\documentclass[10pt,twocolumn,letterpaper]{article}

\usepackage[pagenumbers]{wacv} %

\usepackage{graphicx}
\usepackage{amsmath}
\usepackage{amssymb}
\usepackage{booktabs}

\usepackage[pagebackref,breaklinks,colorlinks]{hyperref}

\usepackage[capitalize]{cleveref}
\crefname{section}{Sec.}{Secs.}
\Crefname{section}{Section}{Sections}
\Crefname{table}{Table}{Tables}
\crefname{table}{Tab.}{Tabs.}

\usepackage{times}
\usepackage{epsfig}
\usepackage{mathtools}
\usepackage{pgfplotstable}
\usepackage{pgfplots}
\usepackage{siunitx}
\usepackage{subcaption}
\usepackage{algorithm}
\usepackage[newfloat,frozencache,cachedir=.]{minted}
\usepackage{caption}
\usepackage{svg}
\usepackage{pifont}
\usepackage{rotating}
\usepackage{tabularray}
\usepackage[export]{adjustbox}

\pgfplotsset{compat=1.18}
\usetikzlibrary{plotmarks}
\usetikzlibrary{shapes}
\usepgfplotslibrary{units} %

\usepackage{tcolorbox}

\tcbuselibrary{minted,breakable,xparse,skins}

\definecolor{bg}{gray}{0.95}
\DeclareTCBListing{mintedbox}{O{}m!O{}}{%
  breakable=true,
  listing engine=minted,
  listing only,
  minted language=#2,
  minted style=tango,
  minted options={%
    linenos,
    gobble=0,
    breaklines=false,
    breakafter=,,
    fontsize=\small,
    numbersep=8pt,
    #1},
  boxsep=0pt,
  left skip=0pt,
  right skip=0pt,
  left=20pt,
  right=0pt,
  top=3pt,
  bottom=3pt,
  arc=5pt,
  leftrule=0pt,
  rightrule=0pt,
  bottomrule=2pt,
  toprule=2pt,
  colback=bg,
  colframe=orange!70,
  enhanced,
  overlay={%
    \begin{tcbclipinterior}
    \fill[orange!20!white] (frame.south west) rectangle ([xshift=15pt]frame.north west);
    \end{tcbclipinterior}},
  #3}
  
\newenvironment{code}{\captionsetup{type=Algorithm}}{}
\SetupFloatingEnvironment{algorithm}{name=Source Code}

\def\wacvPaperID{559} %
\def\confName{WACV}
\def\confYear{2025}

\usepackage{eccvabbrv}
\makeatletter
\robustify\@latex@warning@no@line
\makeatother
\usepackage{authblk}

\newcommand{\qheading}[1]{\noindent\textbf{#1:}}
\newcommand{\zheading}[1]{\textbf{#1:}}

\newcommand{\modelname}{\textit{JointDiffusion}\xspace}%

\begin{document}

\title{A Versatile and Differentiable Hand-Object Interaction Representation}

\author[1]{Théo Morales}
\author[2]{Omid Taheri}
\author[3]{Gerard Lacey}
\affil[1]{Trinity College Dublin} \affil[2]{Max Planck Institute for Intelligent Systems}
\affil[3]{Maynooth University}
\affil[ ]{
\tt\small tmorales@tcd.ie, 
\tt\small omid.taheri@tuebingen.mpg.de,
\tt\small gerard.lacey@mu.ie
}

\twocolumn[{%
\renewcommand\twocolumn[1][]{#1}%
\maketitle
\begin{center}
    \centering
    \captionsetup{type=figure}
    \includegraphics[width=.95\textwidth]{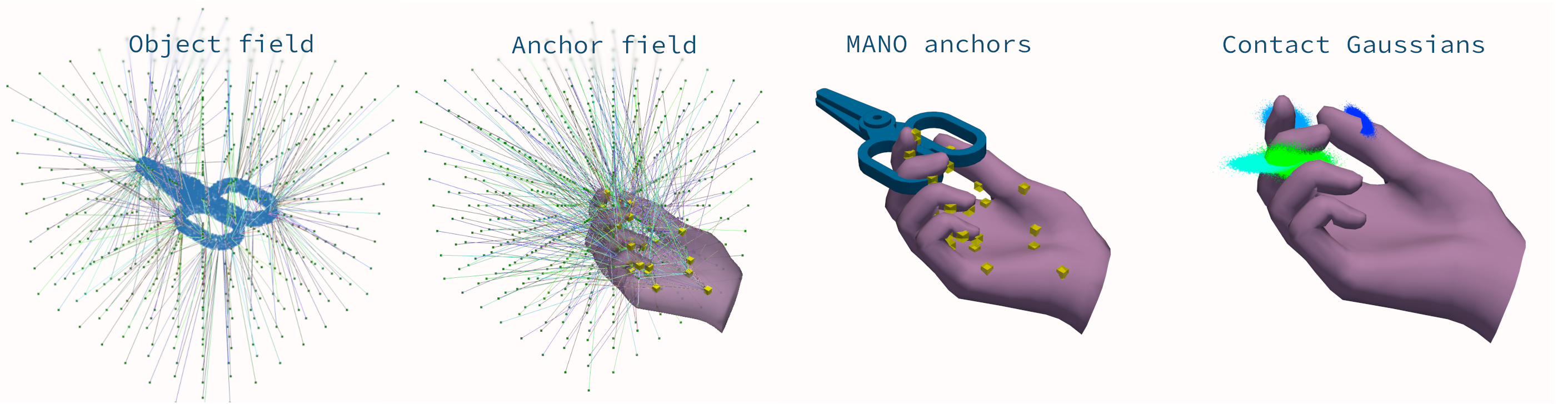}
    \captionof{figure}{ 
     Decomposition of our Coarse Hand-Object Interaction Representation (CHOIR).
     From left to right, our representation encodes the object geometry with point-wise distances in a regular grid
     (coloured rays), the hand shape and pose as point-wise distances to 32 MANO anchors on the mesh surface (yellow cubes),
     and the hand contact points as probability densities from 3D Gaussian distributions (coloured point clouds).
     CHOIR is a fully-differentiable, versatile representation of the hand-object pair in object frame.
     } \label{fig:choir}
\end{center}%
}]
\thispagestyle{empty}

\begin{abstract}
Synthesizing accurate hands-object interactions (HOI) is critical for applications in
Computer Vision, Augmented Reality (AR), and Mixed Reality (MR). Despite recent
advances, the accuracy of reconstructed or generated HOI leaves room for refinement. 
Some techniques have improved the accuracy of dense correspondences by shifting
focus from generating explicit contacts to using rich HOI fields. Still, they lack
full differentiability or continuity and are tailored to specific tasks. 
In contrast, we present a Coarse Hand-Object Interaction Representation (CHOIR),
a novel, versatile and fully differentiable field for HOI modelling. CHOIR leverages
discrete unsigned distances for continuous shape and pose encoding, alongside
multivariate Gaussian distributions to represent dense contact maps
with few parameters. To demonstrate the versatility of CHOIR
we design \modelname, a diffusion model to learn a grasp distribution
conditioned on noisy hand-object interactions or only object geometries, for
both \it{refinement} and \it{synthesis} applications.
We demonstrate \modelname's improvements over the SOTA in both applications:
it increases the contact F1 score by $5\%$ for refinement and decreases the sim. displacement by 
$46\%$ for synthesis. Our experiments show that \modelname with CHOIR yield superior contact accuracy
and physical realism compared to SOTA methods designed for specific tasks. 
Project page: \href{https://theomorales.com/CHOIR}{https://theomorales.com/CHOIR}
\end{abstract}

\section{Introduction}
Numerous computer vision applications could benefit from highly accurate hand
pose prediction in object manipulation scenarios, such as Augmented Reality (AR)
or Mixed Reality (MR), human-robot collaboration, etc. However, SOTA models
still struggle to generalize to novel grasps on unknown objects
\cite{hasson_2019_obman, corona2020ganhand}, in both synthesis and
reconstruction. The problem is challenging because hands are small, dexterous,
with many degrees of freedom, making it hard to be accurately tracked or
reconstructed. Additionally, interactions naturally come with occlusions or
noisy observations, making it harder to estimate accurate hand-object
interactions. Such inaccuracies, like subtle hand-object penetrations or
slightly off-positioned fingers, can significantly affect the realism of the
hand-object interactions. A common approach is to train a model to reason in 3D
space and predict coarse hand-object poses from images, and then refine them
with a model trained on hand-object contacts \cite{Grady2021ContactOptOC,
Zhou2022TOCHSO, Jiang2021HandObjectCC, Wang2022InteractingHP,
Aboukhadra2022THORNetEG}. This coarse-to-fine approach generates an estimate of
how an unknown object is being grasped, while dense hand-object interactions
(learned or simulated) serve as a test-time optimization (TTO) objective to
refine the estimate.

Recently, there has been progress in dense contact map prediction, either
directly from images or meshes \cite{Yu2022UVBased3H, Grady2021ContactOptOC,
Jiang2021HandObjectCC, Hu2023LearningEC, Cho2023TransformerbasedUR}. However,
they still have some limitations such as: compute intensity (typically involving
point cloud processing), the need for feature engineering
\cite{Grady2021ContactOptOC}, and being uninformative in cases where the hand is
approaching but has not yet touched the object. To address these issues, recent
methods have proposed to define the hand pose and shape in an object-centric
space using ray casting or spring systems \cite{Zhou2022TOCHSO, Yang2020CPFLA}.
While these hand-object representations directly improve the refinement
capabilities of TTO, they still require engineering, are compute-intensive, or
are not fully differentiable. To address the gaps, we propose to use a
\textit{Coarse Hand-Object Interaction Representation}, named CHOIR, a novel
field leveraging unsigned distances and multivariate Gaussian distribution to
represent shape, pose, and contact maps for hand-object interactions. 
CHOIR encodes the object geometry as distances to the fixed Basis Point Set
representation (BPS) \cite{bps}, and the hand pose and shape as distances from
the same basis points to the fixed MANO anchors proposed by
\cite{Yang2020CPFLA}. In addition, CHOIR encodes coarse contact maps represented
as 3D Gaussian distributions around the MANO anchors, such that dense contact
maps can be inferred from probability densities.
As such, it is scalable, fully differentiable, and efficient on GPUs.
To demonstrate its effectiveness, we train a conditional Denoising Diffusion
Probabilistic Model (DDPM), named \modelname, to learn the distribution of
hand-object interactions in CHOIR representation. We demonstrate plausible grasp
synthesis alongside noisy grasp refinement through the same model architecture
trained on different condition variables.

Overall, experiments demonstrate that our method outperforms baselines on
denoising and generating static hand interactions and that our approach offers
superior contact-based metrics. Our models and code will be available for
research purposes.

In the direction of solving hand-object interaction challenges, this work makes the following key contributions:
\begin{itemize}
    \item  We propose CHOIR, a versatile and differentiable representation
    that encodes hand-object interactions, enhancing accuracy in contact modelling.
    \item Our method introduces a novel way to represent dense contact maps using
    Gaussian distributions, leading to more accurate hand-object contacts.
    In addition, we propose a novel and simple way to compute contact weights
    for all hand vertices.
    \item We employ a multimodal conditional diffusion model tailored to our
    CHOIR framework, which works for both synthesizing plausible grasps and
    refining noisy ones.
\end{itemize}

\section{Related works}

Despite many advances in hand motion tracking or reconstruction, estimating accurate hand-object interaction poses is still a challenging and unsolved problem. Recently, there has been a push towards the coarse-to-fine paradigm for hand-object interaction, where a coarse hand pose is first generated or reconstructed, and then is refined via optimization \cite{Grady2021ContactOptOC, Zhou2022TOCHSO, Hasson2020PhotometricConsistency, hasson_2019_obman} with pseudo-ground-truth or using learning-based methods ~\cite{GRAB:2020, taheri2023grip}.
In this section, we review the most relevant works and their limitations.

\zheading{ Hand-Object Interaction Reconstruction}
With the growing availability of rich annotated datasets for hand-object interaction  \cite{hand_meshes, Oikonomidis_1hand_object, srinath_eccv2016_handObject}, many recent works focus on simultaneously reconstructing the hands and objects from images \cite{Grady2021ContactOptOC, hasson_2019_obman, Shreyas2020HOnnotate, Brahmbhatt_2020_ECCV, freihand, FirstPersonAction_CVPR2018, h2o}. Many of these works leverage deep learning techniques to estimate the hand and object poses \cite{hasson_2019_obman, Hasson2020PhotometricConsistency}. However, the initial hand and object pose from these methods are often approximate and require further refining. To achieve this, some work optimizes the results further using contact constraints or interaction constraints \cite{zhang2021manipnet, goal2022}. Zhou et al. \cite{Zhou2022TOCHSO} proposed to use a spatiotemporal field for hand-object interaction and train a network to refine this instead. They then use the refined field in a two-step optimization process to get the refined hand poses. This is however slow due to the field not being fully differentiable and requiring a search algorithm.
This pre-optimization imposes a lower bound on the optimization time. 
Here we propose a lightweight and fully differentiable field on which we can optimize hand meshes solely based on the L2 norm.

\zheading{Grasp Synthesis}
Grasp synthesis, split into static and dynamic domains, has received much attention recently. In the static domain, many classic methods use physical constraints to satisfy realistic grasps  \cite{pollard2005physically,sahar2011garsp3d, kry2006interaction, Bohg2014DataDrivenGrasp, li2007datadrivenGrasp}. Newer methods take a learning-based approach and use big datasets of hand-object interactions to learn grasps \cite{GRAB:2020, Corona_2020_CVPR, contactDB_2019, Brahmbhatt_2020_ECCV, jiang2021graspTTA, Karunratanakul2021SkeletonDriven, karunratanakul2020graspField, Zhu2021DexterousGrasping}. These often generate the pose parameters of a model directly \cite{contactDB_2019, Brahmbhatt_2020_ECCV}, or estimate an implicit representation for the grasp \cite{karunratanakul2020graspField, Zhou2022TOCHSO}.

Another body of work focuses on generating dynamic grasps. Similar to static grasps, some methods define contact constraints and use optimization to satisfy them \cite{liu2009dextrous, mordatch2012contactInvariant, zhao2013robustRealTime, liu2012synthesisHandonly, yeL2012synthesis, goal2022}. For better motion realism, recent methods use reinforcement learning (RL) for hand grasp generation \cite{openAI2020rubic, rajeswaran2018learningComplex, Bergamin2019DReCon, Park2019UnorganizedMotion, peng2018deepmimic, Peng2017DeepLoco,christen2022dgrasp}. However, in both static and dynamic grasp generations, the grasps are mostly inaccurate and require further refinement. To further refine grasps there exist optimization-based methods \cite{jiang2021graspTTA, Grady2021ContactOptOC, Zhou2022TOCHSO, zhang2021manipnet} or learning-based ones \cite{GRAB:2020, taheri2023grip}. Some methods like  \cite{GRAB:2020, zhang2021manipnet} directly refine the hand pose. Instead of directly operating on the poses, \cite{Zhou2022TOCHSO} proposes to refine an implicit interaction field for the hand motions and then use it in an optimization process to refine the hand poses. This, however, is very slow due to the complicated nature of the proposed interaction field.
Here we train a diffusion model on our novel CHOIR representation, where we can both generate and refine hand-object interaction by only conditioning the model on different observations.

\zheading{Hand-Object Interaction Representation}
Implicit representations are increasingly gaining traction in the field,
especially for hand-object interaction representation. The Grasping Field,
proposed by Karunratanakul et al. \cite{Karunratanakul2020GraspingFL},
introduces an SDF with hand-parts labels. While it has the advantages of our
proposed interaction field, namely being coarse and distance-based, it primarily
utilizes whole hand and object point clouds as inputs, leading to a
high-dimensional model. Their method, however, does not emphasize grasp
refinement or denoising. In contrast, ContactOpt \cite{Grady2021ContactOptOC}
advocates for a dense contact map based on hand and object meshes. Although
high-dimensional, it offers significant value in grasp refinement and denoising.

Yang et al. \cite{Yang2020CPFLA} propose CPF with coarse anchors and an
innovative spring system. Despite its promising direction, it involves
minimizing relatively intricate energy functions, proving time-consuming at TTO.
Furthermore, its inability to handle non-full or dynamic grasps presents
limitations. Jiang et al. \cite{Jiang2021HandObjectCC} introduced a distinct
approach with hand-object contact consistency reasoning. By employing a CVAE for
initial predictions followed by a contact network, they present dense contact
maps enriched with prior contact regions. Diverging from the above methods, Yu
et al. \cite{Yu2022UVBased3H} used a UV-Based 3D hand-object reconstruction for
grasp optimization. While valuable, its image-centric nature is not well-suited
for hands and fingers that are not in direct contact with the object.
Additionally, the proposed gSDF by Chen et al. \cite{Chen2023gSDFGS}, despite
its precise functionalities remain to be explored further.

The SOTA in grasp refinement is represented by TOCH \cite{Zhou2022TOCHSO},
recognized for its features like accommodating approaching hands and fingers not
in direct contact. While it excels in dynamic grasps, it has not been evaluated
for static ones. In this work, we focus on designing a contact-dense, expressive
interaction field for multiple applications.

\section{Method}
At the heart of our approach is the development of a novel representation for
hand-object interactions, coined CHOIR, which addresses the limitations inherent
in state-of-the-art techniques, particularly those relying on ray-based dense
correspondence fields such as TOCH\cite{Zhou2022TOCHSO}. Our representation is
designed to encode shape, pose, and contacts while remaining fully
differentiable and continuous. We exploit it in two settings: (1) to refine
grasps from noisy predictions of an off-the-shelf method for hand-object
reconstruction, and (2) to synthesize realistic grasps given an object shape. To
do so, we design a DDPM backbone based on the U-Net architecture
which jointly decodes the contact parameters and unsigned
distances from a shared latent space for efficient learning.

In this section, we first go through the details of the proposed representation
and its implementation. We then describe the DDPM and context encoder.
\subsection{Coarse hand-object interaction representation (CHOIR)} \label{sec:choir}
\begin{figure}[tb]
    \centering
    \begin{subfigure}{.49\columnwidth}
        \centering
        \includegraphics[width=0.8\columnwidth]{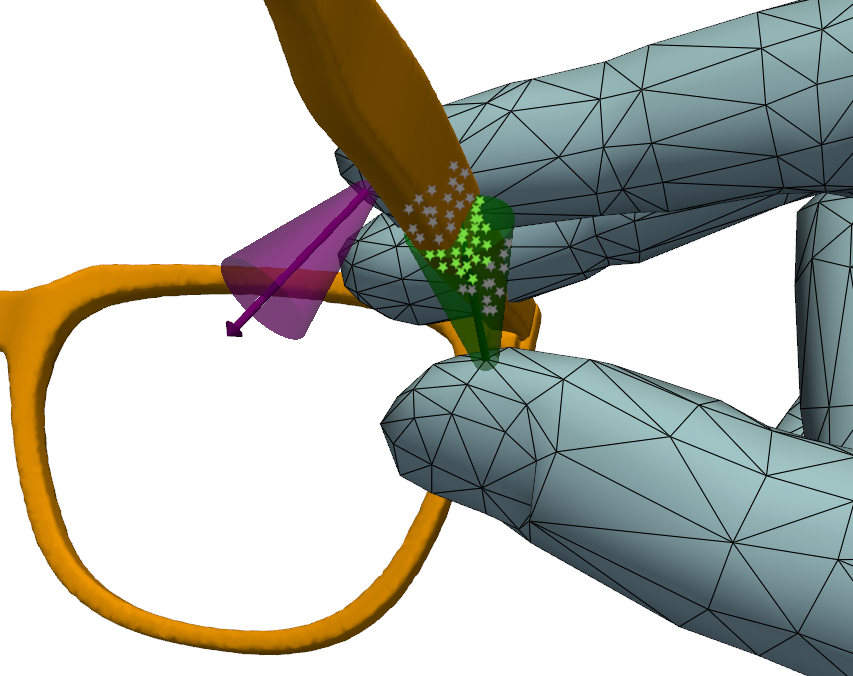}
    \end{subfigure}%
    \begin{subfigure}{.49\columnwidth}
        \centering
        \includegraphics[width=0.8\columnwidth,trim={3cm 2cm 1cm 0},clip,left]{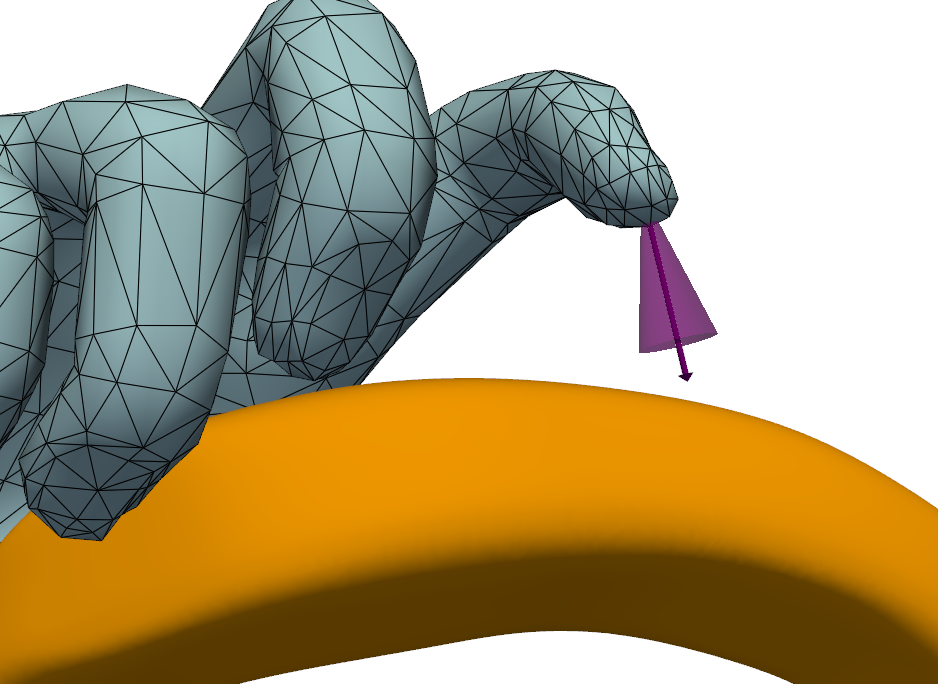}
    \end{subfigure}
    \caption{
    Illustration of the cone of tolerance used to determine the raw hand contact
    weights. For each hand vertex, the weights are the count of object points
    inside the vertex's cone defined along its normal vector. (Left) The green
    points on the object's surface are inside the cone, hence contributing to the
    hand vertex's weight while the grey points do not. (Left \& Right) No
    object points are inside the purple cone: its vertex has a contact
    weight of 0.
    }
    \label{fig:cone_of_tolerance}
\end{figure}

\qheading{Notation} In the following, we denote a dataset sample as $X$ and a 3D vector as
$\boldsymbol{x}$. Superscripts indicate a sample index while subscripts indicate a point index. We differentiate
vectors from scalars by bold symbols for the former. Note that $x$ on its own denotes a generic data
sample of any form.\\

The proposed Coarse Hand-Object Interaction Representation called
\textit{CHOIR}, is a novel field for representing hand-object interaction using
unsigned distances and multivariate Gaussian distributions. The object geometry
and relative hand pose are represented with unsigned distances from a common set
of points, following the Basis Point Set (BPS) representation \cite{bps}. It is
a lightweight 3D point cloud representation with fixed dimensionality that
enables the use of convolutions with a regular point grid. Define a dataset
$\mathcal{D} = \{X^1, \ldots, X^N\}$ consisting of $N$ point clouds where each
point cloud $X^n$ is composed of points $\{\boldsymbol{x}_1^n, \ldots,
\boldsymbol{x}_{K_n}^n\}$. A basis point set $\mathcal{B} = \{\boldsymbol{b}^1,
\ldots, \boldsymbol{b}^M\}$ is defined as a regular grid in $\mathbb{R}^3$.
Then, the dataset is normalized such that $X^n$ fits in the grid and its
centroid is at the origin. Finally, the BPS representation of each point cloud
$X^n$ is computed as
\begin{equation}
X_{\text{BPS}}^n = 
\begin{bmatrix}
    \min\limits_{k} \lVert \boldsymbol{b}^1 - \boldsymbol{x}_k^n \rVert_2^2 \\
    \min\limits_{k} \lVert \boldsymbol{b}^2 - \boldsymbol{x}_k^n \rVert_2^2 \\
    \ldots \\
    \min\limits_{k} \lVert \boldsymbol{b}^M - \boldsymbol{x}_k^n \rVert_2^2
\end{bmatrix}.
\end{equation}

CHOIR represents the hand-object interaction as (a) a coarse hand pose in a
canonical object frame using the MANO parametric hand mesh
\cite{MANO:SIGGRAPHASIA:2017}, and (b) probabilistic hand contact points.

\renewcommand{\thesubsubsection}{\thesubsection.\alph{subsubsection}}
\subsubsection{Shape and pose representation}
The pose part of CHOIR is defined as a concatenation of the BPS representation
of the object mesh and the distances from the BPS to  the $32$ pre-assigned
MANO anchors proposed by \cite{Yang2020CPFLA}, \textit{i.e.}, a CHOIR specifies
an object point cloud $X^n$ together with a hand mesh $H$. The anchor distances
$\boldsymbol{d}_H = [d_H^1, \ldots, d_H^M]^T$ are given by 
\begin{equation}\label{eq:anchors}
    d_H^j = \lVert \boldsymbol{b}^j - \delta_{H}(j) \rVert_2^2
\end{equation}
where the function $\delta_H(j)$ returns the anchor for point $\boldsymbol{b}^j$ and hand mesh $H$.
Note that the same MANO anchor can be assigned to multiple basis points.
We propose two assignment schemes $\delta_H$: (1) a repeating pattern of the 32 ordered indices and 
(2) a shuffled version of the latter.
We did not find any difference in accuracy between the two, which we show
quantitatively in \cref{ap:anchors-ass}, thus we use assignment (1).

\subsubsection{Probabilistic contact representation}
Instead of representing hand contact points as a discrete vector mapping each
MANO vertex to a contact weight or binary class,
we opt for a lightweight and continuous representation based on 3D multivariate Gaussian distributions.
Given a hand-object pair $(H, X^n)$, we first compute the contact weights $w_i$ for each MANO vertex $\boldsymbol{v}_i$. 
We define it as the count of all object surface points within a cone of tolerance defined at the root of $\boldsymbol{v}_i$,
in the direction of the vertex normal $\boldsymbol{n}_i$ (see \cref{fig:cone_of_tolerance}).
This approach is inspired by ContactOpt's contact capsules \cite{Grady2021ContactOptOC}, which includes vertices inside the mesh.
With a cone, we exclude these vertices such that the contact map does not encode penetration patches in favour of a simpler optimization objective. 
Effectively, $\boldsymbol{w}_i = |\mathcal{S}|$ where $\mathcal{S}$ is the set of points $\boldsymbol{x}$ obeying the following conditions 
(where we set $\lambda = 4\text{mm}$ and $\kappa = \frac{4}{3}\pi$):
\begin{align}
    \boldsymbol{x} \in X^n,\\
    \lVert \boldsymbol{x} - \boldsymbol{v}_i \rVert \leq \lambda,\\
    \arccos{\frac{\boldsymbol{n}_i \cdot (\boldsymbol{x} - \boldsymbol{v}_i)^T}{\boldsymbol{n}_i \lVert \boldsymbol{x} - \boldsymbol{v}_i \rVert}}
    \leq \kappa.
\end{align}

From this discrete contact map, we fit one 3D multivariate Gaussian distribution
per MANO anchor. This is done using the weighted mesh vertices such that the
probability densities match the location of the vertices with the most contact
weight. In effect, given the set of vertices $\mathcal{V}$ and the set of
associated weights $\mathcal{W}$, we define the multiset $\mathcal{V}_w$
composed of each element $\boldsymbol{v}_i \in \mathcal{V}$ repeated $w_i \in
\mathcal{W}$ times:%
\begin{equation}
    \mathcal{V}_w = \{ 
        \underbrace{%
            \boldsymbol{v}_i, \boldsymbol{v}_i, \ldots, \boldsymbol{v}_i
        }_{w_i \text{times}}
        \text{for all } \boldsymbol{v}_i \in \mathcal{V}, w_i \in \mathcal{W}
    \}.
\end{equation}

\begin{figure}[tb]
    \centering
    \vspace{-4mm}
    \begin{subfigure}[t]{.49\columnwidth}\label{fig:contact_gaussians_full}
        \centering
        \includegraphics[width=0.6\columnwidth]{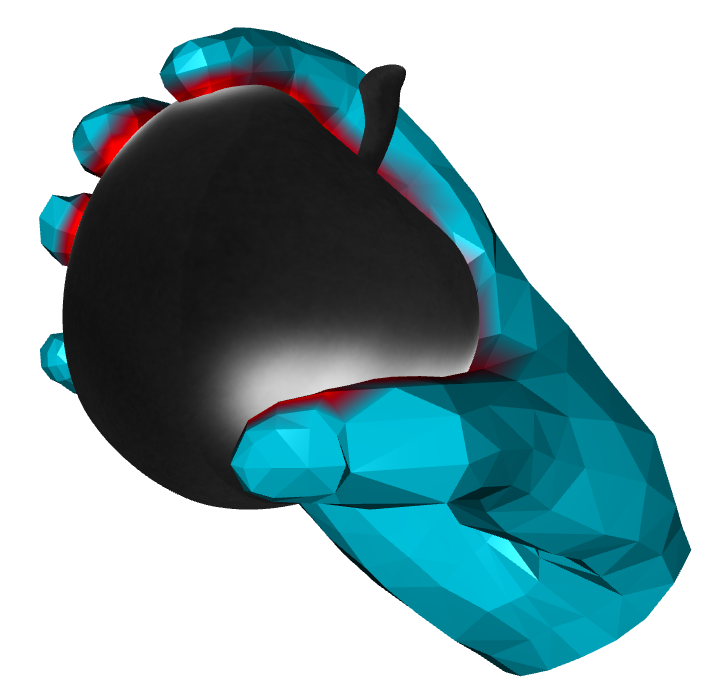}
        \caption{Raw hand contact weights \\in red.}
    \end{subfigure}%
    \begin{subfigure}[t]{.49\columnwidth}\label{fig:contact_gaussians_densities}
        \centering
        \includegraphics[width=0.6\columnwidth]{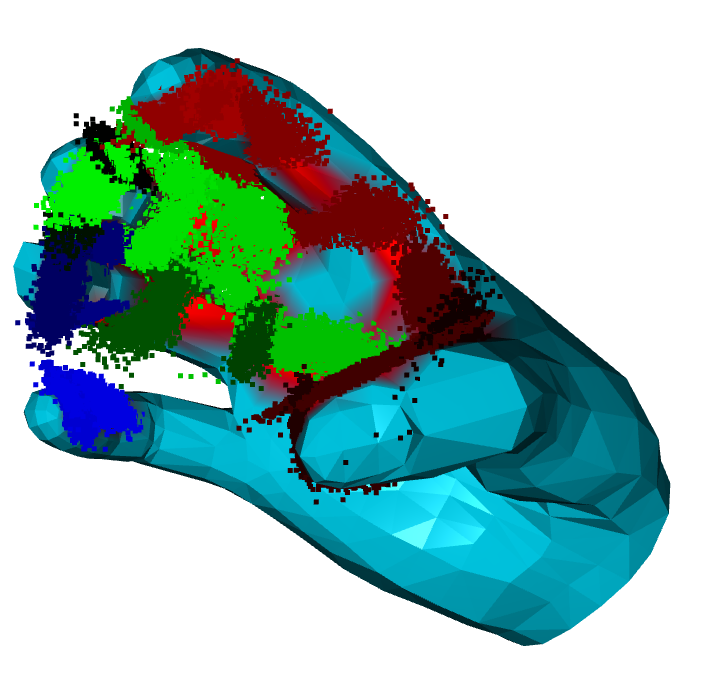}
        \caption{3D Gaussian distributions \\as coloured point clouds.}
    \end{subfigure}\\
    \begin{subfigure}[t]{\columnwidth}\label{fig:contact_gaussians_maps}
        \centering
        \includegraphics[width=0.5\columnwidth]{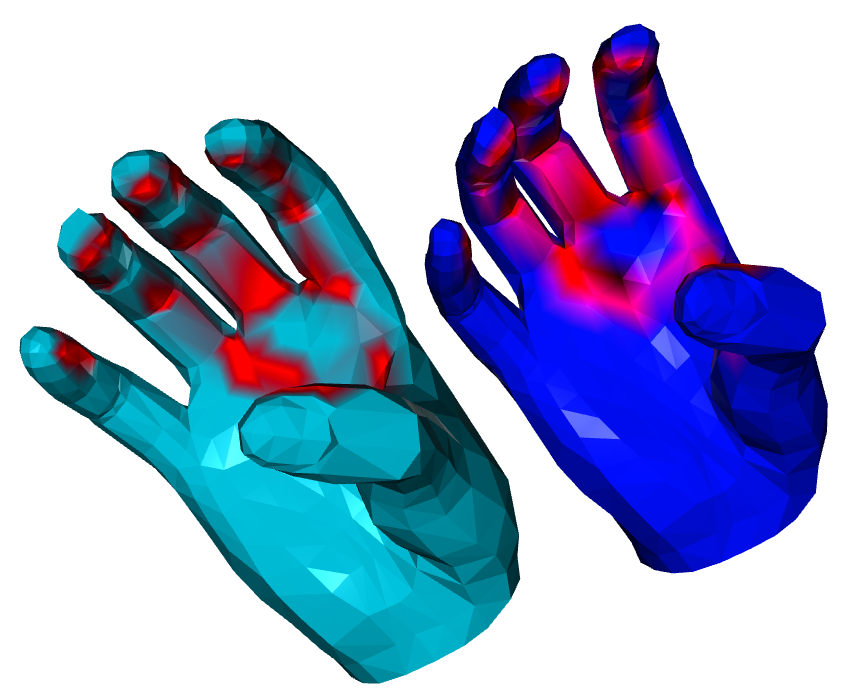}
        \caption{Recovered contact weights (left) \vs raw contact weights (right).}
    \end{subfigure}
    \caption{Visualization of our probabilistic contact maps (best seen in colour).
    (a) The raw hand contact weights are computed with our cone of tolerance method.
    (b) 32 3D Gaussian distributions are fitted -- one for each MANO anchor -- on the weights to obtain
    contact probability densities.
    (c) Comparison of the recovered probabilistic dense contact map and of the raw contact weights.
    Our method leaves gaps in the contact map to allow for a $2mm$ penetration and improve contact fitting.}
    \label{fig:contact_gaussians}
\end{figure}

We then maximize the likelihood of the Gaussian parameters given the multiset $\mathcal{V}_w$.
This allows us to encode a probabilistic dense contact map for the hand mesh as a set of 32 multivariate normal distributions (MVN).
Each hand vertex then gets a contact probability by querying the probability density function of the nearest anchor's Gaussian (see \cref{fig:contact_gaussians}).
For anchor $j$, the MVN is parameterized by a mean vector $\boldsymbol{\mu}^j \in \mathbb{R}^3$ and a covariance
matrix $\Sigma^j \in \mathbb{R}^{3 \times 3}$.
Since the latter must be positive semi-definite, it can be challenging to predict it with a neural network. One approach
is to represent it as a lower triangular matrix $L^j$ obtained via Cholesky decomposition, such that $\Sigma^j = L^j L^{jT}$,
and to enforce the diagonal entries to be positive.%
The final form of our probabilistic contact representation for a given hand mesh $H$ is thus:
\begin{equation}\label{eq:contact_gaussian}
    \boldsymbol{c}_H = \begin{bmatrix}
        \boldsymbol{\mu}^0 & \boldsymbol{l}^0 \\
        \boldsymbol{\mu}^1 & \boldsymbol{l}^1 \\
        \ldots \\
        \boldsymbol{\mu}^{31} & \boldsymbol{l}^{31} \\
    \end{bmatrix} \in \mathbb{R}^{32 \times 9}
\end{equation}
where $\boldsymbol{l}^j \in \mathbb{R}^6$ is the vector containing the elements of and below the diagonal of the matrix $L^j$.
In summary, a CHOIR (see \cref{fig:choir}) is defined as
\begin{equation} \label{eq:choir}
    x_{\text{CHOIR}} = [X_{\text{BPS}}^n \in \mathbb{R}^{M}, \boldsymbol{d}_H \in \mathbb{R}^{M}, \boldsymbol{c}_{H} \in \mathbb{R}^{32 \times 9}].
\end{equation}
By encoding coarse hand-object correspondences in this way,
we can fit hand meshes onto ground-truth CHOIRs with less than $1$mm absolute mean per-joint pose error.
Thus, generating valid CHOIRs is the accuracy bottleneck; in the next subsection, we present our learning method.

\subsection{Learning conditional distributions of CHOIR}

Denoising Diffusion Probabilistic Models (DDPM) have recently made their prowess
in distribution learning for high dimensional problems
\cite{Rombach2021HighResolutionIS}. The combination of recent improvements to
the U-Net architecture and the DDPM framework enables the modelling of complex
relationships between context and target information.
We propose to exploit these advances to model complex conditional CHOIR distributions with multiple
modalities of context, such as noisy hand-object pairs or object shapes.

\begin{figure*}[tb]
    \centering
    \includegraphics[width=0.85\textwidth]{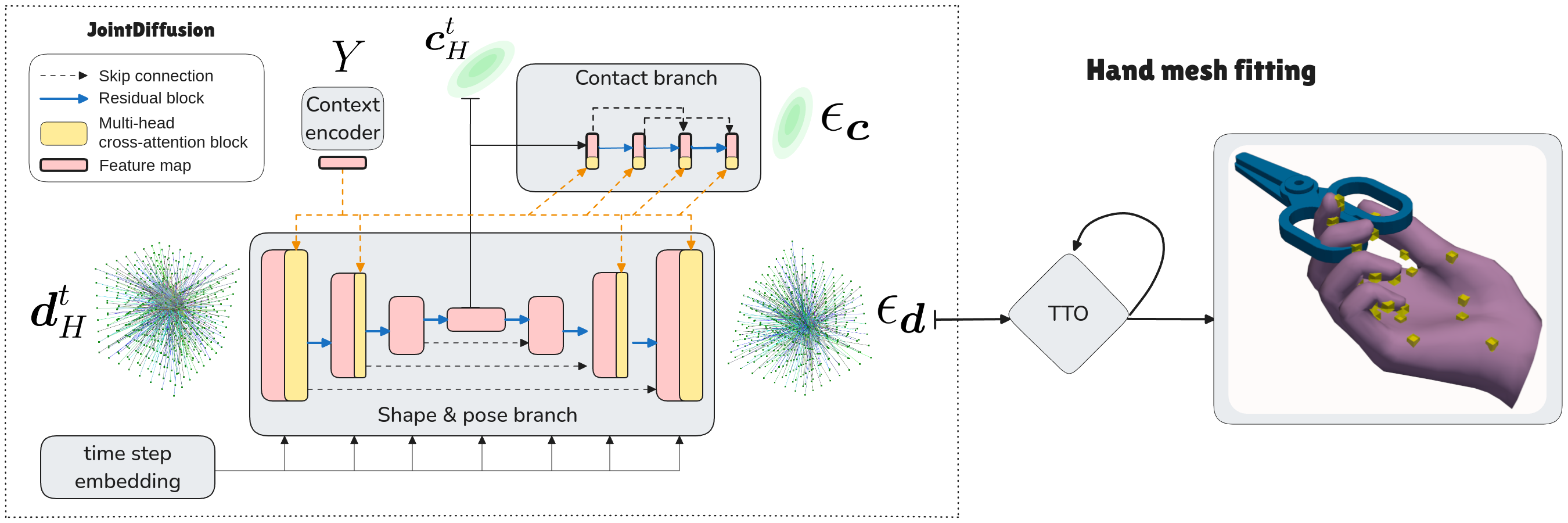}
    \caption{
    Architecture of \modelname. The 3D U-Net predicts the noise sample $\epsilon_{\boldsymbol{d}}$ for the hand distance
    field $\boldsymbol{d}_H$. The contact prediction branch predicts the noise sample $\epsilon_{\boldsymbol{c}}$ for the
    contact Gaussian parameters $\boldsymbol{c}_H$ from the features of the U-Net's bottleneck. This joint learning encourages the U-Net to extract features relevant to both tasks, enhancing the accuracy of the learned CHOIR distribution.
    }
    \label{fig:unet}
\end{figure*}

Our goal is to determine the conditional distribution of hand poses 
$p(\boldsymbol{d}_H, \boldsymbol{c}_H | y)$ based on an observation $y$, where $\boldsymbol{d}_H$ and $\boldsymbol{c}_H$ are parts of CHOIR (see \cref{eq:choir}).  The context $y$ is either (1) a noisy hand-object pair, encoded as a CHOIR with missing contacts $\boldsymbol{c}_H$, or (2) an object point cloud encoded as $X_\text{BPS}^n \in \mathbb{R}^M$, \ie a CHOIR
with missing contacts and hand pose.
To learn this distribution we separately predict the noise samples for the distance field $\boldsymbol{d}_h$ and the contact Gaussians $\boldsymbol{c}_H$, denoted $\boldsymbol{\epsilon}_{\boldsymbol{d}}$ and $\boldsymbol{\epsilon}_{\boldsymbol{c}}$ respectively.
This is motivated by the structure of $\boldsymbol{d}_H$ which allows the use of convolutions, while $\boldsymbol{c}_H$ is a vector in $\mathbb{R}^{32\times9}$.
Thus, our DDPM backbone (see \cref{fig:unet}) is composed of a 3D U-Net for the prediction of  $\boldsymbol{\epsilon}_{\boldsymbol{d}}$,
and of a second decoder for the prediction of $\boldsymbol{\epsilon}_{\boldsymbol{c}}$. This contact decoder is a fully connected residual network 
whose input is a concatenation of the latent variable $\boldsymbol{z}_t$ and the latent features from the bottleneck layer of the U-Net.
This encourages the model to learn pose and contact features in a shared space, such that the latent codes are relevant to both
$\boldsymbol{d}_H$ and $\boldsymbol{c}_H$.

Our U-Net implementation uses Multi-Head Self-Attention (MHSA) to encourage relevant feature extraction and Multi-Head Cross-Attention (MHCA) to condition the network on the context.
The latter is embedded with an encoder identical to the U-Net encoder, with an additional spatial pooling mechanism via a shallow fully-connected network. 
We train one model per modality of context and include experiments on a multi-modal model in \cref{ap:multimodal}.

Ultimately, we propose to learn the conditional distribution $p(\boldsymbol{d}_H, \boldsymbol{c}_H | y)$ in two settings:
\begin{enumerate}
    \item Where $y$ is a noisy observation of a CHOIR, with missing contacts $\boldsymbol{c}_H$, such that\\
    $x_\text{CHOIR}=[X_{\text{BPS}}^n \in \mathbb{R}^{M}, \boldsymbol{d}_H \in \mathbb{R}^{M}]$. %
    \item Where $y$ is an object point cloud in BPS representation $X_\text{BPS}^n \in \mathbb{R}^M$.
\end{enumerate}
We then sample from this distribution to obtain a full CHOIR.
In the next subsection, we show how to obtain MANO parameters in the object coordinate system.

\subsection{Test-Time Optimization (TTO)}

While the state-of-the-art HOI fields are either not fully differentiable
\cite{Zhou2022TOCHSO} or rely on random restarts \cite{Grady2021ContactOptOC,
Zhou2022TOCHSO}, fitting a hand mesh to CHOIR is done by gradient descent in two
stages. Firstly, we fit a MANO mesh to the unsigned distance field of CHOIR.
Secondly, we adjust the hand contacts to the nearest object points using the
contact Gaussians of CHOIR. These two stages rely on distance minimization with
continuous losses, giving a smoother loss landscape than methods using contact
agreement objectives \cite{Grady2021ContactOptOC, jiang2021graspTTA}.

\subsubsection{Coarse pose and shape fitting stage}
The first stage's objective $\mathcal{L}_\text{PoseShape}$ is composed of a reconstruction loss $\mathcal{L}_{\text{rec}}$,
a shape regularizer $\mathcal{L}_{\text{shape}}$ and a pose regularizer $\mathcal{L}_{\text{pose}}$:
\begin{equation}\label{eq:stage_1_tto}
\begin{split}
    \mathcal{L}_{\text{PoseShape}} &= \lambda_1 \cdot \mathcal{L}_{\text{rec}} + \lambda_2 \cdot \mathcal{L}_{\text{shape}} 
    + \lambda_3 \cdot \mathcal{L}_{\text{pose}}\\
    \mathcal{L}_{\text{rec}} & = \lVert \boldsymbol{d}_H - \hat{\boldsymbol{d}_H} \rVert_2^2\\
    \mathcal{L}_{\text{shape}} &= \lVert \boldsymbol{\beta}_\text{MANO} \rVert_2\\
   \mathcal{L}_{\text{pose}} &= \lVert \boldsymbol{\theta}_\text{MANO} - \boldsymbol{\theta}_\text{MANO}^{\text{init}} \rVert_2 
\end{split}
\end{equation}
where $\boldsymbol{d}_H$ and $\hat{\boldsymbol{d}_H}$ are the respective ground-truth and predicted
anchor distances (see \cref{eq:anchors}, $\boldsymbol{\beta}_\text{MANO}$ and $\boldsymbol{\theta}_\text{MANO}$ 
are the MANO shape and pose parameters, respectively.
The shape regularizer prevents the hand mesh from over-deforming to satisfy the reconstruction loss, 
while the pose regularizer prevents strong deviation from the initial MANO pose estimate
$\boldsymbol{\theta}_\text{MANO}^\text{init}$.
Note that in the grasp synthesis case, we remove the pose regularizer.

We minimize $\mathcal{L}_\text{PoseShape}$ \wrt the MANO parameters alongside rotation and translation of the wrist joint
with the Adam optimizer \cite{Kingma2014AdamAM}.%
We set $\lambda_1$ to $1000$ to bring the loss into the millimetre scale and found $\lambda_2 = \num{1e-4}$ 
and $\lambda_3 = \num{1e-8}$ to work well in practice. 
In the grasp refinement setting, all parameters are initialized with the noisy inputs
from which CHOIR observations are built, leading to fast convergence ($\sim150$ iterations).
Since the pose and shape encodings are unsigned distances, the fitting loss is obtained in very few lines
of Python code (see supplementary material).
This TTO stage fits a hand mesh to the predicted distance field but does not
account for contacts and penetration.
For this, we introduce stage two.

\subsubsection{Dense contact fitting stage}
In the second stage, we refine the obtained hand grasp by minimizing the weighted distances from the MANO vertices to
their nearest object points under some constraints. The weights of vertices $\boldsymbol{v}$ are obtained
from the probability distribution function (PDF) $\Phi_j(\boldsymbol{v}_i)$ of the nearest anchor's contact Gaussian
(see \cref{eq:contact_gaussian}). 
For each MANO vertex $\boldsymbol{v}_i$, nearest anchor $j$ and nearest set of $K$ (such as $5$) object points
 $\mathcal{K}^n$, the reconstruction loss is:
\begin{equation}
    \mathcal{L}_\text{rec} = \sum_{i=1}^{N} \sum_{k=1}^{K} \Phi_j(\boldsymbol{v}_i) \cdot 
    \lVert \boldsymbol{v}_i - \boldsymbol{p}_k \rVert_2^2, \quad \boldsymbol{p} \in \mathcal{K}^n.
\end{equation}
This objective is minimized in conjunction with a penetration regularizer following \cite{Grady2021ContactOptOC},
the shape regularizer defined in \cref{eq:stage_1_tto}, and a pose regularizer to avoid deviating from the initial solution,
defined as
\begin{equation}
    \mathcal{L}_\text{pose} = \eta_1 \cdot \lVert R_\text{MANO} - R_\text{MANO}^\text{stage1} \rVert_2
    + \eta_2 \cdot \lVert \boldsymbol{t}_\text{MANO} - \boldsymbol{t}_\text{MANO}^\text{stage1} \rVert_2
\end{equation}
where $R_\text{MANO}$ and $\boldsymbol{t}_\text{MANO}$ are the respective rotation matrix and translation vector of 
the MANO mesh, $\eta_1 = \num{1e-2}$ and $\eta_2 = \num{1e-1}$.
The final objective is
\begin{equation}
    \mathcal{L}_\text{Contacts} = \lambda_4 \cdot \mathcal{L}_\text{rec} + \lambda_5 \cdot \mathcal{L}_\text{penetration}
    + \lambda_6 \cdot \mathcal{L}_\text{pose} + \lambda_2 \cdot \mathcal{L}_\text{shape}
\end{equation}
where $\lambda_4=10$,  $\lambda_5=1000$,  and $\lambda_6=0.5$. We optimize this loss \wrt the same parameters and with the
same method as in the previous stage.
In the next section, we evaluate the combined CHOIR + \modelname + TTO solution on grasp refinement and synthesis
benchmarks, and show how much each component contributes to the accuracy and plausibility of our grasps.

\section{Evaluation} \label{sec:eval}
\begin{table*}[bt]
    \centering
    \vspace{-4mm}
    \caption{
        Evaluation of our approach on static grasp refinement against SOTA methods 
        on the Perturbed ContactPose benchmark. * means reported figures.
        All methods are evaluated with one non-cherry-picked output per sample.
        \modelname shows greater contact accuracy and outperforms all methods on
        most metrics, especially contact metrics (F1, Precision, Recall) and intersection volume,
        showing greater contact fidelity on the hand locations.
        Best results are in bold and second best are underlined.
    }
    \resizebox{0.8\textwidth}{!}{
        \begin{tabular}{lcccccc}
        \textbf{Method} & \textbf{MPJPE} (mm) $\downarrow$ & \textbf{R-MPJPE} (mm) $\downarrow$ & \textbf{IV} ($\text{cm}^3$) $\downarrow$ & \textbf{F1} (\%) $\uparrow$ & \textbf{Precision} (\%) $\uparrow$ & \textbf{Recall} (\%) $\uparrow$\\
        \hline  
        Perturbed data & 83.02 & 21.55 & 6.99 & 1.55 & 1.88 & 2.74\\
        ContactOpt \cite{Grady2021ContactOptOC} & 32.88 & \underline{28.17} & 12.83* & 17.27 & 13.24 & \textbf{34.30}\\ %
        TOCH \cite{Zhou2022TOCHSO} & \textbf{26.96} & 29.24 & \underline{10.14} & \underline{22.23} & \underline{21.46} & 25.09 \\ 
        \modelname (ours) & \underline{27.69} & \textbf{23.54} & \textbf{6.04} & \textbf{27.20} & \textbf{25.21} & \underline{32.80} \\
        \end{tabular}
    }
    
    \label{tab:contactpose}
\end{table*}

\begin{table*}[tb]
    \centering
    \begin{tabular}{cc|ccc}
        Ground truth & Observation & ContactOpt \cite{Grady2021ContactOptOC} & TOCH \cite{Zhou2022TOCHSO} & \modelname \\
        \includegraphics[scale=0.13]{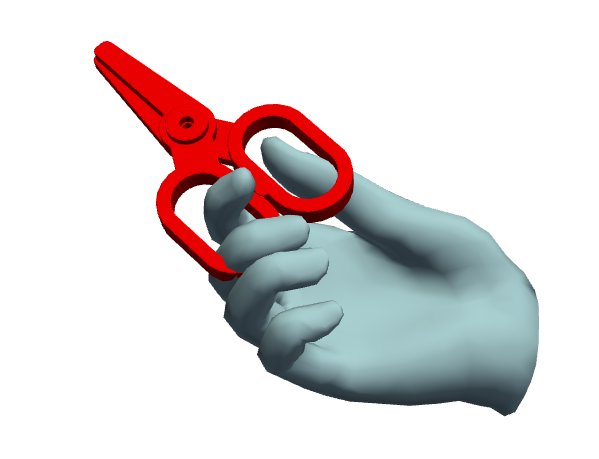} & 
        \includegraphics[scale=0.15]{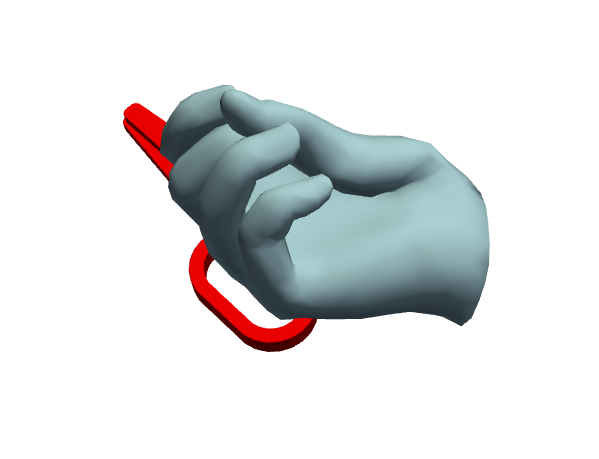} &
        \includegraphics[scale=0.11]{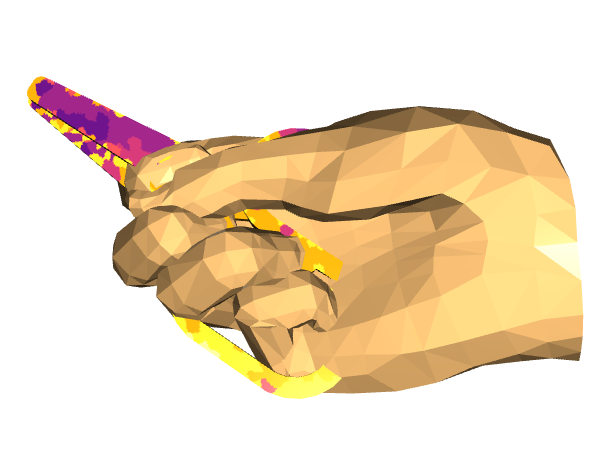} &
        \includegraphics[scale=0.20]{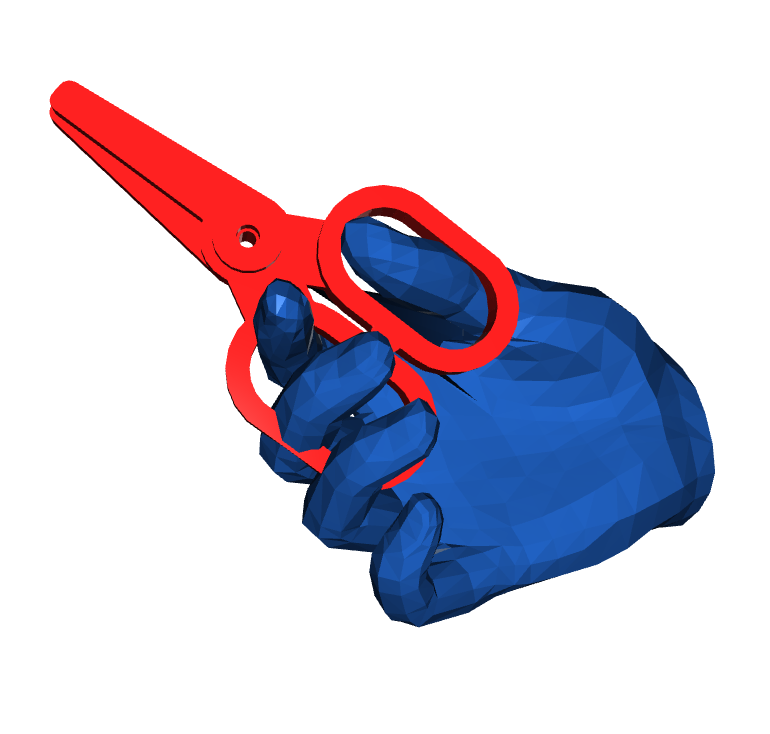} &
        \includegraphics[scale=0.16]{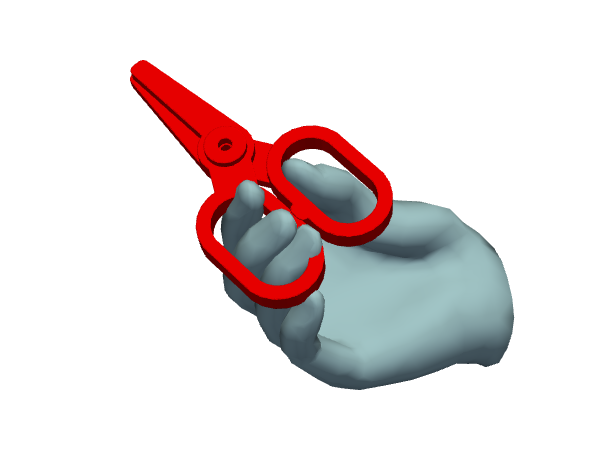} \\ 
        
        \includegraphics[scale=0.12]{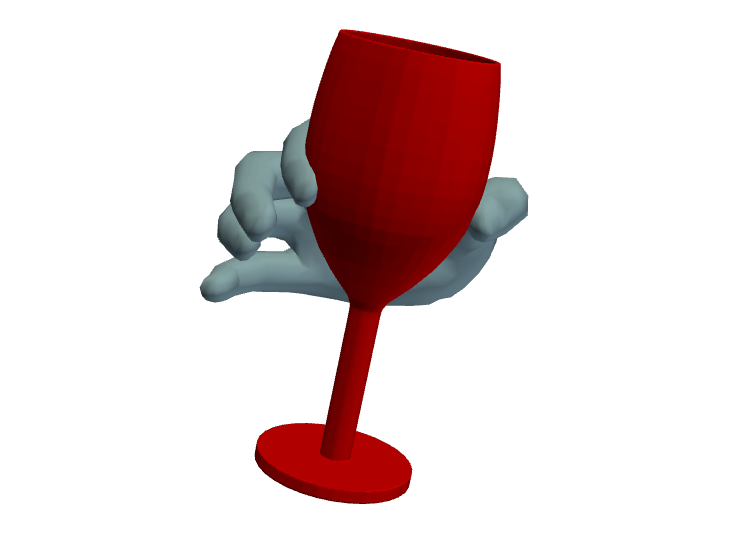} & 
        \includegraphics[scale=0.11]{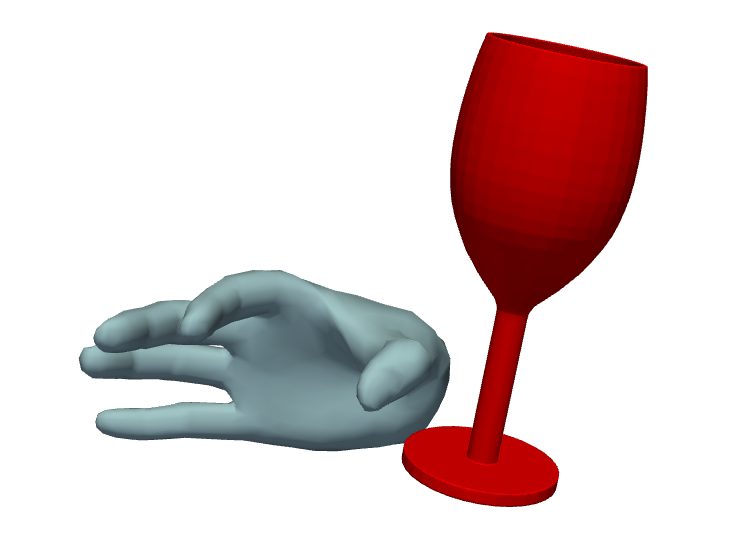} & 
        \includegraphics[scale=0.12]{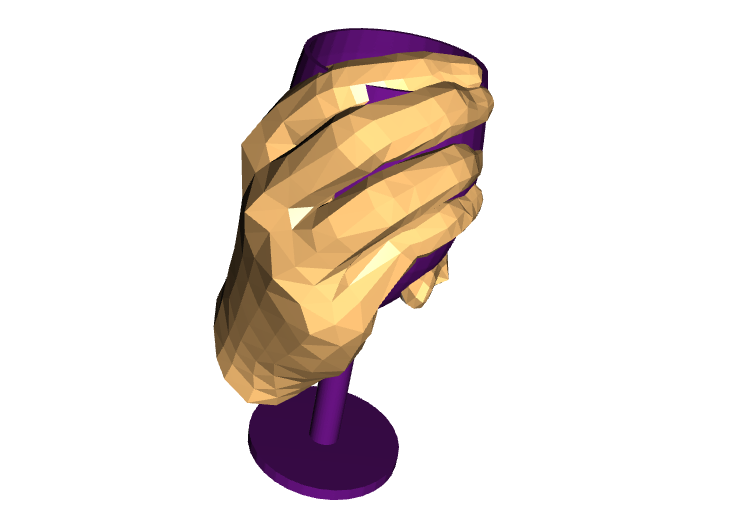} &
        \includegraphics[scale=0.07]{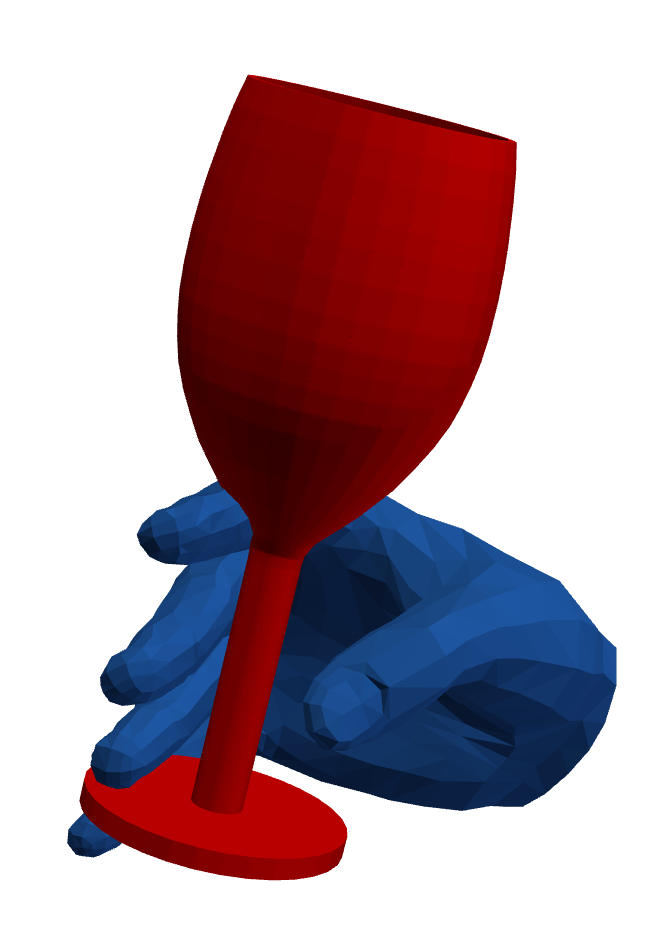} &
        \includegraphics[scale=0.13]{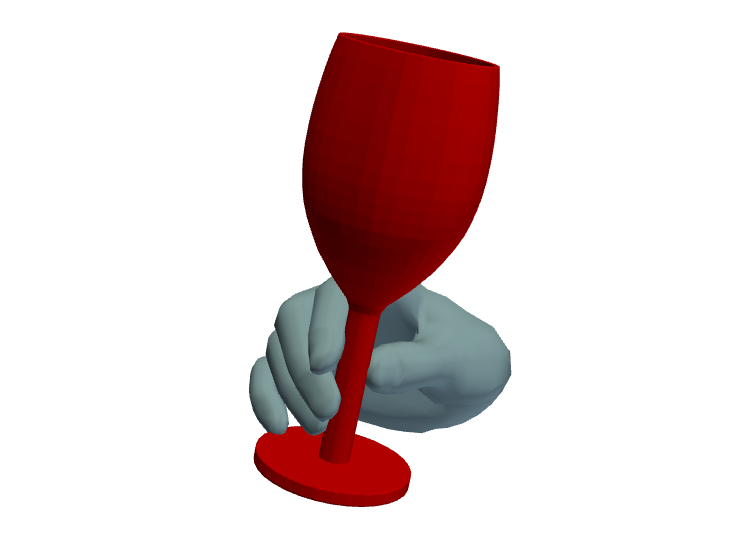} \\ 

        \includegraphics[scale=0.13]{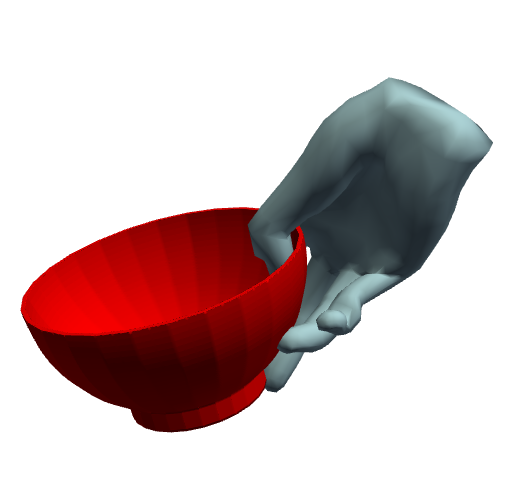} & 
        \includegraphics[scale=0.13]{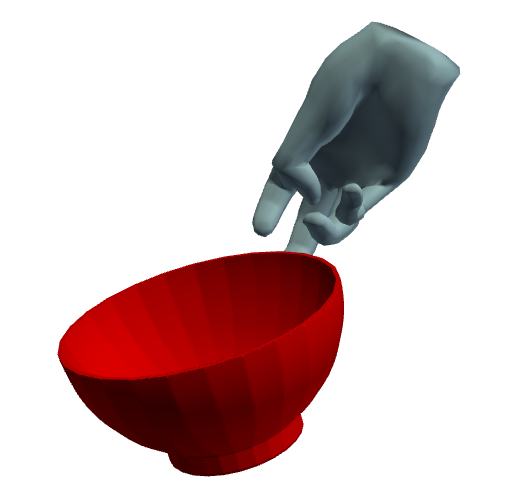} & 
        \includegraphics[scale=0.14]{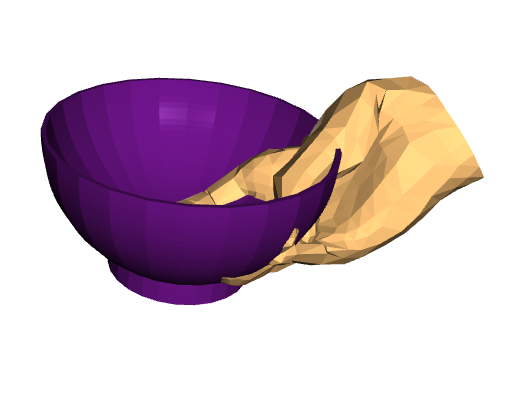} &
        \includegraphics[scale=0.07]{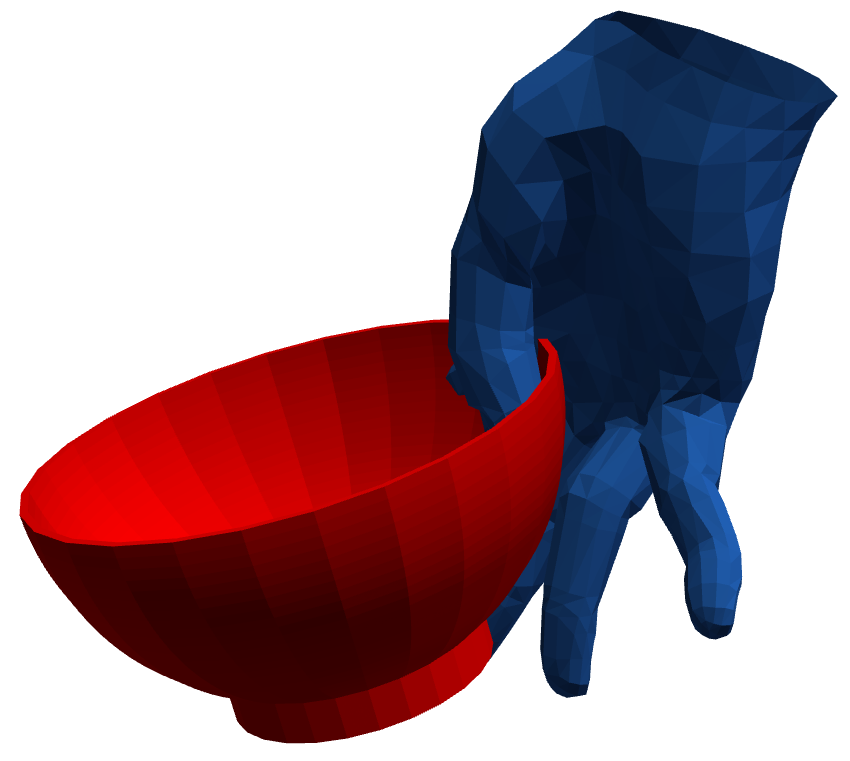} &
        \includegraphics[scale=0.13]{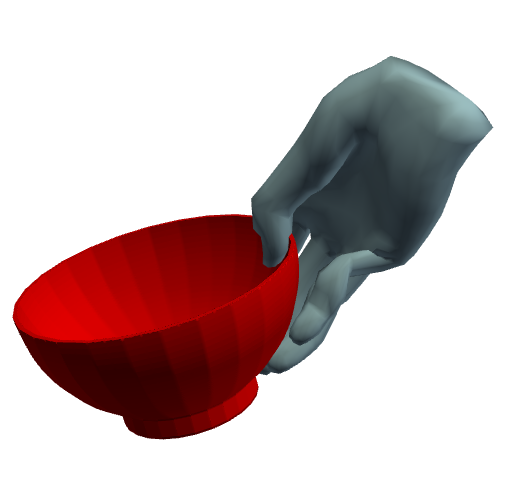} \\

        \includegraphics[scale=0.13]{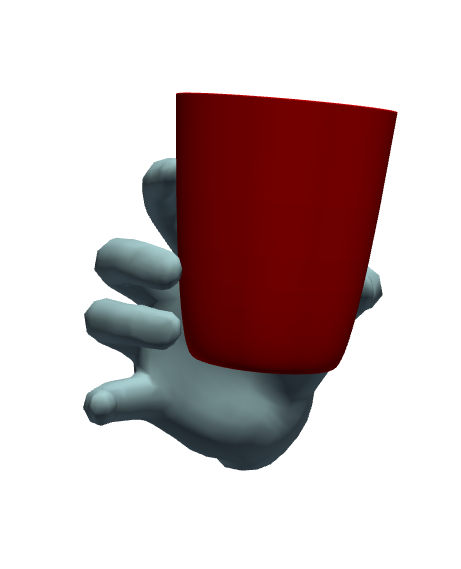} & 
        \includegraphics[scale=0.12]{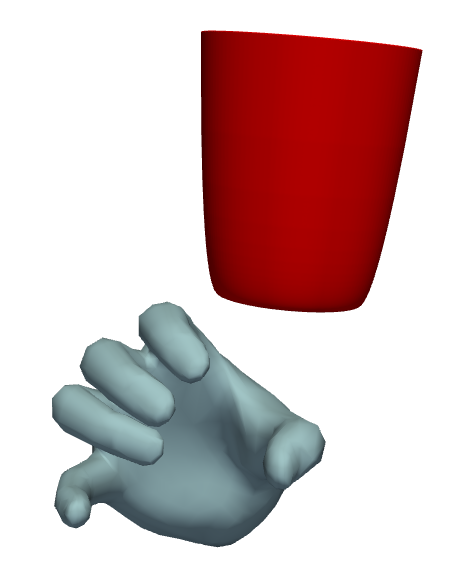} & 
        \includegraphics[scale=0.12]{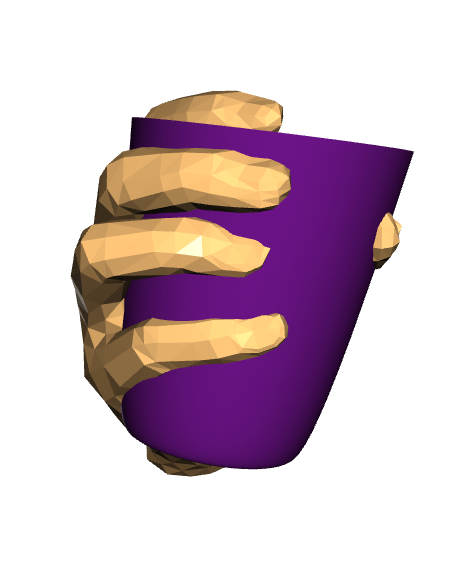} & 
        \includegraphics[scale=0.08]{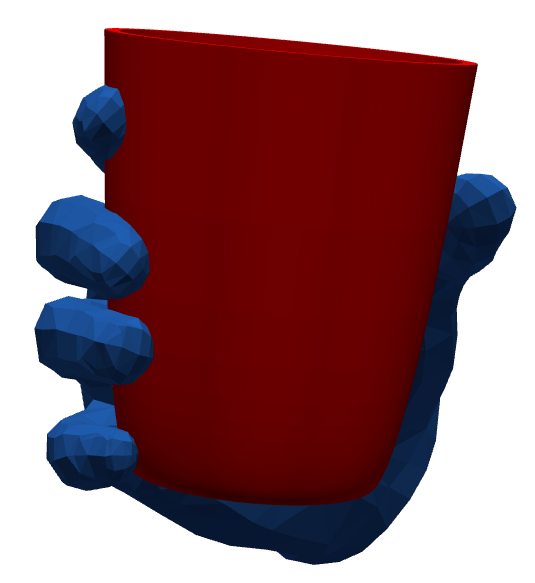} & 
        \includegraphics[scale=0.13]{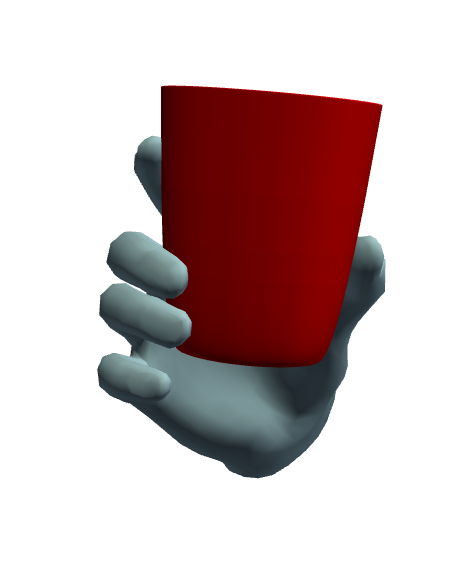} \\ 
    \end{tabular}
    
    \captionsetup{type=figure}
    \captionof{figure}{Qualitative comparison of grasp denoising on one
    challenging case of the Peturbed ContactPose benchmark. Our method produces
    less penetration than TOCH\cite{Zhou2022TOCHSO}, and substantially better
    output than ContactOpt\cite{Grady2021ContactOptOC} which maximizes
    hand-object contact.}
    \label{fig:qualitative_cp}
    \vspace{-0.4cm}
\end{table*}

\begin{table*}[tb]
    \centering
    \begin{tabular}{c|cccc}
    Input object & Sample 1 & Sample 2 & Sample 3 & Sample 4\\
    \includegraphics[scale=0.15,trim={0 0 0 20px},clip]{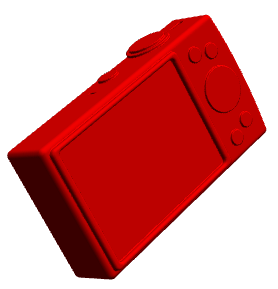} &
    \includegraphics[scale=0.15,trim={0 0 0 20px},clip]{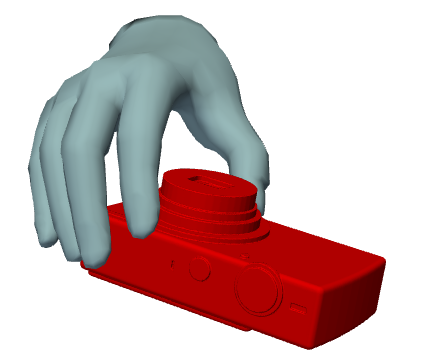} &
    \includegraphics[scale=0.15,trim={0 0 0 30px},clip]{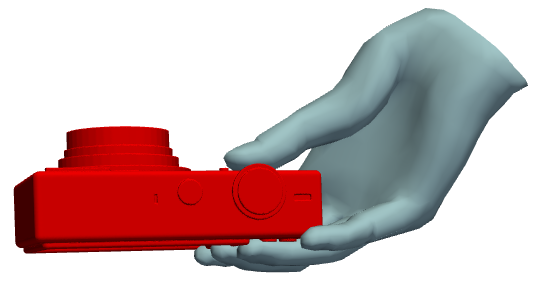} &
    \includegraphics[scale=0.15,trim={0 0 0 3px},clip]{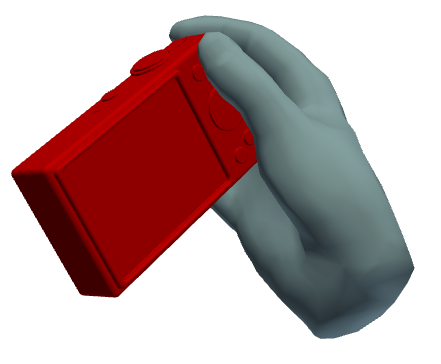} &
    \includegraphics[scale=0.13,rotate=90]{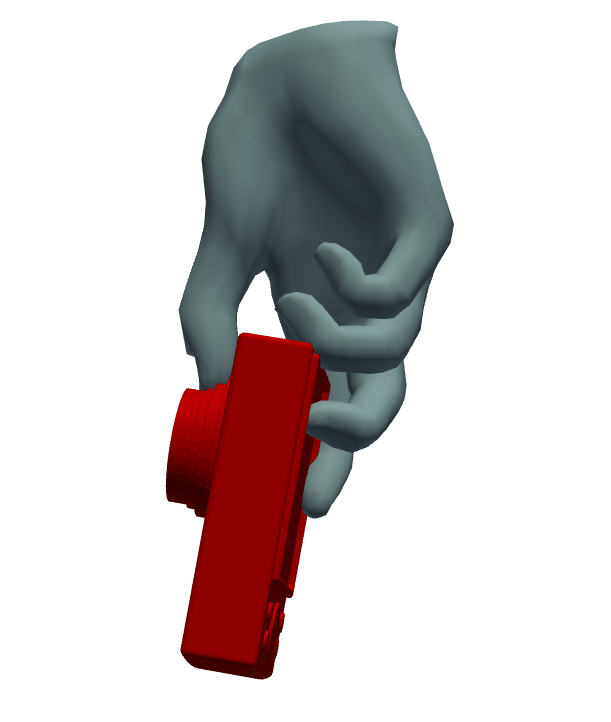} \\
    \includegraphics[scale=0.13]{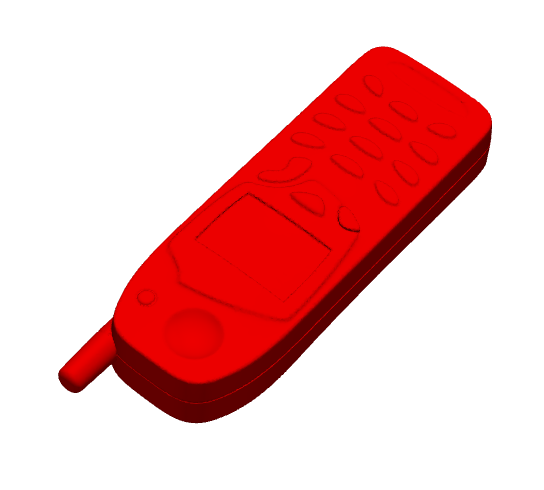} &
    \includegraphics[scale=0.13]{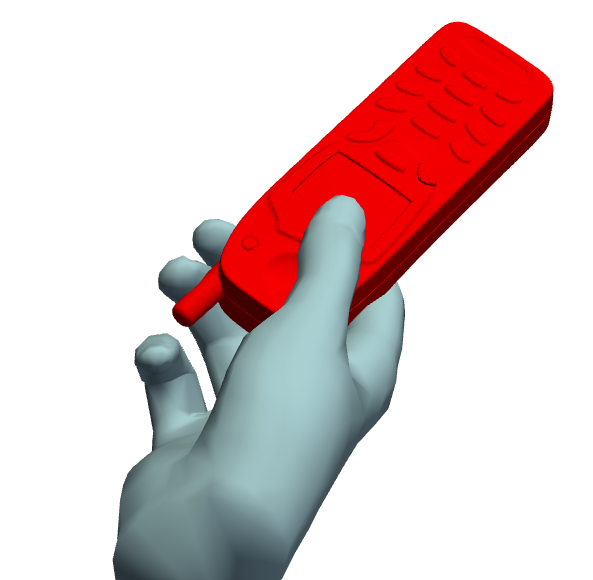} &
    \includegraphics[scale=0.13]{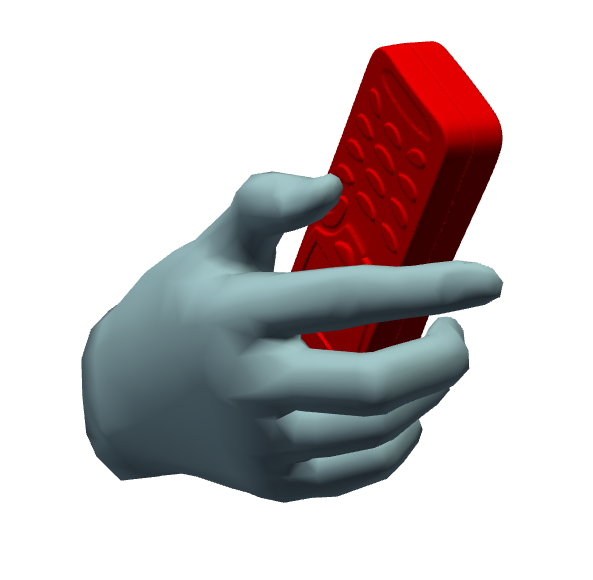} &
    \includegraphics[scale=0.13]{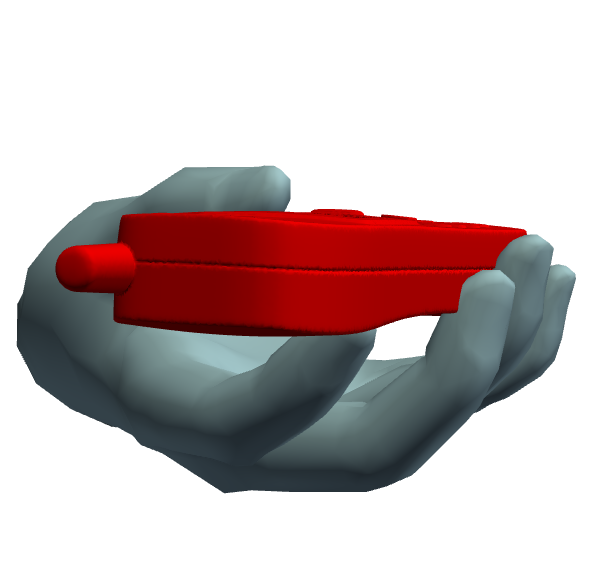} &
    \includegraphics[scale=0.13]{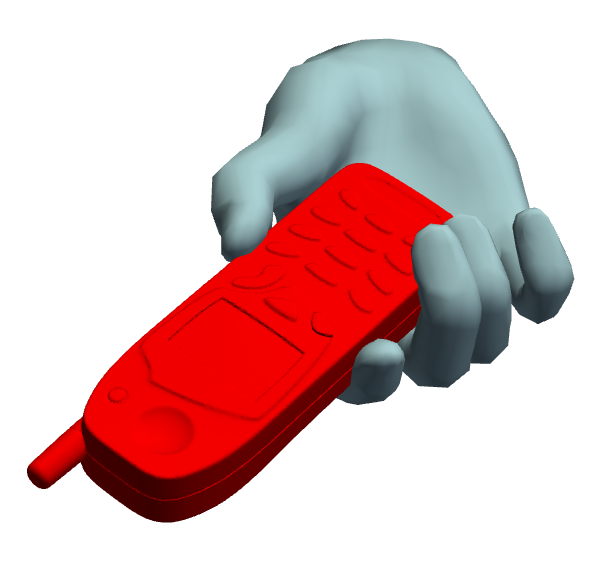} \\
    \end{tabular}
    \captionsetup{type=figure}
    \captionof{figure}{Generated grasps obtained with \modelname on the ContactPose benchmark. The
    synthesized grasps show plausible grasps with good finger-object contact and minimal penetration, showing the expressive contact modelling of CHOIR.}
    \label{fig:synthesis}
    \vspace{-2mm}
\end{table*}

\begin{table}[bt]
    \centering
    \caption{
        Evaluation of our approach against GraspTTA \cite{jiang2021graspTTA} 
        on static grasp generation for the ContactPose benchmark.
        \textdagger \hspace{1px}: contact fitting enabled. Best results are in bold, second best are underlined.
    }
    \resizebox{0.7\columnwidth}{!}{
        \begin{tabular}{lcc}
        \textbf{Method} & \textbf{IV} ($\text{cm}^3$) $\downarrow$ & \textbf{SD} ($\text{cm}$) $\downarrow$ \\
        \hline  %
        GraspTTA \cite{jiang2021graspTTA} & \underline{5.17} & 3.81 \\
        \modelname & 8.13 & \underline{2.07}\\
        \modelname \textdagger & \textbf{4.51} & \textbf{2.05} \\
        \end{tabular}
    }
    \label{tab:generation}
    \vspace{-2mm}
\end{table}

Our solution consists of (a) a representation (CHOIR), (b) a learning method
(\modelname) for denoising and synthesizing interactions, and (c) a hand-mesh fitting algorithm (TTO). We evaluate this
solution in two settings
and in \cref{ap:multimodal}, we evaluate a multi-modal variant
trained in both settings. %

\zheading{Grasp refinement}\label{sec:refinement}
We replicate the benchmark for refining noisy grasps, proposed by Grady et al. \cite{Grady2021ContactOptOC},
which consists of a perturbed version of the ContactPose dataset
\cite{Brahmbhatt2020ContactPoseAD}. ContactPose comprises highly accurate hand
poses for static grasps of 25 objects performed by 50 participants. Grady et al.
\cite{Grady2021ContactOptOC} define large perturbations on the hand poses as 3
additive and i.i.d. noise components: (1) translation noise $\epsilon_t \sim
\mathcal{N}(0, 5)$ in cm, (2) pose noise $\epsilon_\theta \sim \mathcal{N}(0,
0.05)$ in PCA space, and (3) rotation noise $\epsilon_R \sim \mathcal{N}(0, 15)$
in radians. However, they omit a validation split to have more training data. We
instead split the dataset with $70\%$ data for training, $10\%$ for validation
and the last $20\%$ for testing. For training, we use $16$ perturbed versions of
each sample and $4$ for validation and testing. We retrain
ContactOpt\cite{Grady2021ContactOptOC} on these new splits and also train all
methods on object splits to evaluate generalizability in \cref{ap:objects}.
Additionally, we retrain the recent SOTA method TOCH\cite{Zhou2022TOCHSO}, specialized in denoising
dynamic grasps on the GRAB\cite{GRAB:2020} benchmark, which has not been evaluated on static grasps with this amount of noise.

\zheading{Grasp synthesis}\label{sec:synthesis}
In addition to refining noisy interactions, we demonstrate that our \modelname can be used to synthesize novel interactions for unseen objects. To do this, we train \modelname on ContactPose with the object mesh encoded in BPS
representation as input to the context encoder. For quantitative comparison, we
retrain a recent SOTA method in grasp synthesis:
GraspTTA\cite{jiang2021graspTTA}. This method also uses test-time
adaptation, which makes it a good baseline to compare against. We use the same training,
validation and test splits as for \cref{sec:refinement}. This benchmark
evaluates the capabilities of our model to learn the complex interaction
between hands and objects. In grasp refinement, partial noisy information is given to
the model during inference, but in grasp synthesis, the model must generate
plausible grasps without prior information other than the training data. In
\cref{ap:multimodal}, we evaluate our multimodal variant on the OakInk
benchmark\cite{YangCVPR2022OakInk} against GrabNet\cite{GRAB:2020}.

\subsection{Qualitative \& quantitative results}
We use several key metrics to quantify hand pose and contact error, as detailed
in \cref{ap:metrics}. In particular, we employ the (Root-aligned) Mean Per-Joint
Pose Error (R-MPJPE and MPJPE) for either
world space error (MPJPE) or object space error (R-MPJPE). However, a low pose
error is not always indicative of a realistic or well-refined
grasp. 
For denoising, recovering intended contacts is more important: the Precision score
captures intended grasp locations, while a high Recall score implies fewer false negatives.
The latter can be misleadingly increased by maximizing hand contact, leading to
less dexterous grasps (as shown by ContactOpt\cite{Grady2021ContactOptOC} on
\cref{fig:qualitative_cp}). The F1 sore (harmonic mean of Precision and Recall)
is the most meaningful metric for hand contact fidelity.
For synthesis, we employ the simulation displacement (SD) metric with IV to
evaluate the feasibility and stability of grasps.

For grasp refinement, \cref{tab:contactpose} shows that our method \modelname
outperforms two SOTA methods on most metrics, and comes second best in the
remaining metrics. In particular, our method brings a $5\%$ improvement over
TOCH\cite{Zhou2022TOCHSO} and $10\%$ over ContactOpt\cite{Grady2021ContactOptOC}
in contact F1 score. Our method demonstrates the lowest R-MPJPE ($-4.6\text{mm}$
over ContactOpt) and highest contact precision ($+3.8\%$ over TOCH), indicating
a higher contact and grasp fidelity than both methods. With the highest
contact precision, the lowest intersection volume ($-4\text{mm}$ over
TOCH), and a high contact recall, \modelname demonstrates the most accurate
contact inference, as seen with a challenging case on \cref{fig:qualitative_cp}.
This comes at the cost of a negligible penalty in absolute pose, where TOCH
outperforms \modelname by less than $1\text{mm}$. More qualitative comparisons
between  \modelname and ContactOpt are available in \cref{ap:qualitative-cp}.
ContactOpt remains $2\%$ better in hand contact Recall since it aims to
maximize hand-object contact and thus reduces false negatives. However, with a
$12\%$ worse contact Precision than our method, it cannot yield the intended grasp
with high accuracy, as reflected by a $10\%$ lower F1 score and a failure on a
challenging case in \cref{fig:qualitative_cp}.

For grasp synthesis, \cref{fig:synthesis} shows \modelname's ability to
generate plausible and realistic grasps. It shows minimal penetration and
consistent contact between the used fingers and the object, owing to rich
contact modelling through CHOIR. These results are validated quantitatively on
\cref{tab:generation}, where our solution outperforms the SOTA method, GraspTTA
\cite{jiang2021graspTTA}, while being more versatile in applications. \modelname
reduces the intersection volume by $13\%$ and the simulation displacement by
$46\%$, resulting in more stable and feasible grasps. More qualitative results can
be found in \cref{ap:synthesis}.

\begin{table}[bt]
    \centering
    \caption{
    Ablation study of CHOIR on grasp refinement (best seen in colour). Best metrics are in bold and
    second best are underlined, with \textcolor{green}{improvement} or \textcolor{red}{degradation} \wrt the previous row in parenthesis.
    A keypoint diffusion model is used as a baseline (see \cref{ap:kp-baseline}).
    The BPS representation for shape and pose encoding improves all contact metrics at the cost of a slightly higher pose error.
    Adding the probabilistic contacts substantially improves the contact metrics with a small cost in pose accuracy.
    }
    \resizebox{\columnwidth}{!}{
        \begin{tabular}{lccccc}
        \textbf{Method} & \textbf{MPJPE} (mm) $\downarrow$ & \textbf{IV} ($\text{cm}^3$) $\downarrow$ & \textbf{F1} (\%) $\uparrow$ & \textbf{Precision} (\%) $\uparrow$ & \textbf{Recall} (\%) $\uparrow$\\
        \hline  
        KP. Baseline & \textbf{22.11} (-0.00) & 9.18 (-0.00) & 19.45 (+0.00) & 20.22 (+0.00) & 21.18 (+0.00) \\ %
        \textit{+BPS} & \underline{24.69} \textcolor{red}{(+2.58)} &
        \underline{9.12} \textcolor{green}{(-0.06)} & \underline{20.77}
        \textcolor{green}{(+1.32)} & \underline{20.86}
        \textcolor{green}{(+0.64)} & \underline{23.38}
        \textcolor{green}{(+2.20)} \\
        \textit{+BPS +Contacts } & 27.69 \textcolor{red}{(+4.00)} &
        \textbf{6.04} \textcolor{green}{(-3.14)} & \textbf{27.20}
        \textcolor{green}{(+6.43)} & \textbf{25.21} \textcolor{green}{(+4.35)} &
        \textbf{32.80} \textcolor{green}{(+9.42)} \\
        \end{tabular}
    }\label{ablation}
\end{table}

\subsection{Ablation study}

To validate the design choices of CHOIR, we conduct an ablation study of its
components by starting from a keypoint baseline and adding CHOIR components. The
baseline uses a model similar to \modelname to learn hand joint keypoints (see
\cref{ap:kp-baseline}). In \cref{ablation}, \textit{+BPS} corresponds to
\modelname without contact representation, and \textit{+BPS +Contacts}
matches \modelname with the full CHOIR. We evaluate each row on the Perturbed ContactPose
benchmark.
\cref{ablation} shows that while the baseline yields a lower MPJPE,
it also gives the highest IV and lowest contact scores. 
The first two rows yield better pose accuracy by ignoring physical plausibility
and contact fidelity, simplifying the learning objective. The last row improves
on all contact and penetration metrics. Hence, CHOIR offers the best compromise
for pose accuracy and contact fidelity in the denoising setting, while yielding
plausible and stable grasps in the synthesis setting. These findings corroborate
the previous experimental results.

\section{Conclusion}
In this work, we introduced the novel Coarse Hand-Object Interaction
Representation (CHOIR), a versatile and fully-differentiable
hand-object interaction field.  CHOIR represents hand and object
shape and pose as unsigned distance, and dense contacts using probability
distributions, leading to more accurate hand-object interactions. Additionally,
leveraging CHIOR we train a diffusion model, \modelname, to both refine or generate
hand-object interactions. CHOIR demonstrates improvements in 
pose and contact accuracy over existing representations, providing a compact
representation for refining or synthesizing hand-object interaction poses. 

\qheading{Limitations \& Future Work}
Despite its advancements, our method is not without limitations. The reliance on
the BPS representation may limit the ability to capture detailed interactions.
Furthermore, the model's focus on static interactions might restrict its
application in real-world scenarios. To address these limitations and extend the
utility of our framework, future work will focus on: (a) investigating learnable
geometry representations to capture more detailed interactions, (b) handling
dynamic interactions by incorporating temporal information into CHOIR.

\qheading{Acknowledgement}
This work was conducted with the financial support of the Science Foundation Ireland
Centre for Research Training in Digitally-Enhanced Reality (d-real) under Grant No.
18/CRT/6224. For the purpose of Open Access, the author has applied a CC BY public
copyright licence to any Author Accepted Manuscript version arising from this submission

{\small
\bibliographystyle{ieee_fullname}
\bibliography{main}
}

\appendix
\section{Supplementary material}
\section{Method details}
In this section, we include additional information regarding our representation and learning method.
\subsection{CHOIR: Anchor assignment}\label{ap:anchors-ass}
\begin{table}[h]
    \centering
    \caption{Average reconstruction error for MANO meshes fitted onto ground-truth CHOIRs with the \textit{ordered} and \textit{random}
    anchor assignment schemes. Mean Per-Joint Pose Error (MPJPE) and Mean Per-Vertex Pose Error (MPVPE)
    are averaged across the entire ContactPose \cite{Brahmbhatt2020ContactPoseAD} dataset.}
    \label{tab:anchor-assignment}
    \[\begin{array}{c|cc}
     & \textit{Ordered} & \textit{Random} \\
    \hline
    \text{MPJPE (mm)} & 0.18 & 0.19\\
    \text{MPVPE (mm)} & 0.22 & 0.22
    \end{array}\]
\end{table}

\cref{tab:anchor-assignment} shows that both the \textit{ordered} and \textit{random} anchor assignment schemes produce the same reconstruction
error when fitting a ground-truth CHOIR from the ContactPose \cite{Brahmbhatt2020ContactPoseAD} dataset. The Mean Per-Joint Pose Error (MPJPE)
and Mean Per-Vertex Pose Error (MPVPE) metrics were averaged across the entire dataset.
Note that with ground-truth hand-object meshes, the obtained CHOIR allows fitting a MANO mesh with less than $1\text{mm}$ error.

\subsection{Test-Time Optimization: Fitting loss}
The Python code for the stage $1$ of the TTO loss fits in a few lines of code: %
\begin{figure}[h]%
\begin{code}
\label{code:choir_loss}
\vspace{-10pt}
\begin{mintedbox}{python}
anchor_dist = torch.cdist(
    bps, anchors
) # Anchors predicted in TTO
distances = torch.gather(
    anchor_dist, 2, anchor_ids
)
choir_loss = F.mse_loss(
    distances, choir[..., -1]
) # Agreement of anchors and CHOIR
\end{mintedbox}
\vspace{-10pt}
\end{code}
\captionof{algorithm}{Minimal Python code for the stage 1 TTO loss.}
\end{figure}

\subsection{Keypoint baseline}\label{ap:kp-baseline}
To evaluate the expressiveness and efficacy of each component of CHOIR, we
design a diffusion model backbone that allows us to fit a simpler alternative to
CHOIR. This simpler representation only encodes the hand pose and shape as $21$
MANO joints $\boldsymbol{j}_H \in \mathbb{R}^{21\times3}$ and $32$ MANO anchors
$\boldsymbol{a}_H \in \mathbb{R}^{32\times3}$. The object is encoded as a vector
of $K$ randomly sampled surface points $\boldsymbol{p}_O \in
\mathbb{R}^{K\times3}$ where we set $K=4096$ to match CHOIR which uses a grid of
$16\times16\times16$ basis points.
The final keypoint representation is defined as
\begin{equation}
    \boldsymbol{r}_\text{kp} = [
   \boldsymbol{p}_O \in \mathbb{R}^{K\times3}, \boldsymbol{j}_H \in \mathbb{R}^{21\times3},
   \boldsymbol{a}_H \in \mathbb{R}^{32\times3}%
   ].
\end{equation}
However, as in \modelname, this model learns to predict the hand part only, defined as
\begin{equation}
    \boldsymbol{r}_\text{kp}^\text{H} = [
   \boldsymbol{j}_H \in \mathbb{R}^{21\times3},
   \boldsymbol{a}_H \in \mathbb{R}^{32\times3}%
   ]%
\end{equation}

The backbone of this diffusion model is composed only of residual blocks made of multi-layer perceptrons (MLPs).
We use $4$ residual blocks with a hidden dimensionality of $512$.%

In effect, in this baseline, we only replace the 3D U-Net component of
\modelname with a residual MLP and remove the contact prediction branch, while
keeping cross-attention and the same timestep conditioning scheme. The context
encoder is also replaced with a residual MLP of hidden dimensionality $2048$. We
experimented with a PointNet++-based encoder but observed a degradation in
performance. 

\subsection{Runtime costs}\label{ap:runtime}
To evaluate the computational costs of CHOIR, we timed its computation and that
of TOCH\cite{Zhou2022TOCHSO} for 50 grasps on an RTX 2080Ti and Intel i9-7900X.
On average, TOCH takes $\sim 8.89$s ($\pm3.99$) while CHOIR takes $\sim 0.13$s
($\pm0.015$), a $68\times$ reduction. When looking at the total inference time,
including the model representation computation, forward pass and TTO, ours
converges in $\sim 49$s ($\pm$ 16) and TOCH in $\sim 23$s ($\pm$ 4.3). Our
diffusion model accounts for $\sim 13$s of the total ($27\%$), hence is a major
runtime bottleneck. Diffusion Models are inherently slow, but they are becoming
faster, and new alternatives with similar properties can be easily integrated
since our representation is agnostic to the learning method.

\section{Additional experiments and results}
\subsection{Evaluation metrics}\label{ap:metrics}
In our experiments, we use the following metrics to evaluate the fitted hand mesh to the predicted CHOIR:
\begin{itemize}
    \item \textbf{Mean Per-Joint Pose Error (MPJPE)/(R-MPJPE) (mm)}: L2 norm between ground-truth (GT) and predicted hand joints. We compute both absolute (MPJPE) and root-aligned (R-MPJPE) metrics.
    The former tells us about the position of the hand around the object, and the latter tells us about the hand grasp error regardless of the spatial pose.
    \item \textbf{Intersection Volume (IV) ($\text{cm}^3$)}: A measure of hand-object mesh penetration. It is computed by voxelizing the hand and object meshes ($1\text{mm}$ voxels) and computing the volume of the intersecting voxels.
    \item \textbf{Hand contact F1/precision/recall (\%)}: The precision and recall scores are measured on binary hand contact maps obtained by upsampling the MANO mesh and computing the Chamfer distance to the object point cloud. Hand vertices within $2\text{mm}$ of their nearest object point are considered in contact, to emulate soft tissue deformation as in \cite{Grady2021ContactOptOC}.
    A high precision means a low false positives count, while a high recall means a low false negatives count. The F1 score
    is the harmonic mean of both and is a measure of predictive performance.
    \item \textbf{Simulation Displacement (SD) ($\text{cm}$)}: The distance of displacement of the object in world space when
    applying inward forces to the hand grasp in a physics simulation. This tells how stable the grasp is, since more hand-object contact patches result in higher friction and therefore lower displacement.
\end{itemize}

\subsection{Perturbed ContactPose}\label{ap:qualitative-cp}
We show a qualitative comparison of our method \vs ContactOpt \cite{Grady2021ContactOptOC} on several objects. \cref{fig:failure} shows
failure cases in some challenging cases. While ContactOpt \cite{Grady2021ContactOptOC} fails to produce a plausible
grasp for each object and noisy input, our method
delivers satisfying results that still closely match the contacts of the ground-truth hand pose.
Further qualitative samples are shown in \cref{fig:good1}, \cref{fig:good2}, and \cref{fig:good3}, where our method demonstrates
fidelity in the reconstructed finger contacts, as opposed to ContactOpt \cite{Grady2021ContactOptOC}.

\noindent

\begin{table*}[h]
    \centering
    \begin{tblr}{
      colspec = {c|X[c,j]X[c,j]|X[c,j]},
      stretch = 0,
      rowsep = 6pt,
    }
      Method & Ground truth & Observation & Prediction \\
     \hline
      \modelname &
      \includegraphics[scale=0.15]{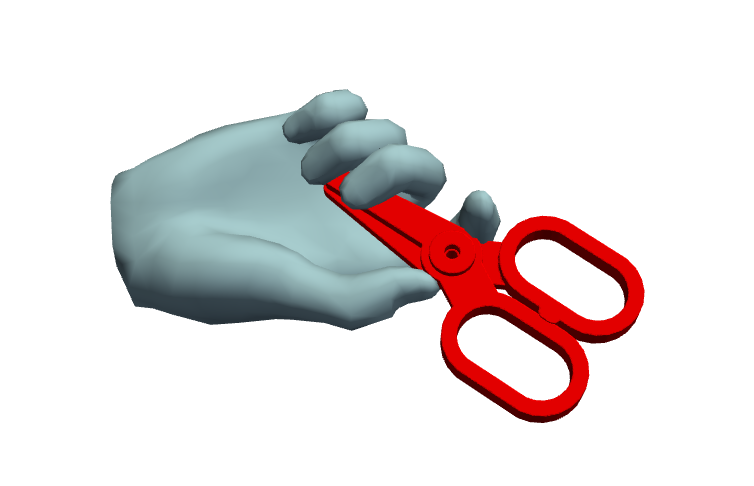} & 
      \includegraphics[scale=0.15]{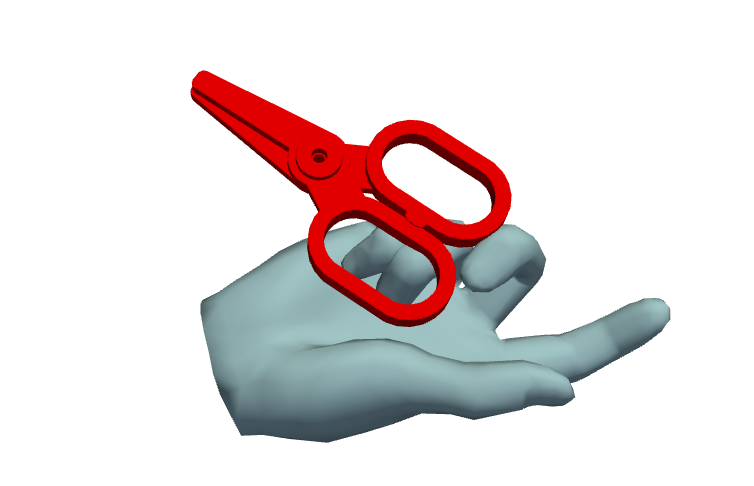} &
      \includegraphics[scale=0.15]{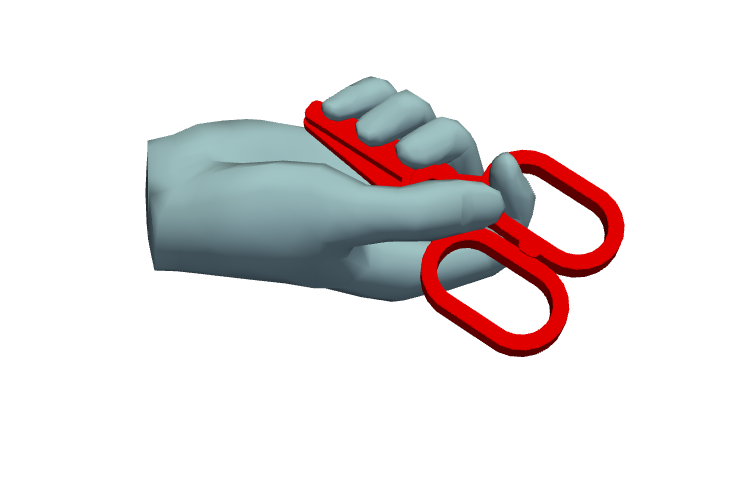} \\ 
      ContactOpt & 
          \includegraphics[scale=0.15]{figures/contactpose/denoising/scissors/Screenshot_2024-03-14_at_15.02.57.png} & 
            \includegraphics[scale=0.15]{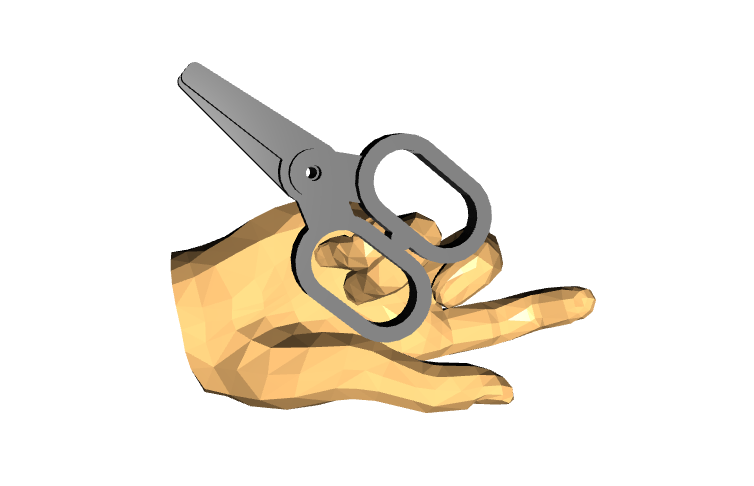}&
            \includegraphics[scale=0.12]{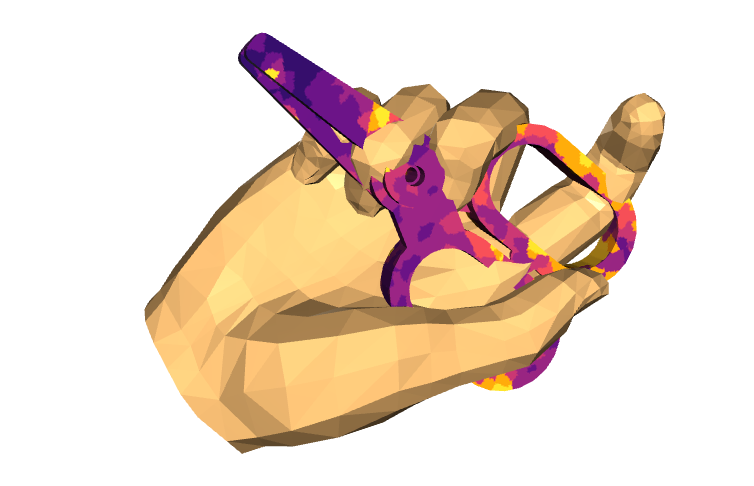} \\
        \hline
      \modelname &
      \includegraphics[scale=0.15]{figures/contactpose/denoising/wine_glass/Screenshot_2024-03-14_at_15.05.53.png} & 
      \includegraphics[scale=0.15]{figures/contactpose/denoising/wine_glass/Screenshot_2024-03-14_at_15.05.59.png} & 
      \includegraphics[scale=0.15]{figures/contactpose/denoising/wine_glass/Screenshot_2024-03-14_at_15.06.02.png} \\ 
      ContactOpt & 
      \includegraphics[scale=0.15]{figures/contactpose/denoising/wine_glass/Screenshot_2024-03-14_at_15.05.53.png} & 
      \includegraphics[scale=0.15]{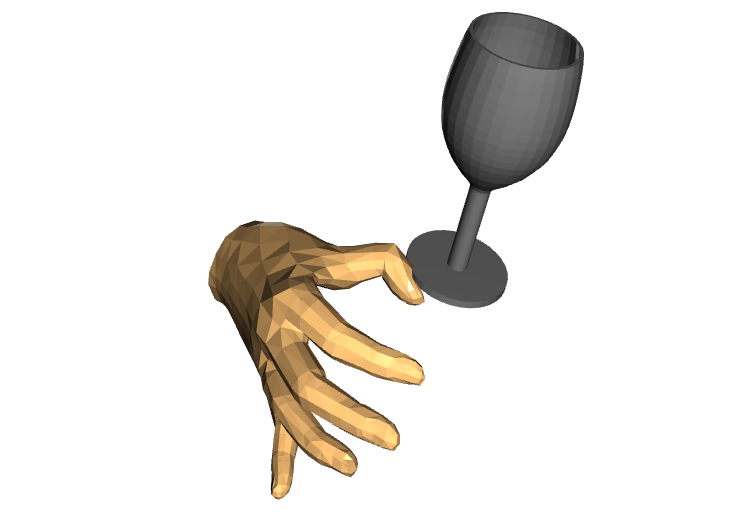} & 
      \includegraphics[scale=0.15]{figures/contactpose/denoising/wine_glass/Screenshot_2024-03-14_at_15.07.22.png} \\ 
      \hline
      \modelname &
      \includegraphics[scale=0.15]{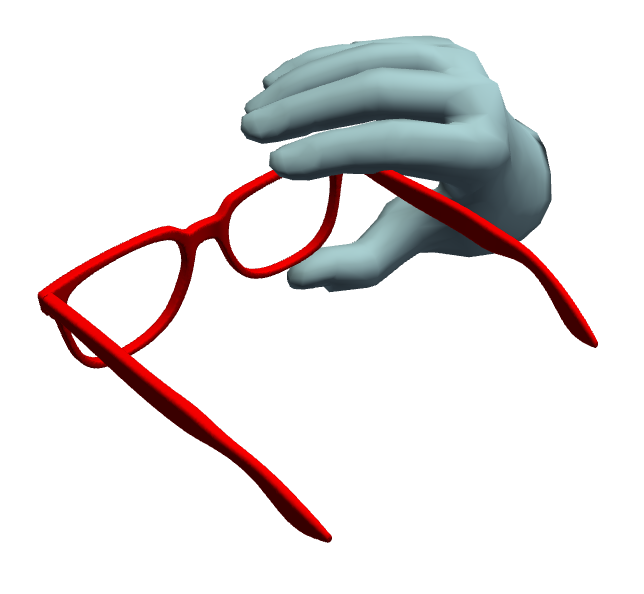} & 
      \includegraphics[scale=0.15]{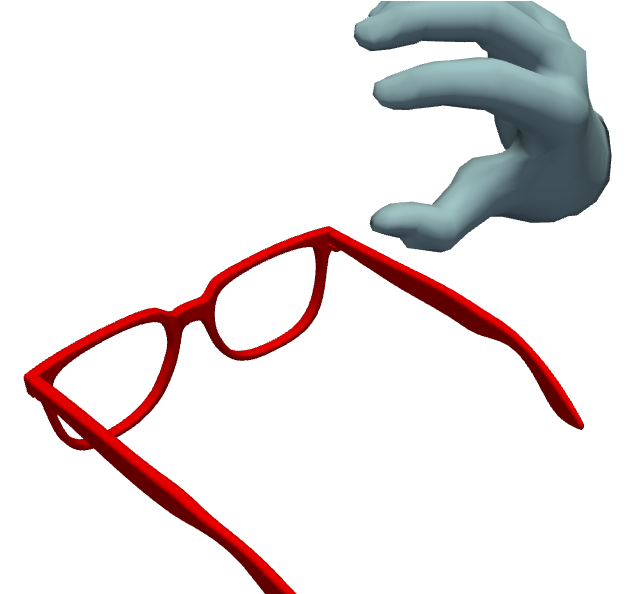} & 
      \includegraphics[scale=0.15]{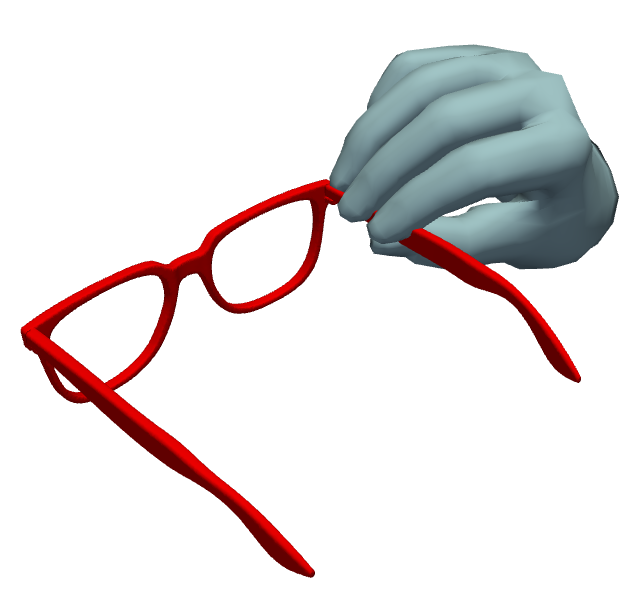} \\ 
      ContactOpt & 
      \includegraphics[scale=0.15]{figures/contactpose/denoising/glasses/Screenshot_2024-03-14_at_14.53.03.png} & 
      \includegraphics[scale=0.15]{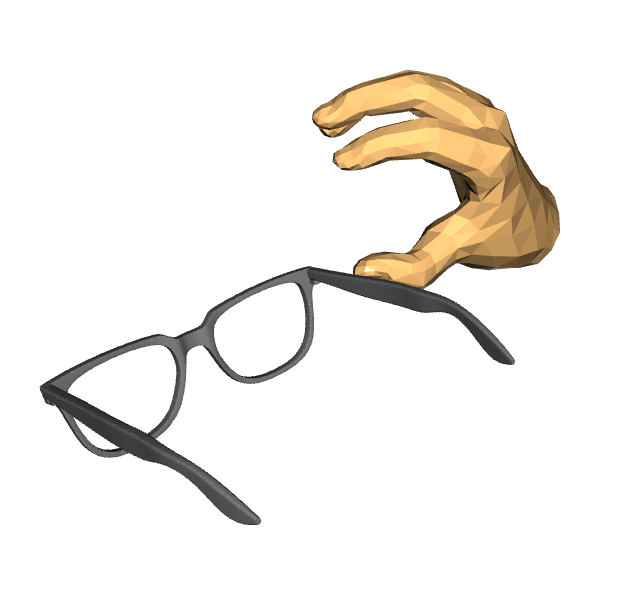} & 
      \includegraphics[scale=0.15]{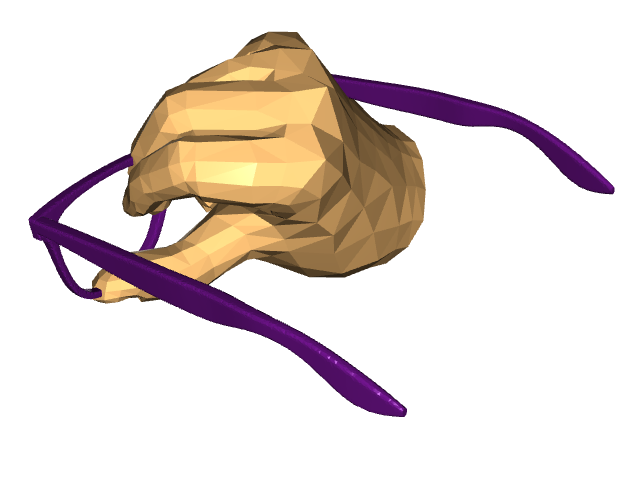} \\
    \end{tblr}
    \captionsetup{type=figure}
     \captionof{figure}{Failure cases on a comparison of \modelname and ContactOpt for the Perturbed ContactPose benchmark.
     While ContactOpt consistently fails at producing a plausible mesh after multiple restarts,
     our method results in minimal penetration and respected finger contacts with only one sample.}
    \label{fig:failure}
 \end{table*}

\bigskip

\begin{table*}[h]
    \centering
    \begin{tblr}{
      colspec = {c|X[c,j]X[c,j]|X[c,j]},
      stretch = 0,
      rowsep = 6pt,
    }
      Method & Ground truth & Observation & Prediction \\
     \hline
      \modelname &
      \includegraphics[scale=0.16]{figures/contactpose/denoising/scissors2/Screenshot_2024-03-14_at_15.09.26.png} & 
      \includegraphics[scale=0.16]{figures/contactpose/denoising/scissors2/Screenshot_2024-03-14_at_15.09.39.png} &
      \includegraphics[scale=0.16]{figures/contactpose/denoising/scissors2/Screenshot_2024-03-14_at_15.09.43.png} \\ 
      ContactOpt & 
            \includegraphics[scale=0.16]{figures/contactpose/denoising/scissors2/Screenshot_2024-03-14_at_15.09.26.png} & 
            \includegraphics[scale=0.16]{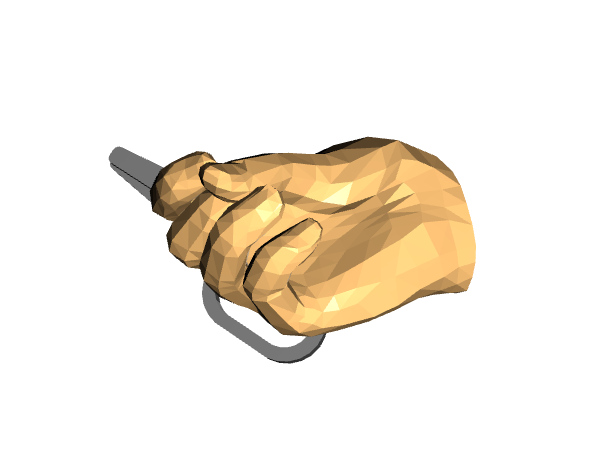}&
            \includegraphics[scale=0.16]{figures/contactpose/denoising/scissors2/Screenshot_2024-03-14_at_15.10.26.png} \\
        \hline
      \modelname &
      \includegraphics[scale=0.16]{figures/contactpose/denoising/bowl/Screenshot_2024-03-14_at_14.30.10.png} & 
      \includegraphics[scale=0.16]{figures/contactpose/denoising/bowl/Screenshot_2024-03-14_at_14.30.17.png} & 
      \includegraphics[scale=0.16]{figures/contactpose/denoising/bowl/Screenshot_2024-03-14_at_14.30.22.png} \\ 
      ContactOpt & 
      \includegraphics[scale=0.16]{figures/contactpose/denoising/bowl/Screenshot_2024-03-14_at_14.30.10.png} & 
      \includegraphics[scale=0.16]{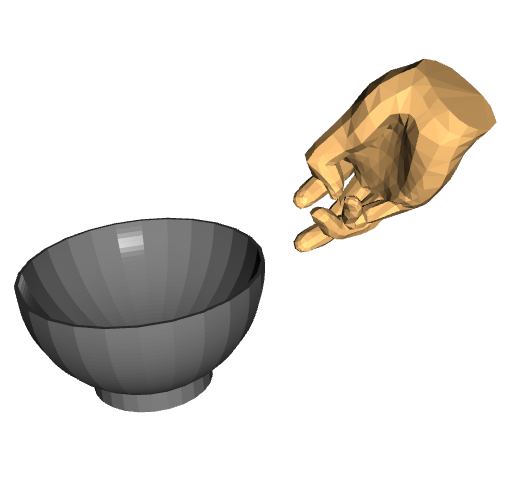} & 
      \includegraphics[scale=0.16]{figures/contactpose/denoising/bowl/Screenshot_2024-03-14_at_14.31.08.png} \\ 
        \hline
      \modelname &
      \includegraphics[scale=0.16]{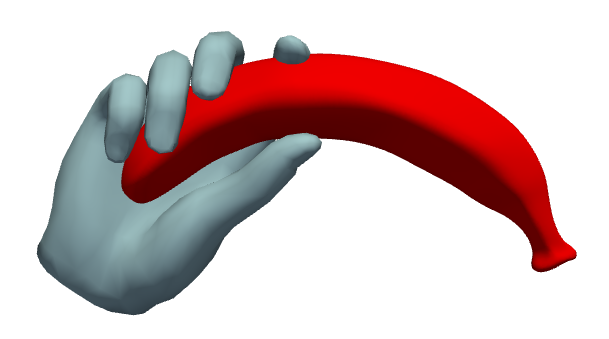} & 
      \includegraphics[scale=0.16]{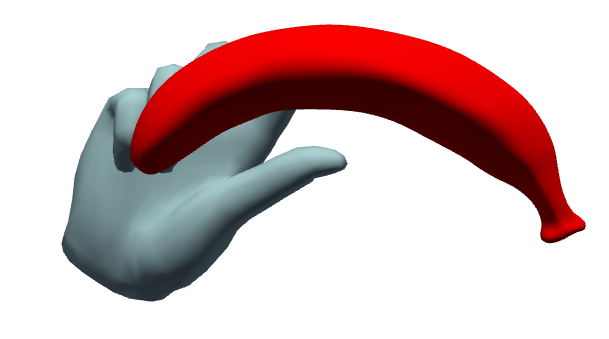} & 
      \includegraphics[scale=0.16]{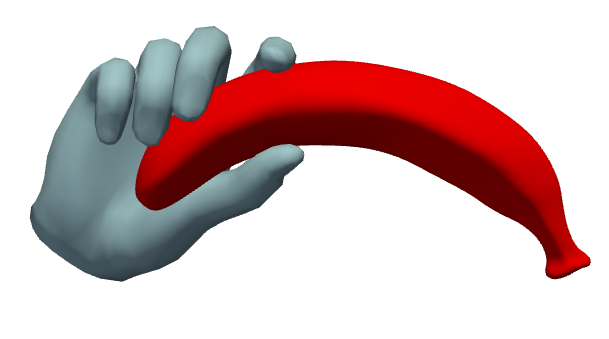} \\ 
      ContactOpt & 
      \includegraphics[scale=0.16]{figures/contactpose/denoising/Banana/Screenshot_2024-03-14_at_14.17.42.png} & 
      \includegraphics[scale=0.16]{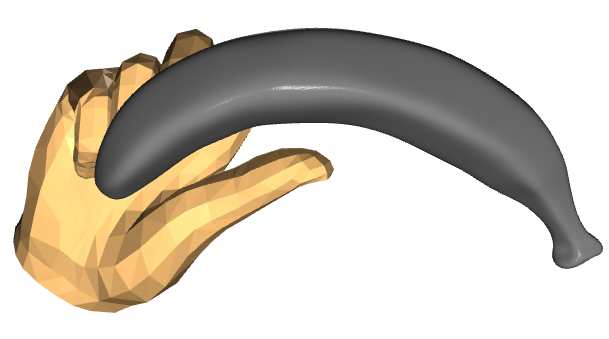} & 
      \includegraphics[scale=0.16]{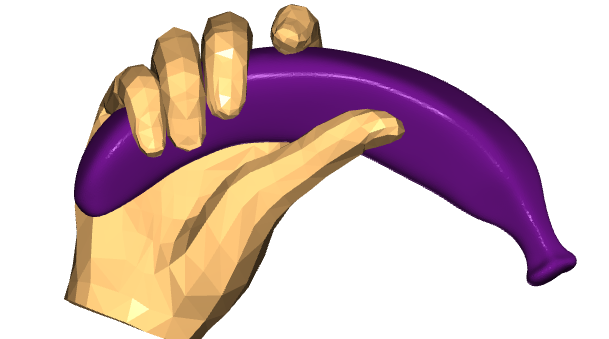} \\ 
    \end{tblr}
    \captionsetup{type=figure}
     \captionof{figure}{Qualitative comparison of \modelname \vs ContactOpt on the Perturbed ContactPose benchmark.
     Our method, \modelname, produces plausible grasps and maintains the fidelity of finger contacts while ContactOpt
     fails in challenging cases even with several random restarts. Our method only draws one sample and performs TTO without
     random restarts.}
    \label{fig:good1}
 \end{table*}

\bigskip

\begin{table*}[h]
    \centering
    \begin{tblr}{
      colspec = {c|X[c,j]X[c,j]|X[c,j]},
      stretch = 0,
      rowsep = 6pt,
    }
      Method & Ground truth & Observation & Prediction \\
     \hline
      \modelname &
      \includegraphics[scale=0.15]{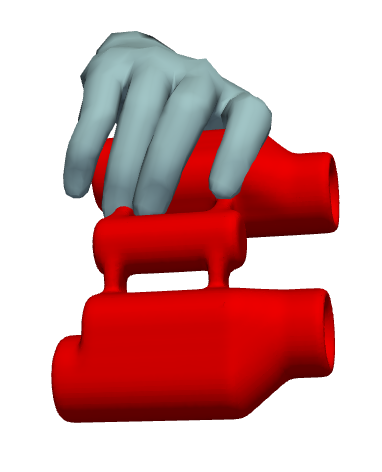} & 
      \includegraphics[scale=0.15]{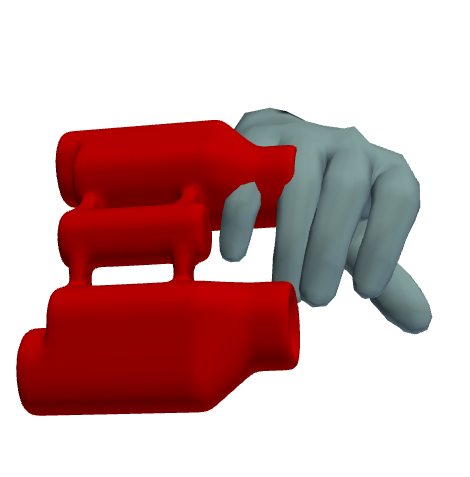} &
      \includegraphics[scale=0.15]{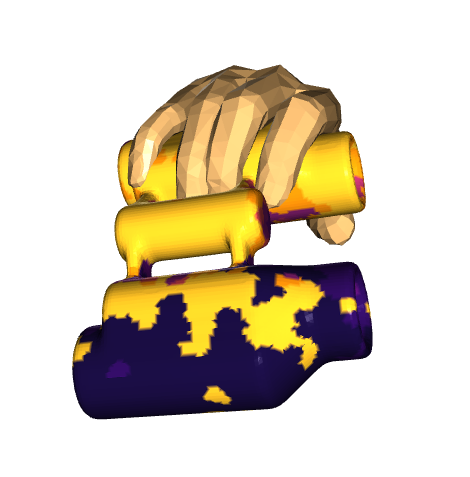} \\ 
      ContactOpt & 
            \includegraphics[scale=0.15]{figures/contactpose/denoising/binoc/Screenshot_2024-03-14_at_14.23.00.png} & 
            \includegraphics[scale=0.15]{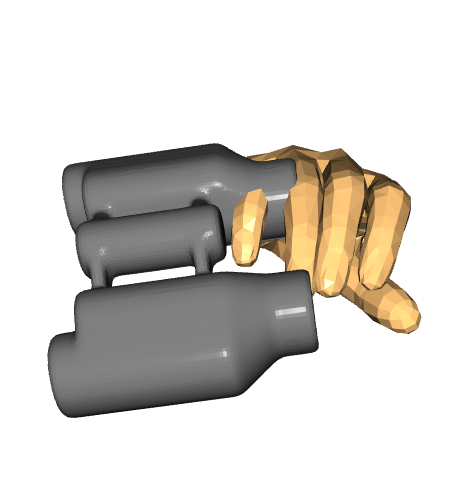}&
            \includegraphics[scale=0.15]{figures/contactpose/denoising/binoc/Screenshot_2024-03-14_at_14.23.22.png} \\
        \hline
      \modelname &
      \includegraphics[scale=0.14]{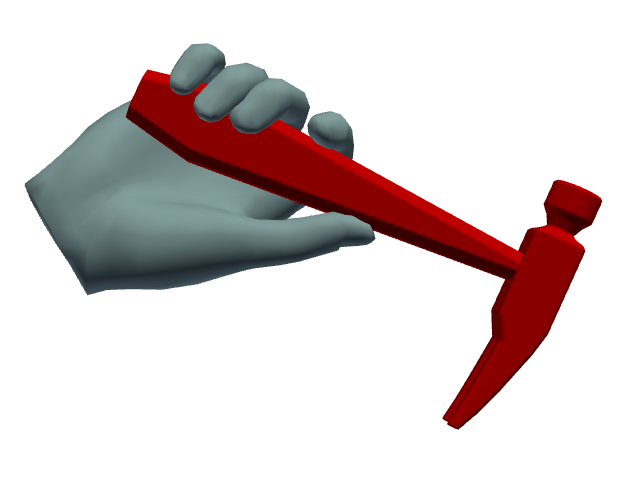} & 
      \includegraphics[scale=0.14]{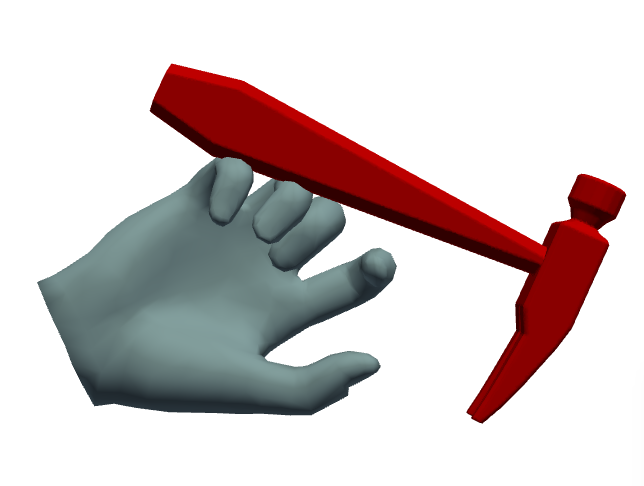} & 
      \includegraphics[scale=0.14]{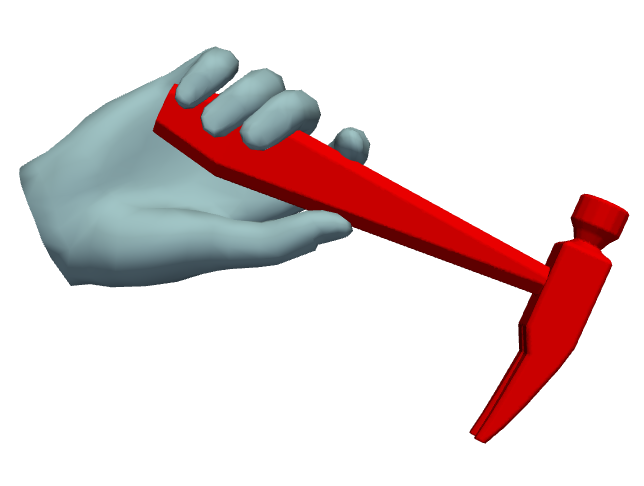} \\ 
      ContactOpt & 
      \includegraphics[scale=0.14]{figures/contactpose/denoising/hammer/Screenshot_2024-03-14_at_14.56.39.png} & 
      \includegraphics[scale=0.14]{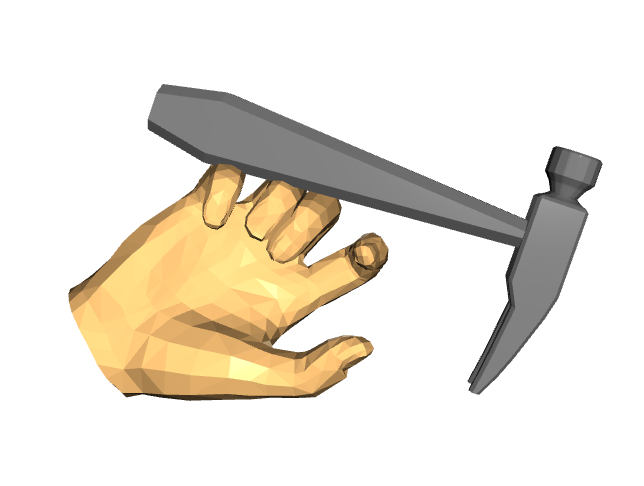} & 
      \includegraphics[scale=0.14]{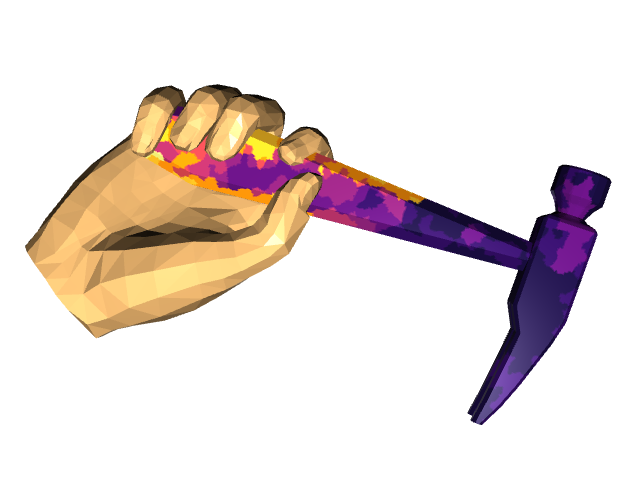} \\ 
        \hline
      \modelname &
      \includegraphics[scale=0.15]{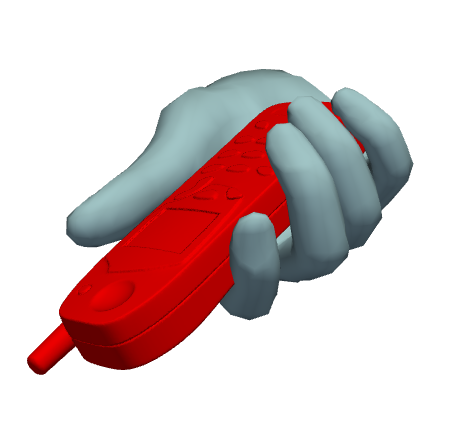} & 
      \includegraphics[scale=0.15]{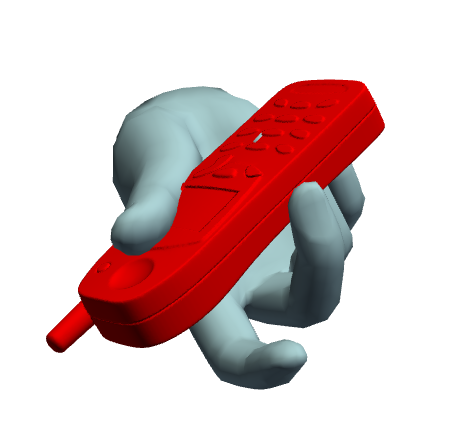} & 
      \includegraphics[scale=0.15]{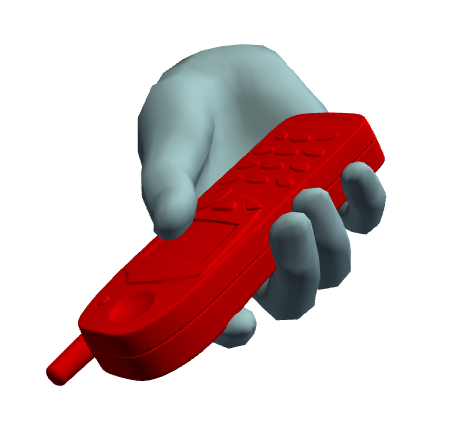} \\ 
      ContactOpt & 
      \includegraphics[scale=0.15]{figures/contactpose/denoising/cell_phone/Screenshot_2024-03-14_at_14.38.34.png} & 
      \includegraphics[scale=0.15]{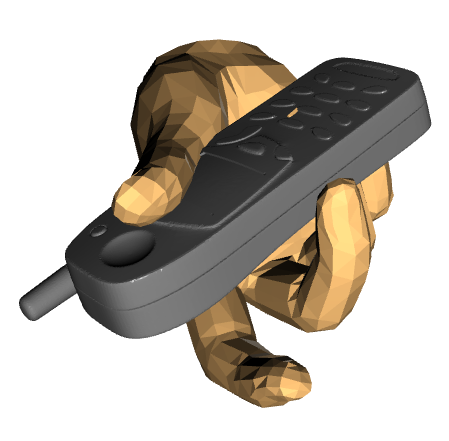} & 
      \includegraphics[scale=0.15]{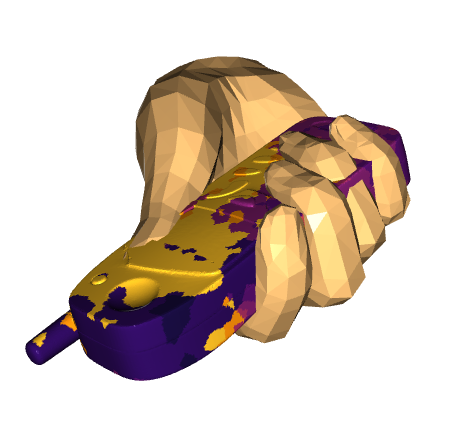} \\ 
    \end{tblr}
    \captionsetup{type=figure}
     \captionof{figure}{Qualitative comparison of \modelname \vs ContactOpt on the Perturbed ContactPose benchmark.
     Our method, \modelname, produces plausible grasps and maintains the fidelity of finger contacts while ContactOpt
     fails in challenging cases even with several random restarts. Our method only draws one sample and performs TTO without
     random restarts.}
    \label{fig:good2}
 \end{table*}

\bigskip

\begin{table*}[h]
    \centering
    \begin{tblr}{
      colspec = {c|X[c,j]X[c,j]|X[c,j]},
      stretch = 0,
      rowsep = 6pt,
    }
      Method & Ground truth & Observation & Prediction \\
     \hline
      \modelname &
      \includegraphics[scale=0.15]{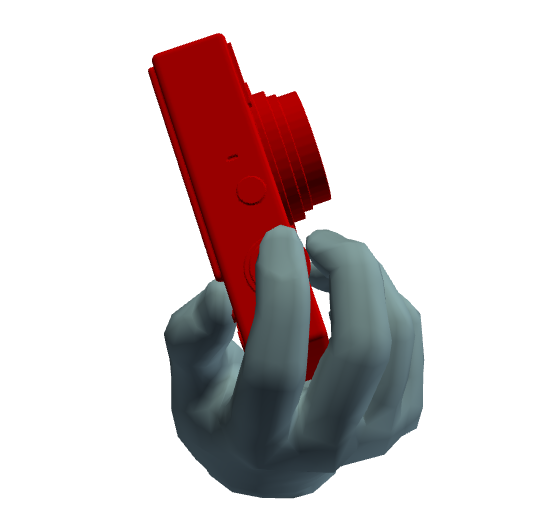} & 
      \includegraphics[scale=0.15]{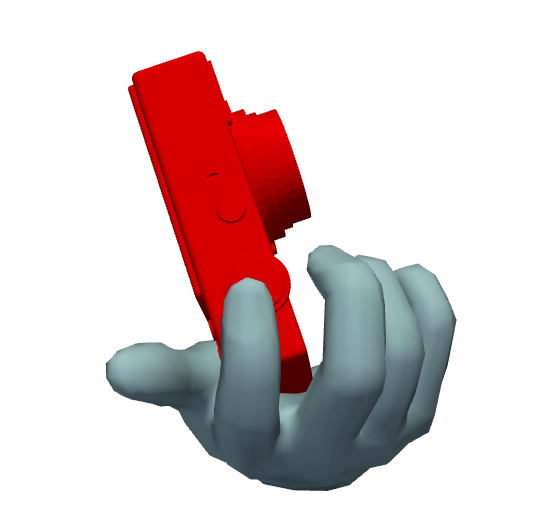} &
      \includegraphics[scale=0.15]{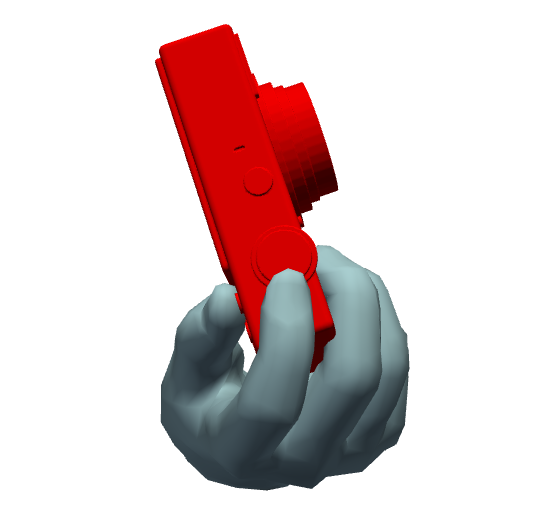} \\ 
      ContactOpt & 
            \includegraphics[scale=0.15]{figures/contactpose/denoising/camera/Screenshot_2024-03-14_at_14.34.36.png} & 
            \includegraphics[scale=0.15]{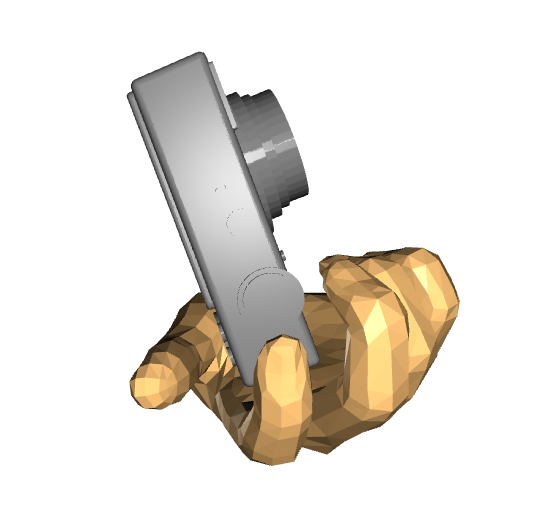}&
            \includegraphics[scale=0.15]{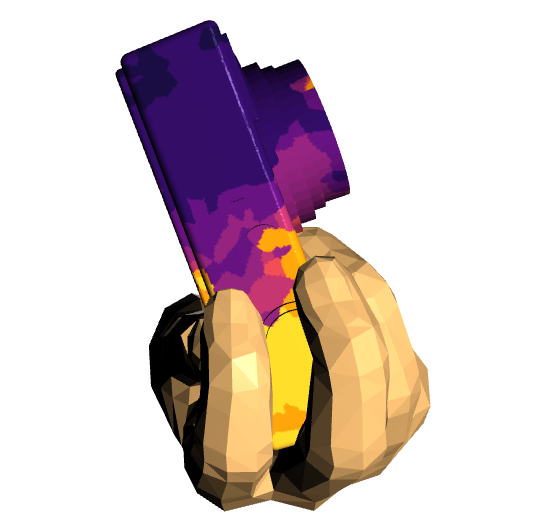} \\
        \hline
      \modelname &
      \includegraphics[scale=0.15]{figures/contactpose/denoising/cup/Screenshot_2024-03-14_at_14.43.45.png} & 
      \includegraphics[scale=0.15]{figures/contactpose/denoising/cup/Screenshot_2024-03-14_at_14.43.52.png} & 
      \includegraphics[scale=0.15]{figures/contactpose/denoising/cup/Screenshot_2024-03-14_at_14.43.56.png} \\ 
      ContactOpt & 
      \includegraphics[scale=0.15]{figures/contactpose/denoising/cup/Screenshot_2024-03-14_at_14.43.45.png} & 
      \includegraphics[scale=0.15]{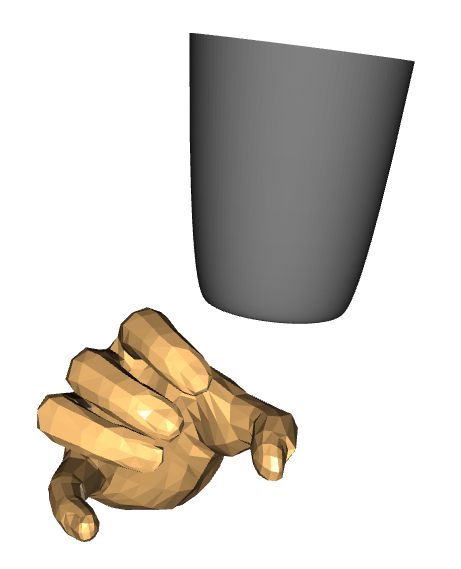} & 
      \includegraphics[scale=0.15]{figures/contactpose/denoising/cup/Screenshot_2024-03-14_at_14.44.15.png} \\ 
        \hline
      \modelname &
      \includegraphics[scale=0.15]{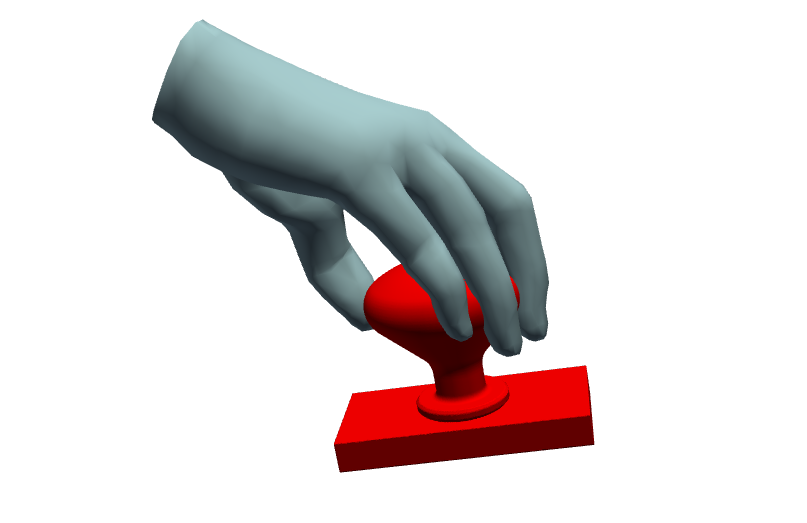} & 
      \includegraphics[scale=0.14]{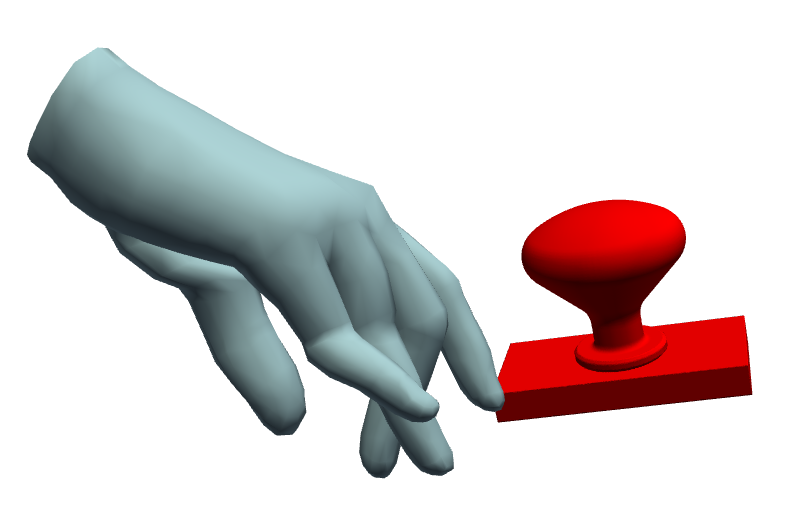} & 
      \includegraphics[scale=0.15]{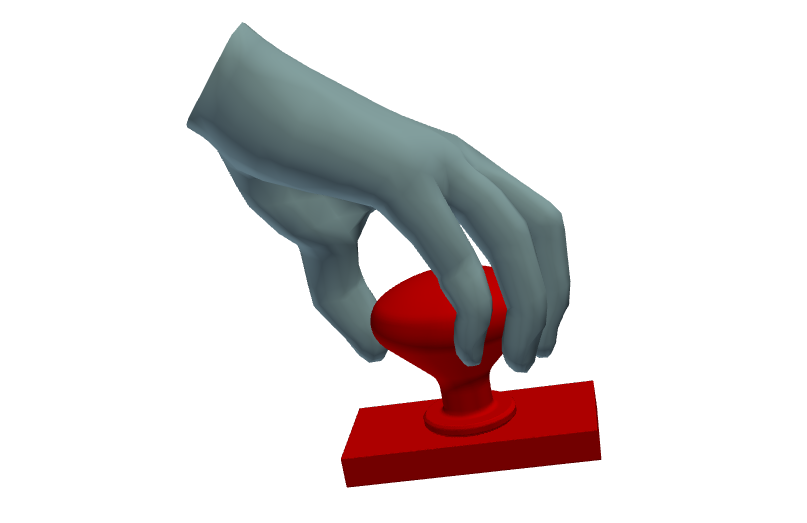} \\ 
      ContactOpt & 
      \includegraphics[scale=0.15]{figures/contactpose/denoising/door_knob/Screenshot_2024-03-14_at_14.47.52.png} & 
      \includegraphics[scale=0.14]{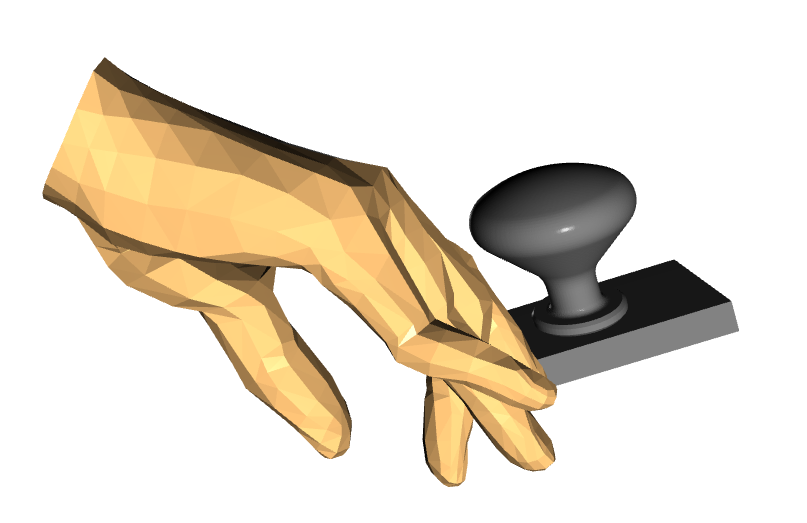} & 
      \includegraphics[scale=0.15]{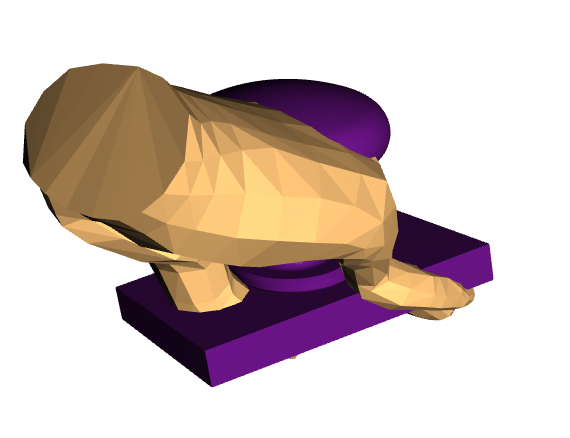} \\ 
    \end{tblr}
    \captionsetup{type=figure}
     \captionof{figure}{Qualitative comparison of \modelname \vs ContactOpt on the Perturbed ContactPose benchmark.
     Our method, \modelname, produces plausible grasps and maintains the fidelity of finger contacts while ContactOpt
     fails in challenging cases even with several random restarts. Our method only draws one sample and performs TTO without
     random restarts.}
    \label{fig:good3}
 \end{table*}

\bigskip

\clearpage

\begin{table*}[h!]
    \centering
    \caption{
        Quantitative evaluation of our approach on static grasp refinement against ContactOpt \cite{Grady2021ContactOptOC} 
        on the Perturbed ContactPose benchmark with object splits. * means reported figures.
        \modelname is evaluated with one non-cherry-picked generated grasp per sample.
        \modelname shows greater contact accuracy and outperforms ContactOpt\cite{Grady2021ContactOptOC} on most contact metrics, although
        ContactOpt\cite{Grady2021ContactOptOC} retains a greater recall score due to its objective which maximizes the hand-object contact ratio,
        hence reducing false negatives. Best results are in bold, second best are underlined.
    }
    \resizebox{\textwidth}{!}{
        \begin{tabular}{lcccccc}
        \textbf{Method} & \textbf{MPJPE} (mm) $\downarrow$ & \textbf{R-MPJPE} (mm) $\downarrow$ & \textbf{IV} ($\text{cm}^3$) $\downarrow$ & \textbf{F1} (\%) $\uparrow$ & \textbf{Precision} (\%) $\uparrow$ & \textbf{Recall} (\%) $\uparrow$\\
        \hline  
        Perturbed data & 83.02 & 21.55 & 6.99 & 1.55 & 1.88 & 2.74\\
        ContactOpt \cite{Grady2021ContactOptOC} & \textbf{35.05} & \textbf{29.13} &
        \underline{12.83}* & \underline{15.39} & \underline{12.04} & \textbf{30.36}\\
        TOCH \cite{Zhou2022TOCHSO} & 48.27 & 51.13 & 17.63 & 11.18 & 10.74 & 13.54 \\
        \modelname & 42.54 & 29.55 & \textbf{2.90} & \textbf{21.40} & \textbf{21.94} & \underline{23.05} \\
        \end{tabular}
    }
    \label{tab:contactpose_objects}
\end{table*}

\subsection{Object splits experiment}\label{ap:objects}
To evaluate the generalizability of our method in the grasp refinement setting, we
retrain all methods on the Perturbed ContactPose benchmark \cite{Grady2021ContactOptOC}
with object splits instead of subject splits. We hold $2$ objects out of the validation
split, and reserve $5$ objects for the test split, namely: \textit{doorknob, eyeglasses,
apple, bowl, toothbrush}. This increases the difficulty of the benchmark, as all test
objects were unseen during training. For a method to perform well in this setting, it
must learn generalizable hand-object interaction in latent space.
\cref{tab:contactpose_objects} shows that our method outperforms ContactOpt
\cite{Grady2021ContactOptOC} on most contact-based metrics, and TOCH
\cite{Zhou2022TOCHSO} on all metrics. ContactOpt \cite{Grady2021ContactOptOC} retains an
edge on the recall score since it maximizes the hand-object contact ratio and therefore
minimizes false negatives, but at the cost of less contact fidelity since its precision
score is significantly lower than \modelname. However, TOCH \cite{Zhou2022TOCHSO} fails
to generalize to these objects, which can be explained by the lack of object
representation in the TOCH field. We consider this task to be a main challenge in
hand-object interaction understanding and will focus on object generalization in future
work.

\subsection{Grasp synthesis}\label{ap:synthesis}
\cref{fig:gen1} and \cref{fig:gen2} show samples of our generative model given an object mesh as input.
The model is trained on the improved Perturbed ContactPose benchmark \cite{Grady2021ContactOptOC},
\ie all objects are seen during training. \modelname generates visually plausible grasps with consistent finger contacts
and minimal mesh penetration. 
In addition, to enhance visibility, we provide non-cherry-picked supplementary videos of generated hand grasps.

\noindent

\begin{table*}[h]
    \centering
    \begin{tblr}{
      colspec = {X[c,j]|X[c,j]X[c,j]X[c,j]},
      stretch = 0,
      rowsep = 6pt,
    }
      Input & Sample 1 & Sample 2 & Sample 3 \\
     \hline
      \includegraphics[scale=0.12]{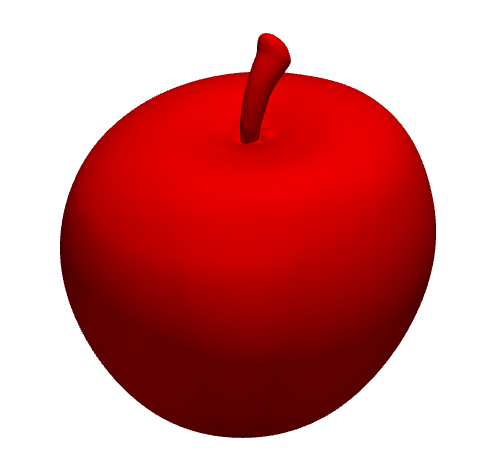} & 
      \includegraphics[scale=0.15]{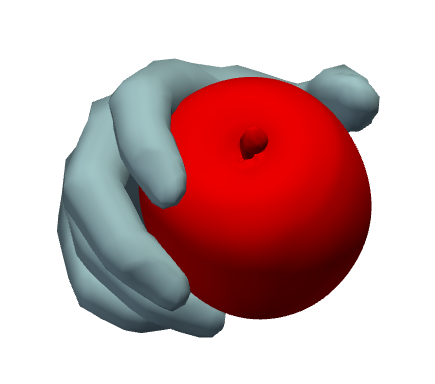} &
      \includegraphics[scale=0.08]{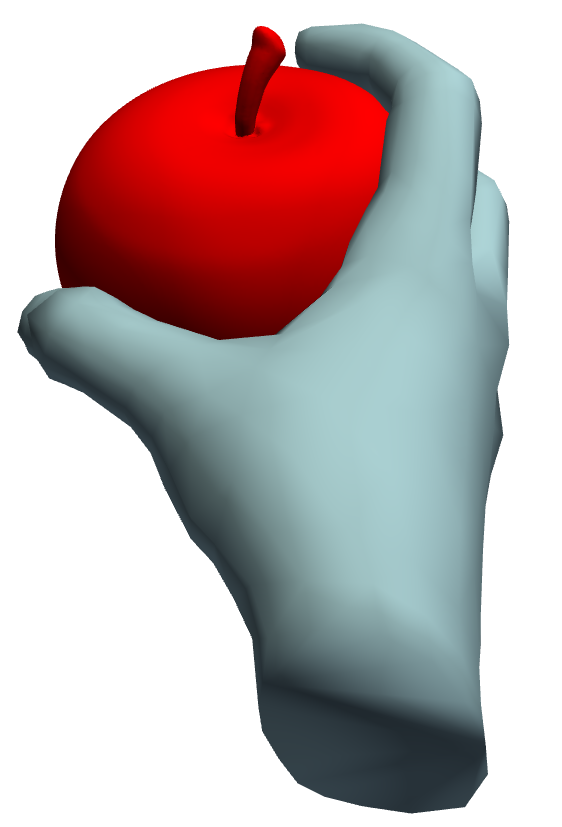} &
      \includegraphics[scale=0.13]{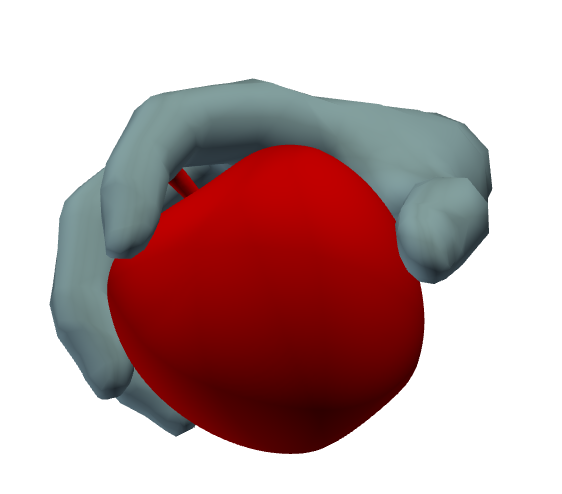} \\ 
      \includegraphics[scale=0.1]{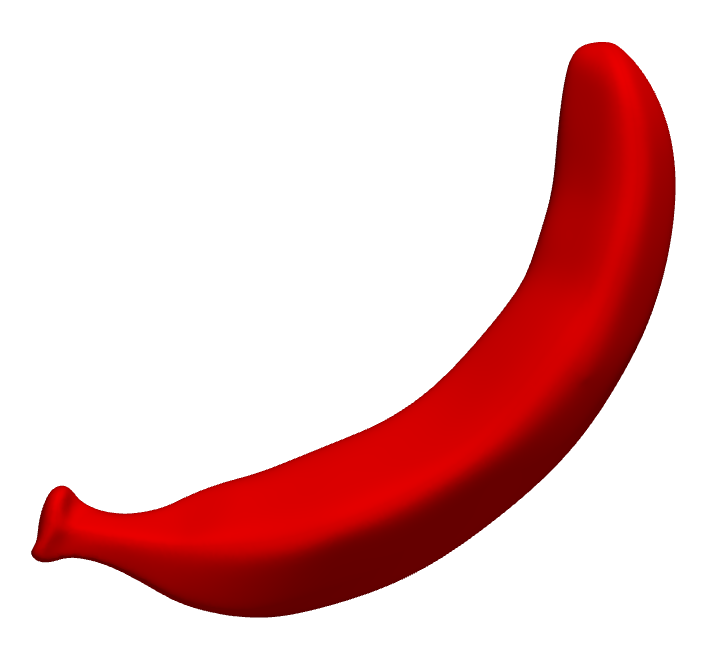} & 
      \includegraphics[scale=0.1]{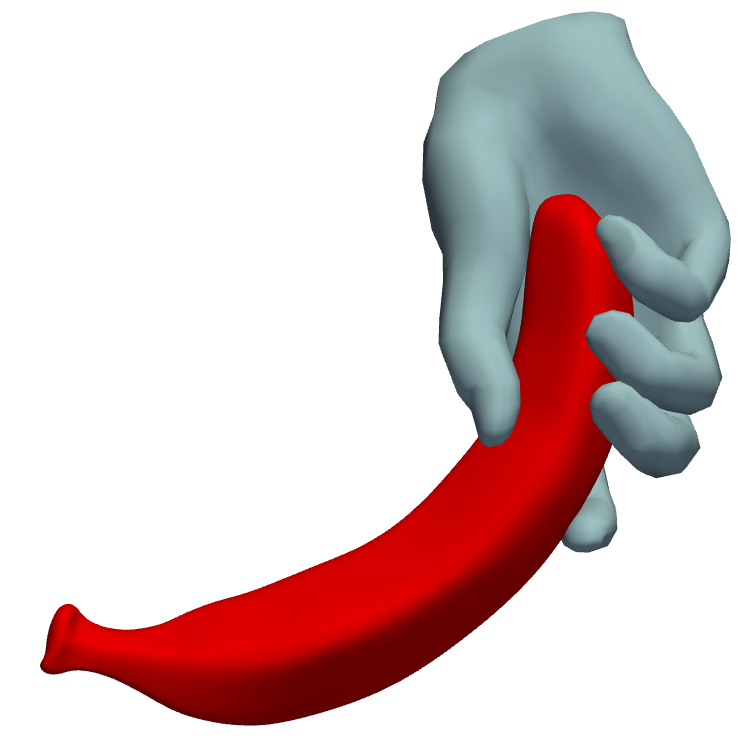} & 
      \includegraphics[scale=0.11]{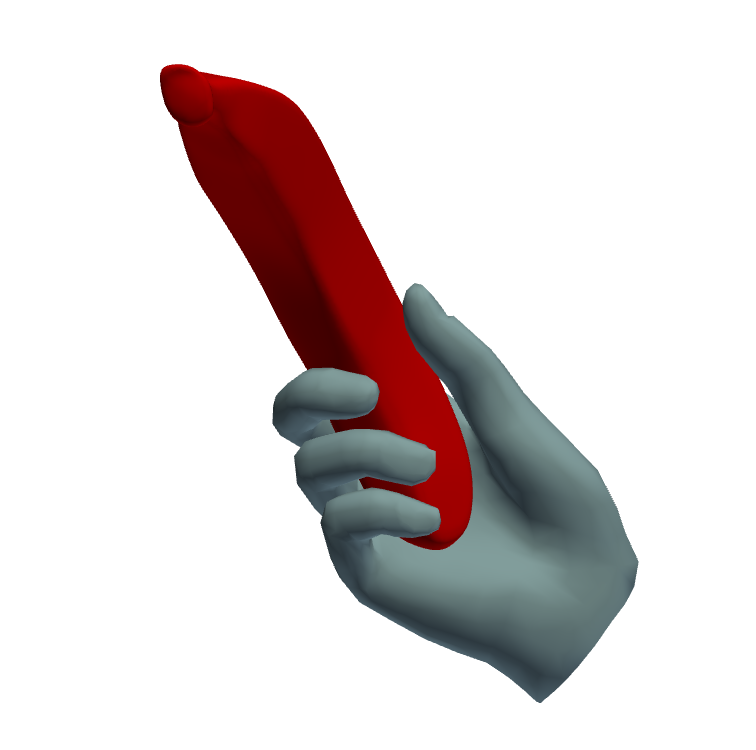} & 
      \includegraphics[scale=0.1]{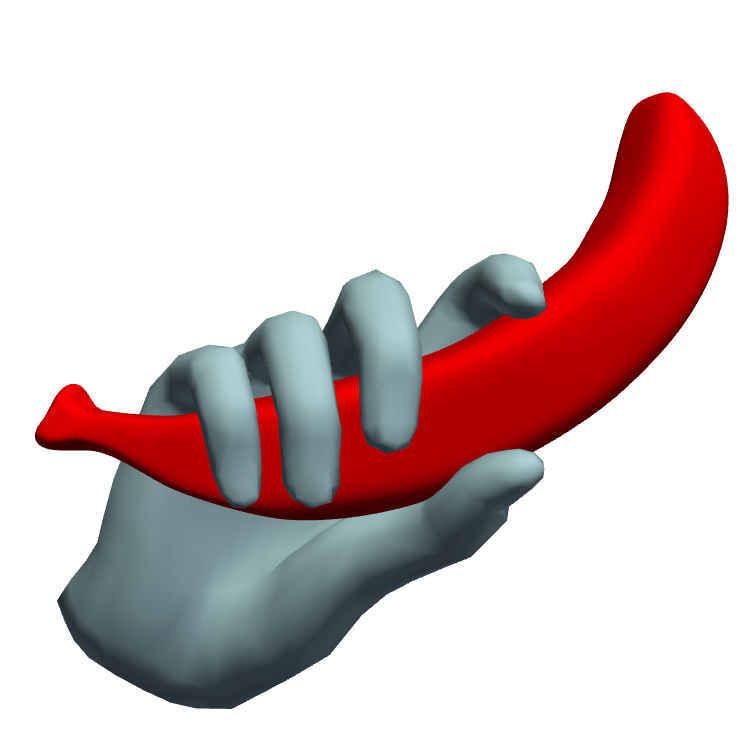} \\ 
      \includegraphics[scale=0.1]{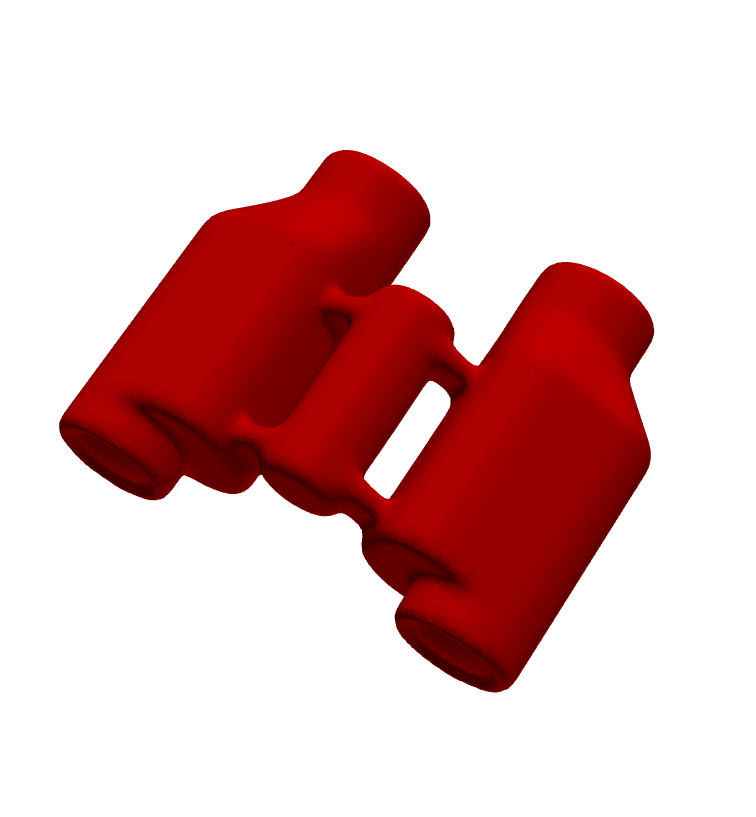} & 
      \includegraphics[scale=0.1]{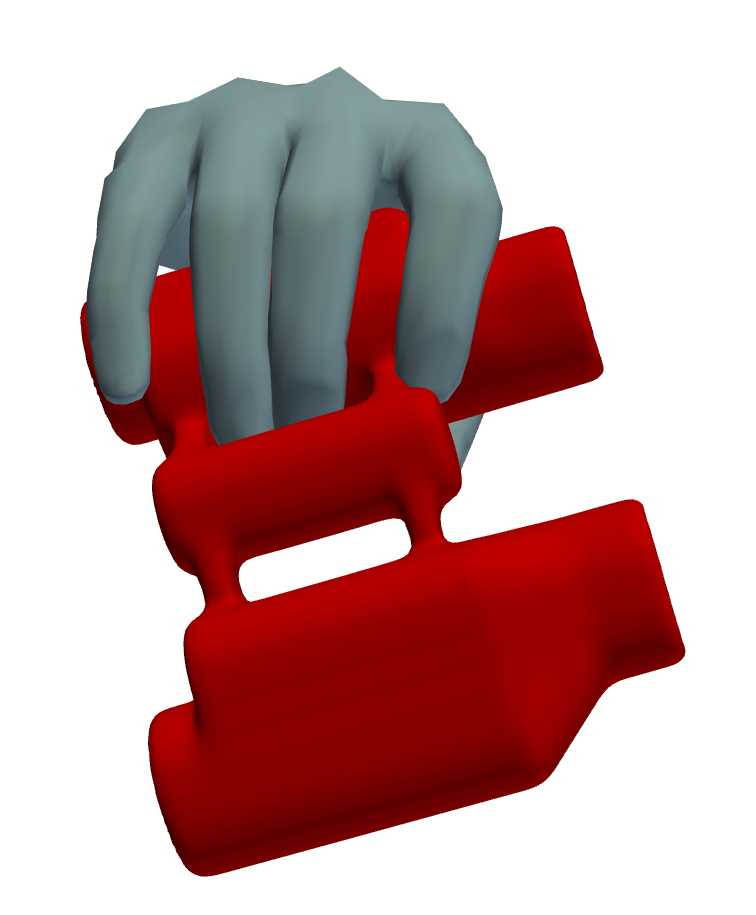} & 
      \includegraphics[scale=0.1]{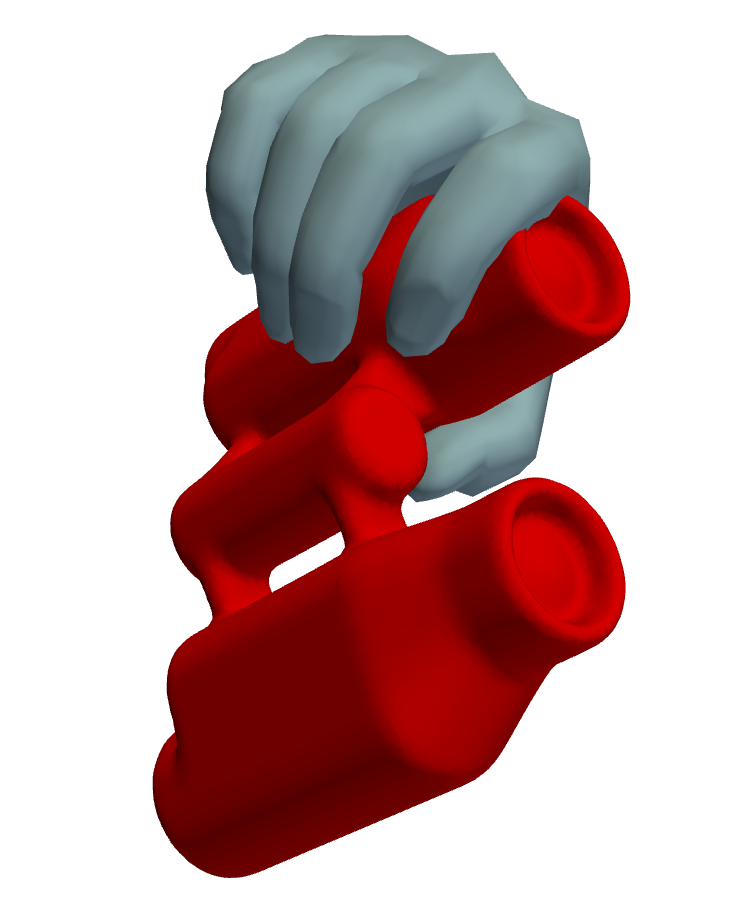} & 
      \includegraphics[scale=0.1]{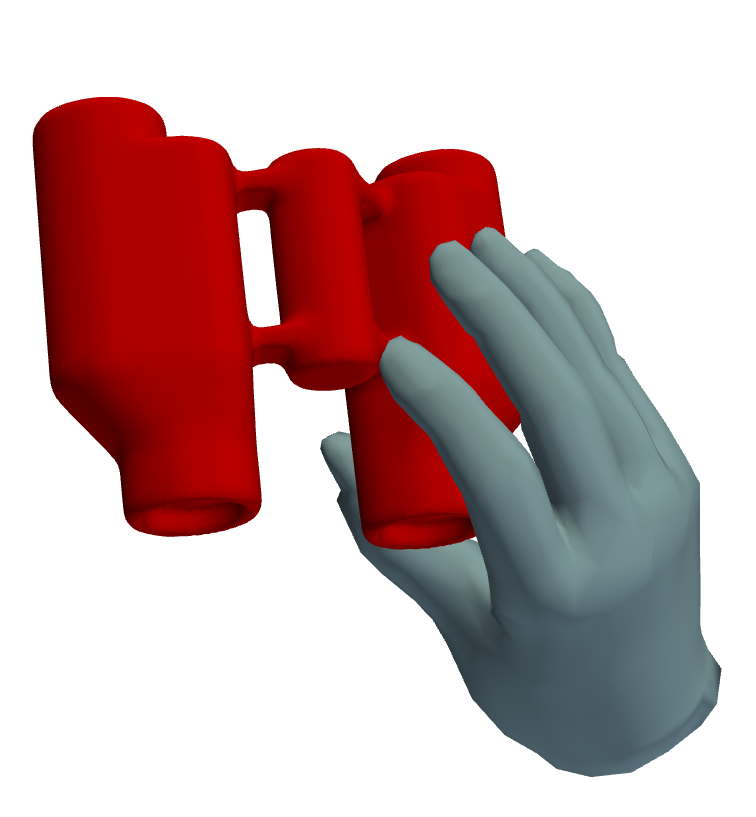} \\
      \includegraphics[scale=0.1]{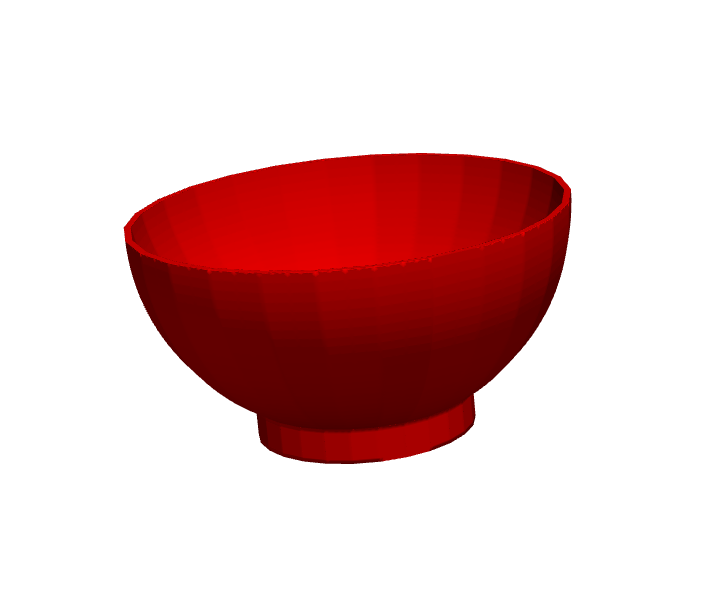} & 
      \includegraphics[scale=0.1]{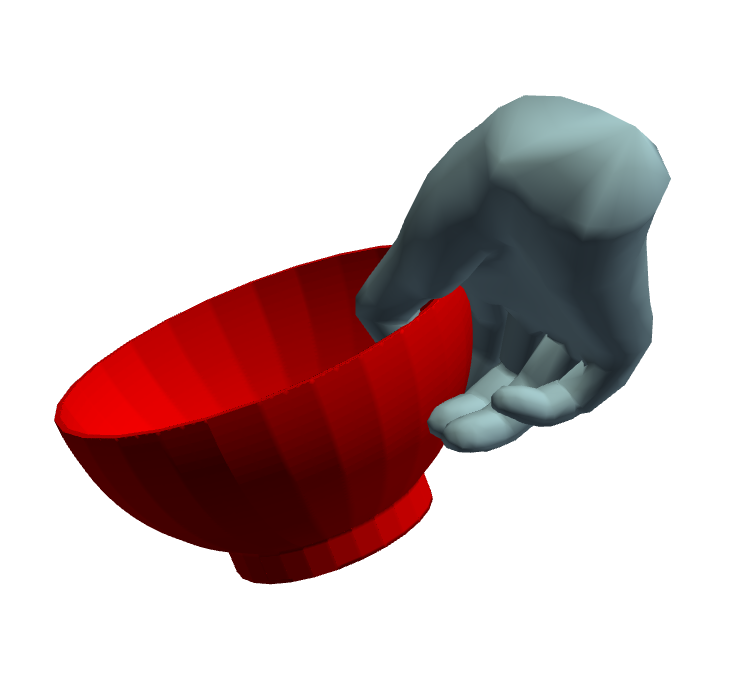} & 
      \includegraphics[scale=0.1]{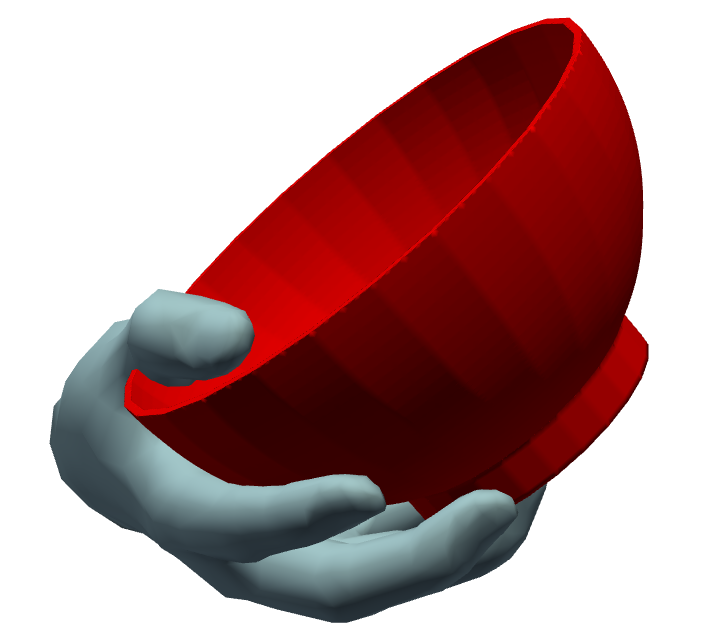} & 
      \includegraphics[scale=0.1]{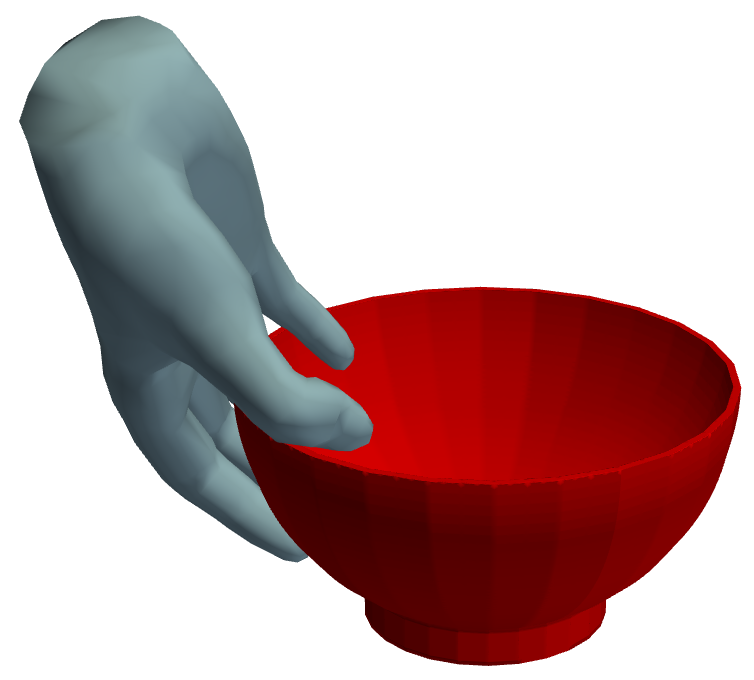} \\
    \end{tblr}
    \captionsetup{type=figure}
     \captionof{figure}{Qualitative evaluation of our method, \modelname, trained on the object modality of input
     for grasp synthesis. Each sample is generated from the same input, the object mesh in canonical pose.
     \modelname produces plausible grasps with minimal mesh penetration and consistent finger contacts.}
    \label{fig:gen1}
 \end{table*}

\bigskip

\begin{table*}[h]
    \centering
    \begin{tblr}{
      colspec = {X[c,j]|X[c,j]X[c,j]X[c,j]},
      stretch = 0,
      rowsep = 6pt,
    }
      Input & Sample 1 & Sample 2 & Sample 3 \\
     \hline
      \includegraphics[scale=0.12]{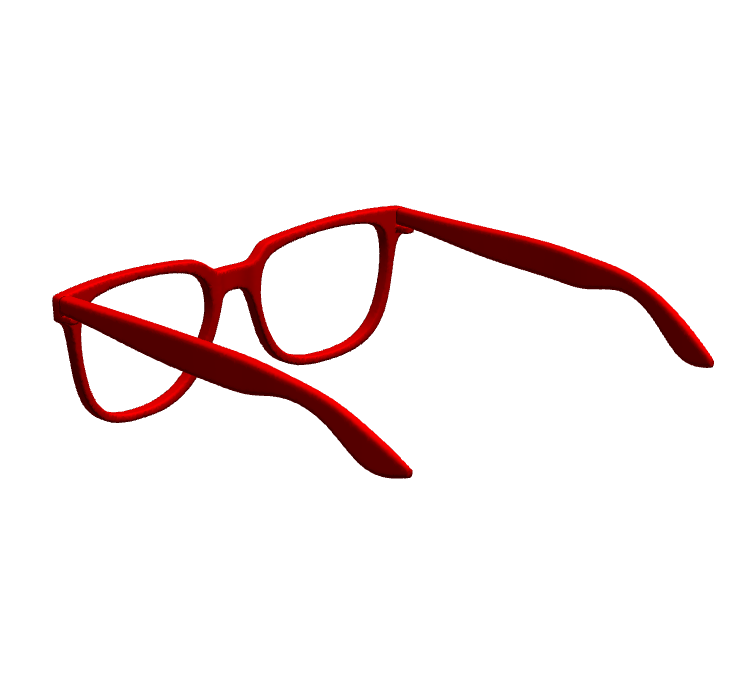} & 
      \includegraphics[scale=0.12]{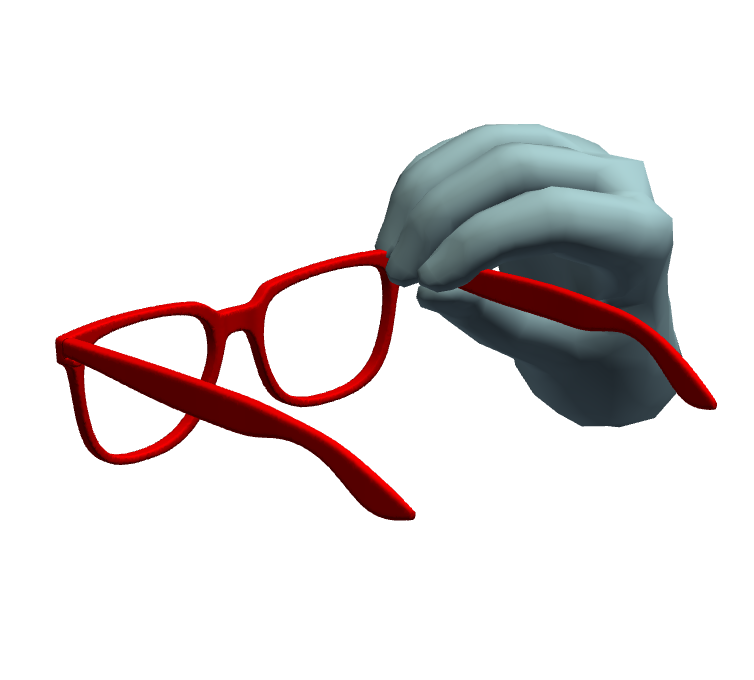} &
      \includegraphics[scale=0.12]{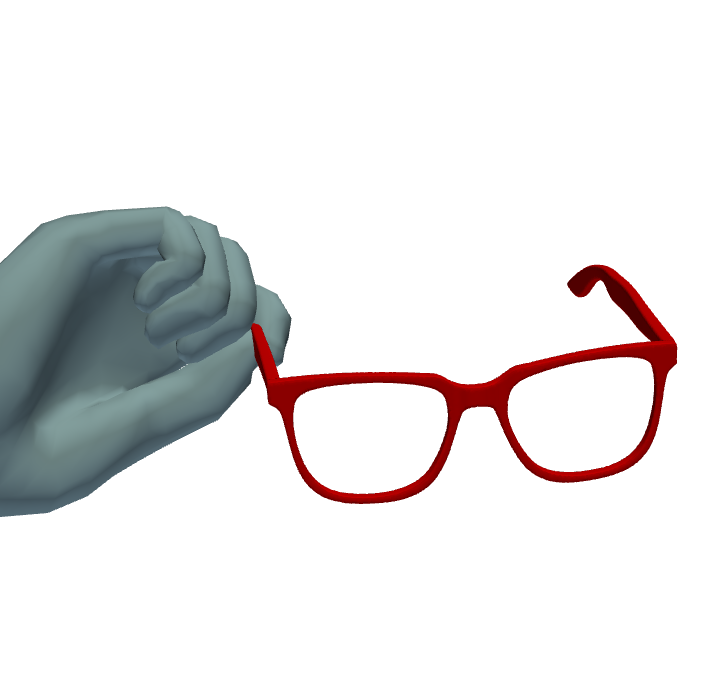} &
      \includegraphics[scale=0.1]{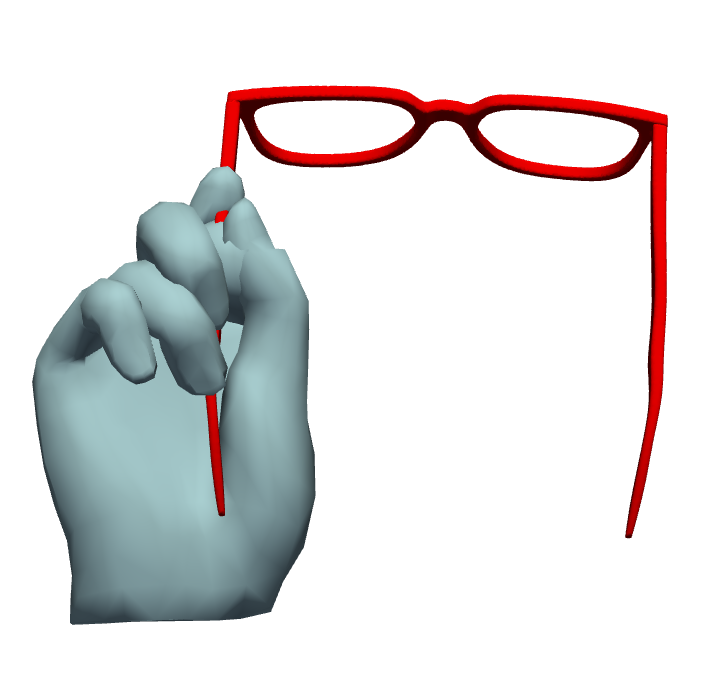} \\ 
      \includegraphics[scale=0.12]{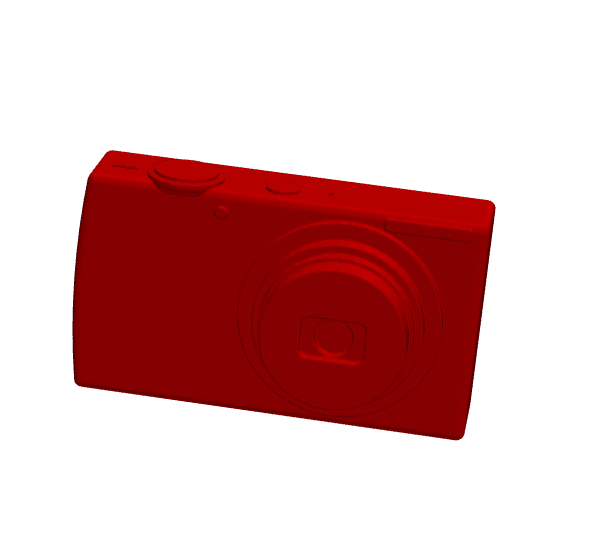} & 
      \includegraphics[scale=0.12]{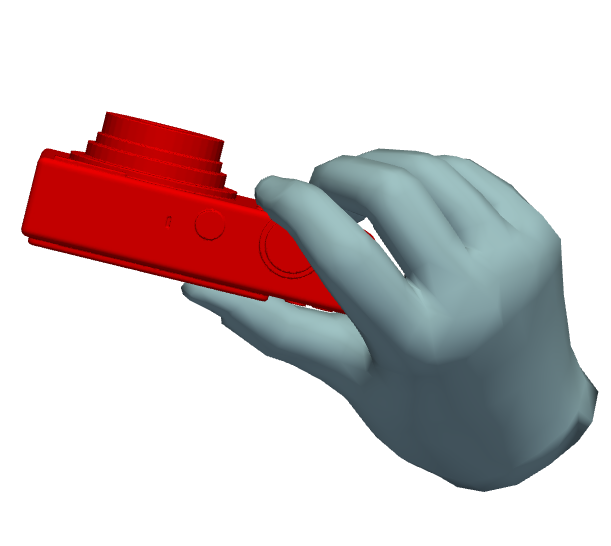} & 
      \includegraphics[scale=0.12]{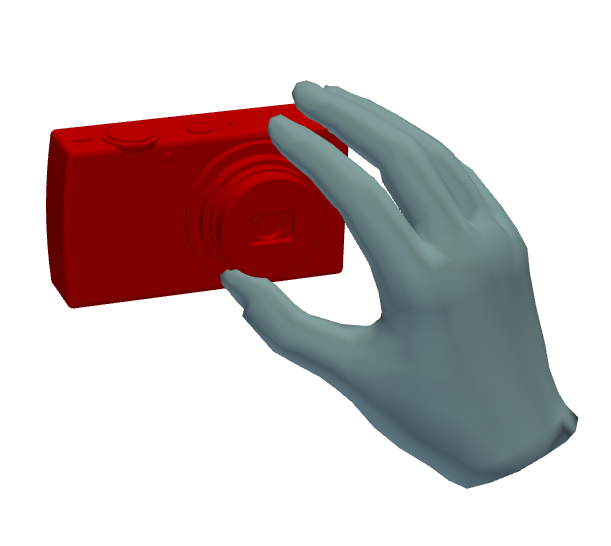} & 
      \includegraphics[scale=0.12]{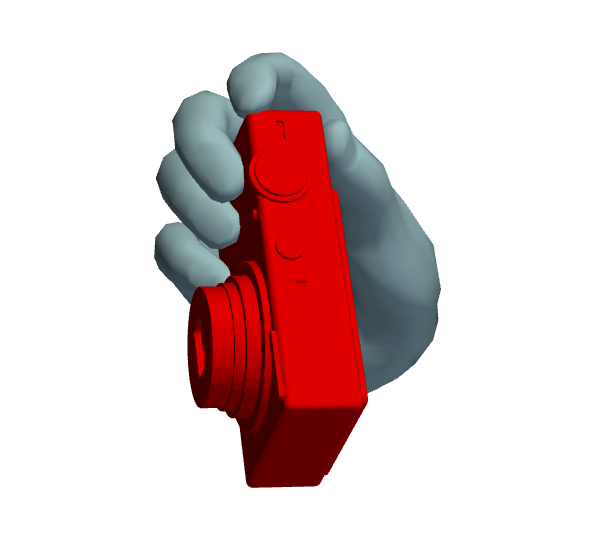} \\ 
      \includegraphics[scale=0.12]{figures/contactpose/synthesis/Screenshot_2024-03-14_at_17.19.47.png} & 
      \includegraphics[scale=0.12]{figures/contactpose/synthesis/Screenshot_2024-03-14_at_17.19.56.png} & 
      \includegraphics[scale=0.12]{figures/contactpose/synthesis/Screenshot_2024-03-14_at_17.20.07.png} & 
      \includegraphics[scale=0.12]{figures/contactpose/synthesis/Screenshot_2024-03-14_at_17.20.21.png} \\
      \includegraphics[scale=0.12]{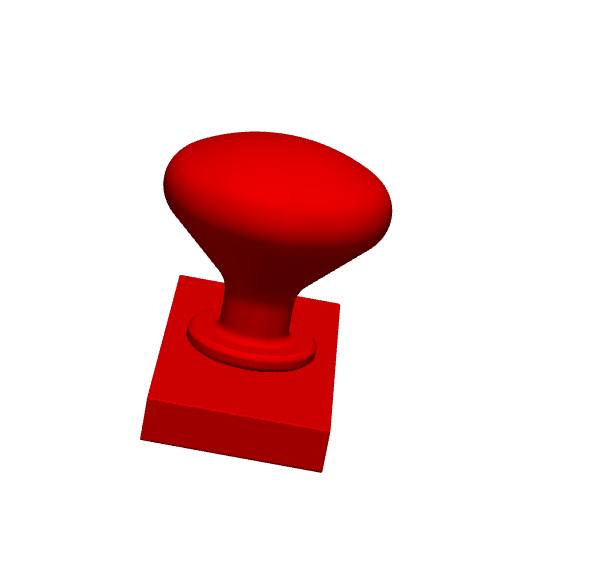} & 
      \includegraphics[scale=0.12]{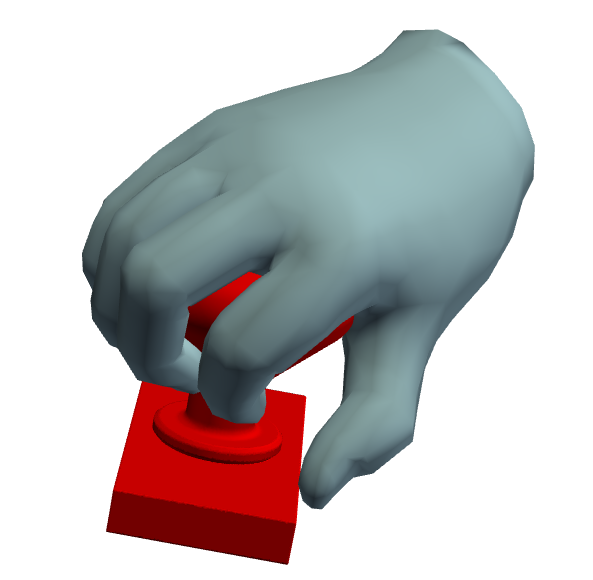} & 
      \includegraphics[scale=0.12]{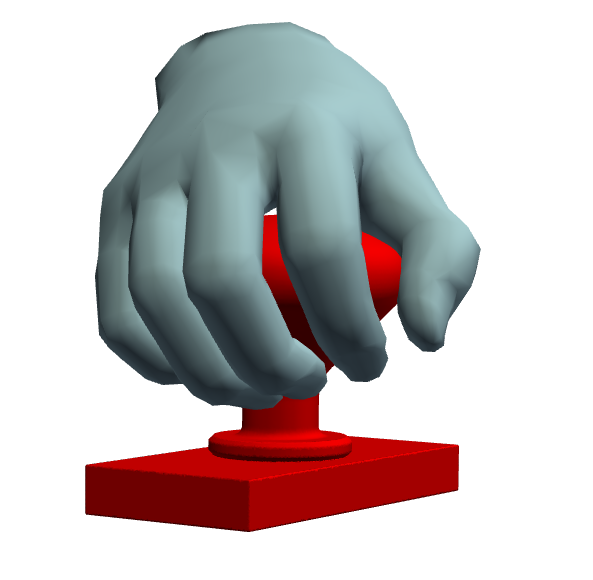} & 
      \includegraphics[scale=0.12]{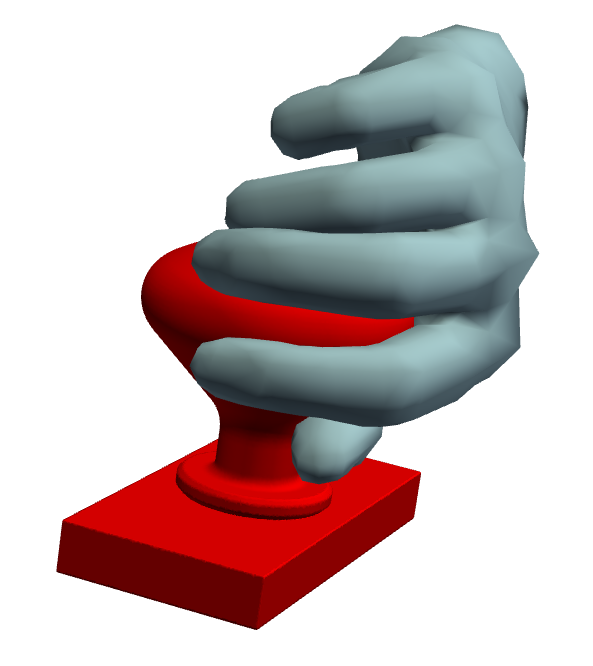} \\
    \end{tblr}
    \captionsetup{type=figure}
     \captionof{figure}{Qualitative evaluation of our method, \modelname, trained on the object modality of input
     for grasp synthesis. Each sample is generated from the same input, the object mesh in canonical pose.
     \modelname produces plausible grasps with minimal mesh penetration and consistent finger contacts.}
    \label{fig:gen2}
 \end{table*}

\bigskip

\clearpage

\subsection{Multimodal model: grasp refinement \& synthesis}\label{ap:multimodal}
We further explore our model expressiveness by jointly training two context
encoders along with the diffusion backbone of \modelname, as opposed to separately
trained models for object conditioning and noisy hand-object pair conditioning.
\cref{fig:multi-gen} shows qualitative results of the grasp synthesis from this
model, while \cref{fig:multi-refinement} shows qualitative results of the grasp
denoising task for the same model. We trained two multimodal models: one on the
ContactPose\cite{Brahmbhatt2020ContactPoseAD} dataset, and one on the
OakInk\cite{YangCVPR2022OakInk} dataset which we only evaluate on grasp synthesis.

\begin{table}[tb]
    \centering
    \caption{
        Evaluation of our multimodal model on static grasp generation against two state-of-the-art methods on two benchmarks.
        \modelname outperforms GraspTTA\cite{jiang2021graspTTA} on the
        ContactPose benchmark\cite{Brahmbhatt2020ContactPoseAD} and is on par
        with GrabNet\cite{GRAB:2020} on the OakInk
        benchmark\cite{YangCVPR2022OakInk}.
        We used reported metrics for GrabNet\cite{GRAB:2020} from the OakInk paper\cite{YangCVPR2022OakInk} and sampled
        one grasp per dataset sample for our method on both benchmarks. Best results are in bold.
    }
    \resizebox{\columnwidth}{!}{
        \begin{tabular}{lcccc}
        & \multicolumn{2}{c}{ContactPose\cite{Brahmbhatt2020ContactPoseAD}} & \multicolumn{2}{c}{OakInk\cite{YangCVPR2022OakInk}}\\
        \hline
        \textbf{Method} & \textbf{IV} ($\text{cm}^3$) $\downarrow$ & \textbf{SD} ($\text{cm}$) $\downarrow$ & \textbf{IV} ($\text{cm}^3$) $\downarrow$ & \textbf{SD} ($\text{cm}$) $\downarrow$  \\
        \hline  \hline
        GraspTTA \cite{jiang2021graspTTA} & 5.17 & \textbf{3.81} & - & - \\
        GrabNet \cite{GRAB:2020} & - & - & 6.60 & \textbf{1.21} \\
        \modelname & \textbf{5.13} & 5.80 & \textbf{5.98} & 5.84 \\
        \end{tabular}
    }
    \label{tab:multimodal-gen}
\end{table}

A quantitative evaluation on the denoising task is shown on
\cref{tab:multimodal-denoising}, and one on the generation task
is shown on \cref{tab:multimodal-gen}. For the latter, the increase in
simulation displacement (SD) for our method with contact fitting suggests that
some hand penetration is helpful to a stable grasp. Note that the synthetic
nature of most OakInk samples results in incorrect vertex normals, adversely
affecting our penetration regularization loss and performance. This could be
solved with a different approach to penetration regularization, such as via the
signed distance function.

\begin{table*}[tb]
    \centering
    \caption{
        Evaluation of our approach on static grasp refinement against two SOTA methods and our baseline
        on the Perturbed ContactPose benchmark. 
        * means reported figures.
        Our multimodal model is marked with \textdagger.
        Both \modelname variants were evaluated with one non-cherry-picked
        generated grasp per sample. While our baseline yields better
        reconstruction accuracy in absolute pose, our full model \modelname
        shows greater contact accuracy and outperforms
        ContactOpt\cite{Grady2021ContactOptOC} and TOCH\cite{Zhou2022TOCHSO} on
        almost all metrics. The multimodal version still outperforms these
        baselines on contact-based metrics and IV score for grasp refinement,
        while also being able to do grasp synthesis. Best results are in bold,
        second best are underlined.
    }
    \resizebox{\textwidth}{!}{
        \begin{tabular}{lcccccc}
        \textbf{Method} & \textbf{MPJPE} (mm) $\downarrow$ & \textbf{R-MPJPE} (mm) $\downarrow$ & \textbf{IV} ($\text{cm}^3$) $\downarrow$ & \textbf{F1} (\%) $\uparrow$ & \textbf{Precision} (\%) $\uparrow$ & \textbf{Recall} (\%) $\uparrow$\\
        \hline  
        Perturbed data & 83.02 & 21.55 & 6.99 & 1.55 & 1.88 & 2.74\\
        ContactOpt\cite{Grady2021ContactOptOC}  & 32.88 & \underline{28.17} & 12.83* & 17.27 & 13.24 & \textbf{34.30}\\ %
        TOCH \cite{Zhou2022TOCHSO} & \textbf{26.96} & 29.24 & 10.14 & 22.23 & 21.46 & 25.09 \\ 
        \modelname & 27.69 & \textbf{23.54} & \underline{6.04} & \textbf{27.20} & \textbf{25.21} & \underline{32.80} \\
        \modelname \textdagger & 35.45 & 33.10 & \textbf{5.62} & \underline{24.88} & \underline{23.87} & 29.24 \\
        \end{tabular}
    }
    \label{tab:multimodal-denoising}
\end{table*}

\begin{table*}[h]
    \centering
    \begin{tblr}{
      colspec = {X[c,j]|X[c,j]X[c,j]X[c,j]},
      stretch = 0,
      rowsep = 6pt,
    }
      Input & Sample 1 & Sample 2 & Sample 3 \\
     \hline
      \includegraphics[scale=0.08]{figures/contactpose/synthesis/Screenshot_2024-03-14_at_17.11.59.png} & 
      \includegraphics[scale=0.15]{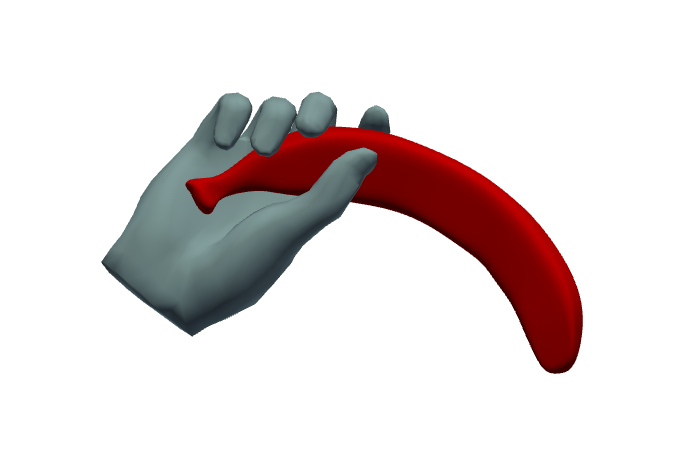} &
      \includegraphics[scale=0.15]{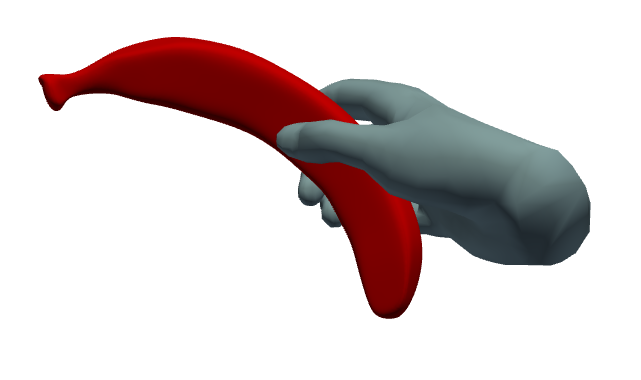} &
      \includegraphics[scale=0.15]{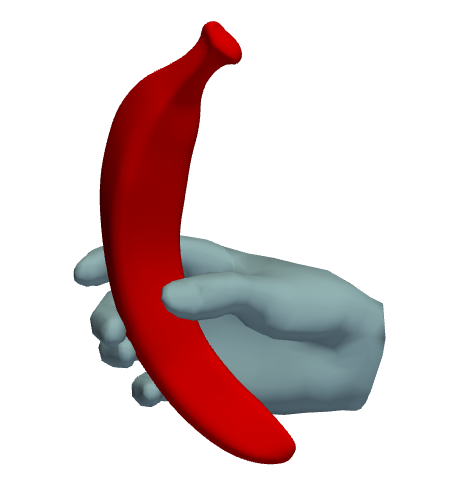} \\ 
      \includegraphics[scale=0.08]{figures/contactpose/synthesis/Screenshot_2024-03-14_at_17.14.08.png} & 
      \includegraphics[scale=0.13]{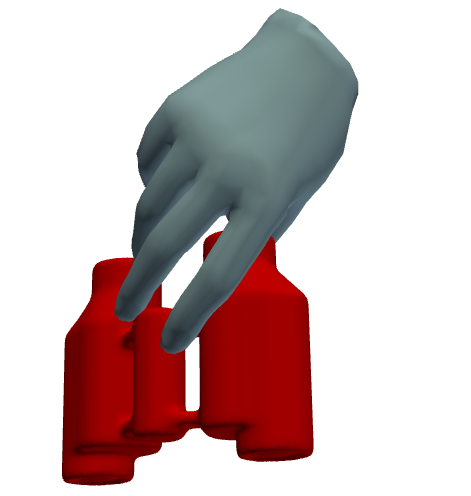} & 
      \includegraphics[scale=0.13]{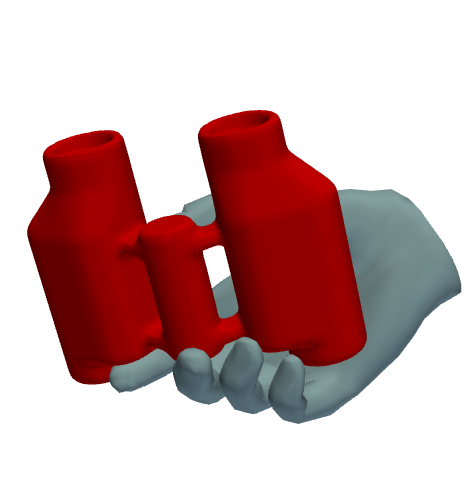} & 
      \includegraphics[scale=0.13]{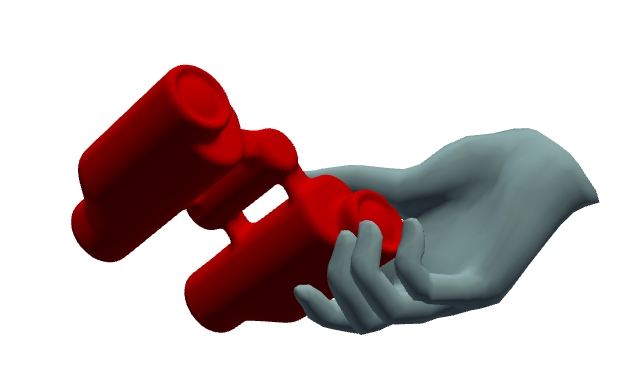} \\ 
      \includegraphics[scale=0.1]{figures/contactpose/synthesis/Screenshot_2024-03-14_at_17.15.21.png} & 
      \includegraphics[scale=0.15]{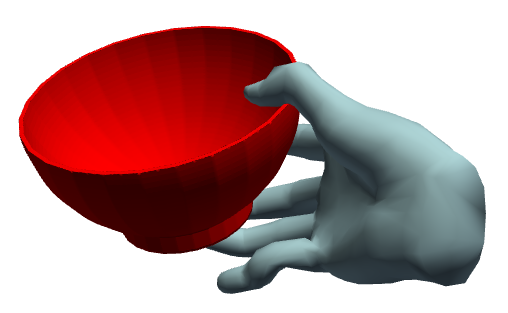} & 
      \includegraphics[scale=0.15]{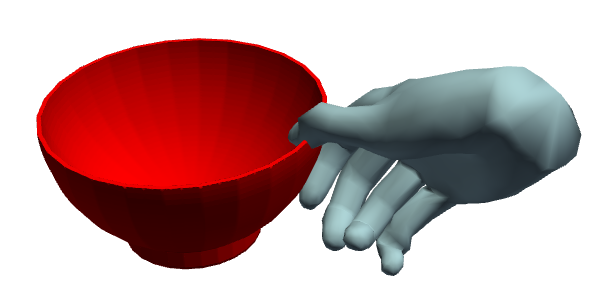} & 
      \includegraphics[scale=0.15]{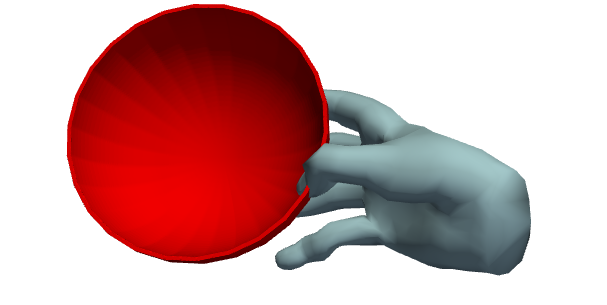} \\
      \includegraphics[scale=0.1]{figures/contactpose/synthesis/Screenshot_2024-03-14_at_17.18.27.png} & 
      \includegraphics[scale=0.15]{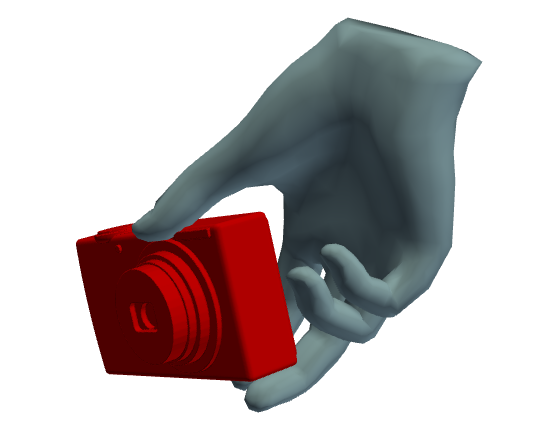} & 
      \includegraphics[scale=0.15]{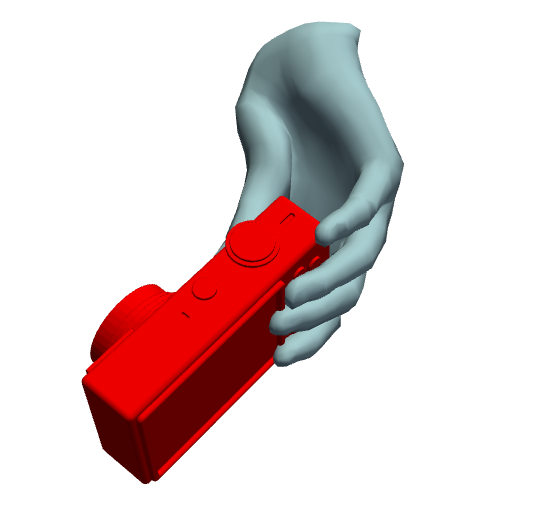} & 
      \includegraphics[scale=0.15]{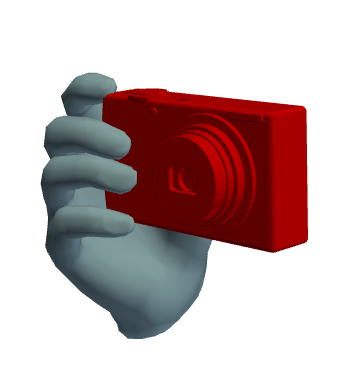} \\
    \end{tblr}
    \captionsetup{type=figure}
     \captionof{figure}{Qualitative evaluation of our multimodal \modelname, trained on both object and noisy hand-object pair
     modalities, in the grasp synthesis setting. Each sample is generated from the same input, the object mesh in canonical pose.
     \modelname produces plausible grasps with minimal mesh penetration and consistent finger contacts.}
    \label{fig:multi-gen}
 \end{table*}

\bigskip

\begin{table*}[h]
    \centering
    \begin{tblr}{
      colspec = {X[c,j]X[c,j]|X[c,j]},
      stretch = 0,
      rowsep = 6pt,
    }
      Ground truth & Observation & Prediction \\
     \hline
      \includegraphics[scale=0.15]{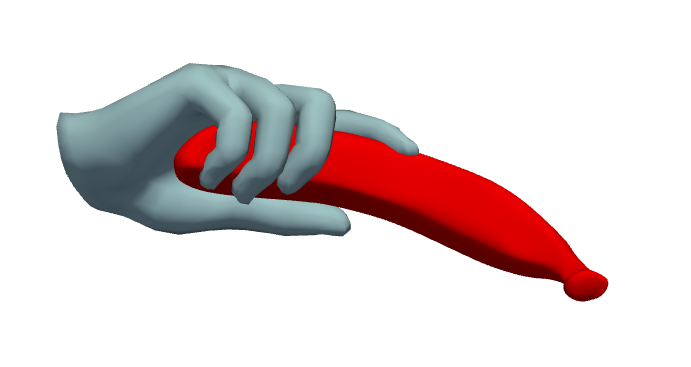} & 
      \includegraphics[scale=0.12]{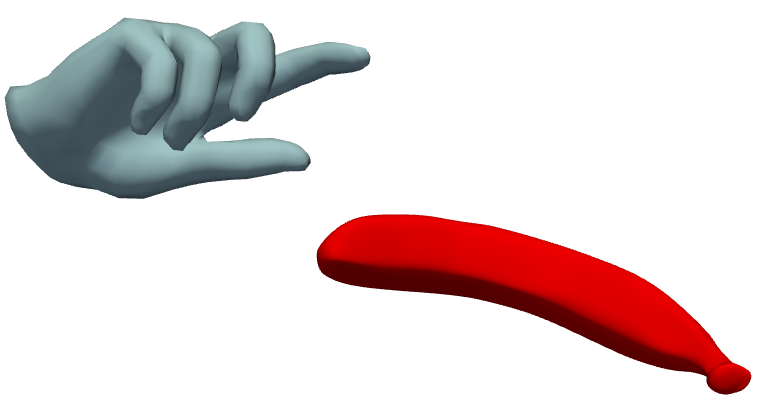} &
      \includegraphics[scale=0.15]{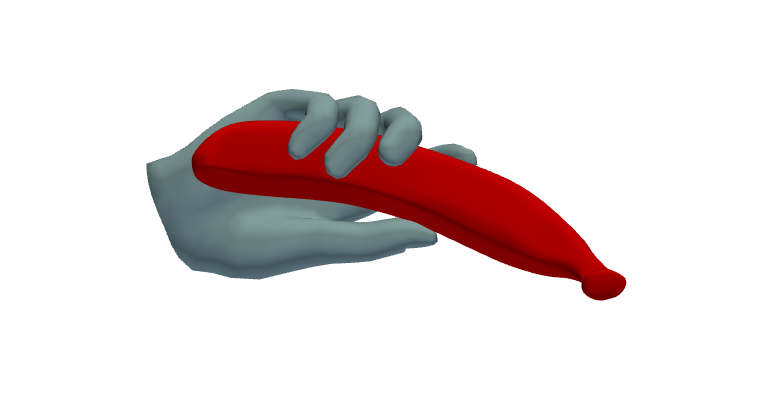} \\ 
      \includegraphics[scale=0.15]{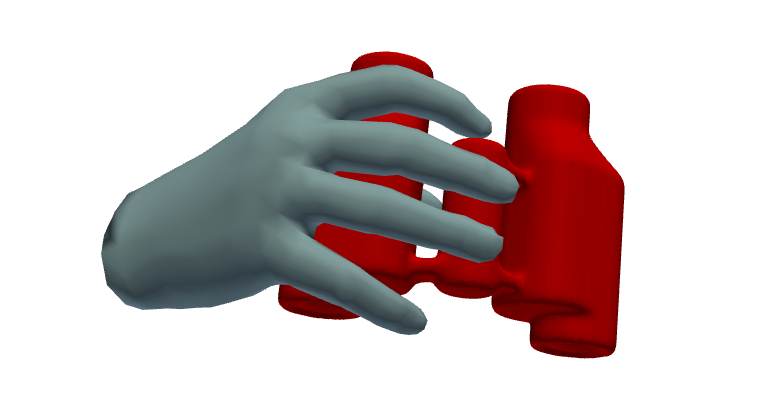} & 
      \includegraphics[scale=0.15]{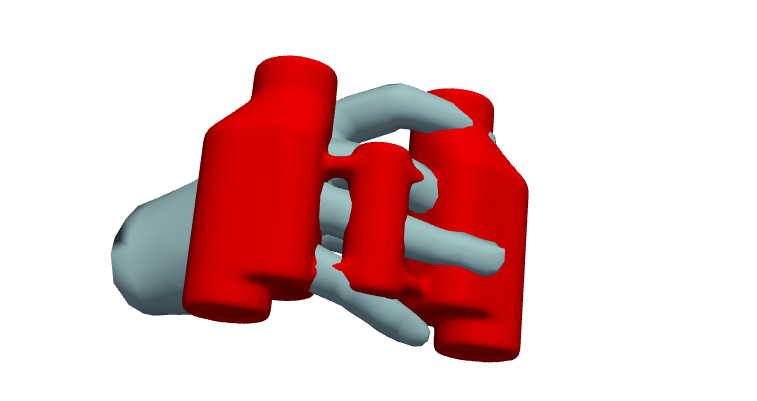} & 
      \includegraphics[scale=0.15]{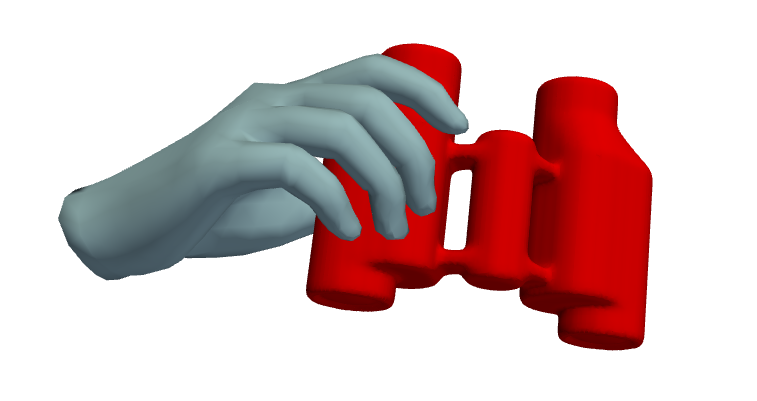} \\ 
      \includegraphics[scale=0.15]{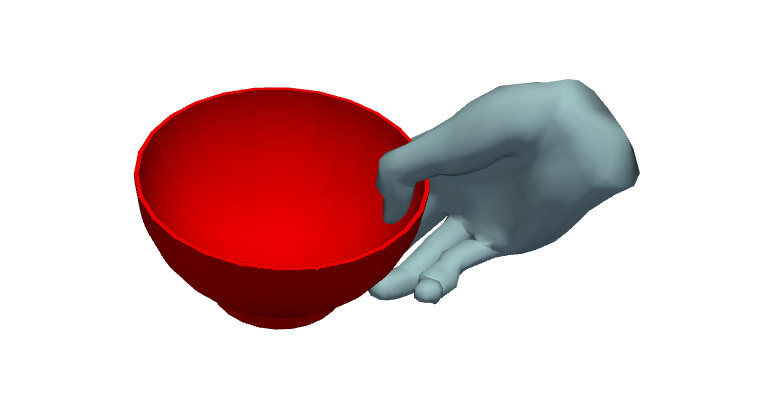} & 
      \includegraphics[scale=0.15]{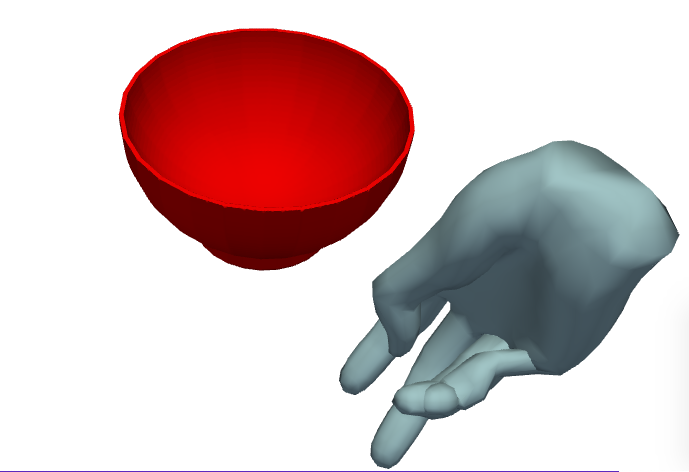} & 
      \includegraphics[scale=0.15]{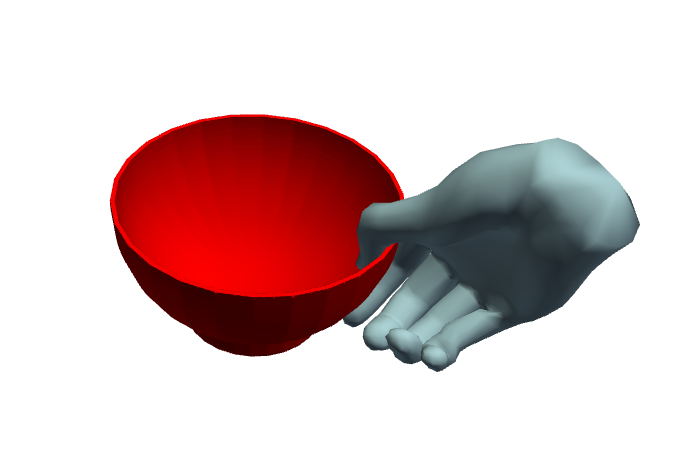} \\ 
      \includegraphics[scale=0.15]{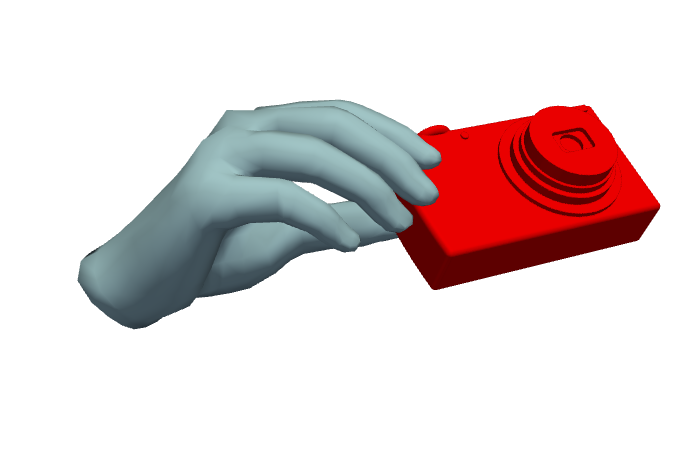} & 
      \includegraphics[scale=0.15]{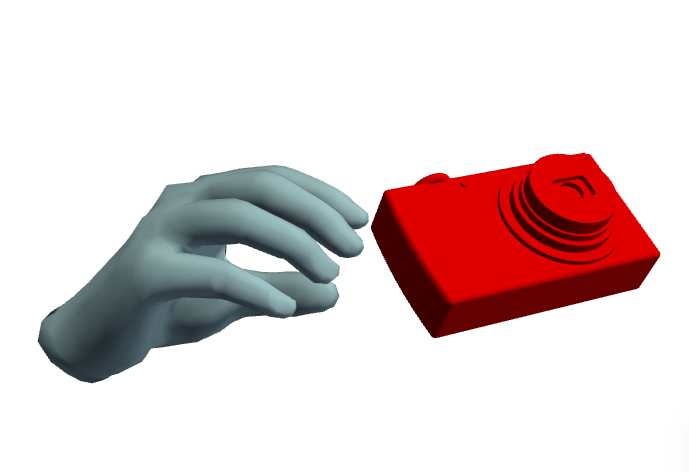} & 
      \includegraphics[scale=0.15]{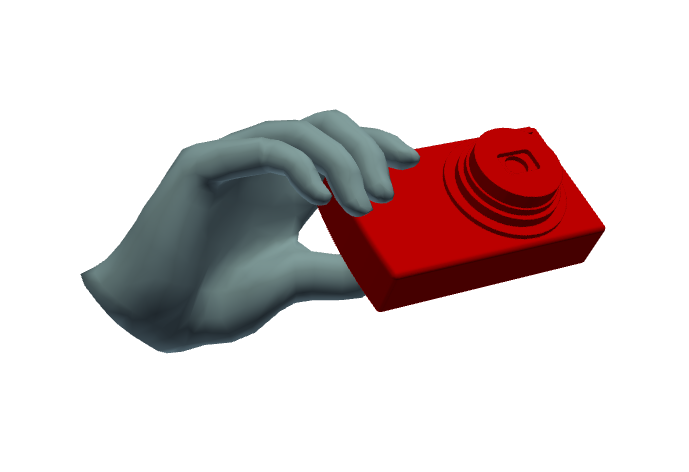} \\ 
    \end{tblr}
    \captionsetup{type=figure}
     \captionof{figure}{Qualitative evaluation of our multimodal \modelname, trained on both object and noisy hand-object
     pair modalities, in the grasp refinement setting. \modelname produces plausible grasps with minimal mesh penetration
     and respects finger contacts from the ground-truth mesh.}
    \label{fig:multi-refinement}
 \end{table*}

\bigskip

\end{document}